\documentclass{article} 
\usepackage{arxiv_verrsion,times} 

\usepackage[dvipsnames]{xcolor}
\usepackage[pagebackref,breaklinks,colorlinks,allcolors=Plum]{hyperref}
\usepackage{url}

\usepackage{amsmath,amsfonts,bm}

\def\eqref#1{equation~\ref{#1}}

\def\1{\bm{1}}

\DeclareMathAlphabet{\mathsfit}{\encodingdefault}{\sfdefault}{m}{sl}
\SetMathAlphabet{\mathsfit}{bold}{\encodingdefault}{\sfdefault}{bx}{n}

\usepackage[dvipsnames]{xcolor}

\usepackage{algorithm}
\usepackage{algpseudocode}
\usepackage{booktabs}
\usepackage{multirow}      
\usepackage{makecell}  
\usepackage{amsmath,amssymb}
\usepackage{amsthm}        
\usepackage{courier}       
\newtheorem{theorem}{Theorem}[section]
\newtheorem{proposition}{Proposition}[section]
\theoremstyle{remark}

\usepackage{xspace}
\usepackage{graphicx}
\usepackage{xcolor}
\usepackage{colortbl}
\usepackage{wrapfig}
\usepackage{graphicx}
\usepackage{subcaption}
\usepackage{bbm}
\usepackage{mathtools}
\usepackage[nameinlink,capitalize,noabbrev]{cleveref}

\algrenewcommand\algorithmicrequire{\textbf{Input:}}
\algrenewcommand\algorithmicensure{\textbf{Output:}}

\newcommand{\val}[2]{%
  \begin{tabular}{@{}c@{}}#1\\[-2pt]{\scriptsize \(\pm\)#2}\end{tabular}%
}
\newcommand{\bval}[2]{%
  \begin{tabular}{@{}c@{}}\textbf{#1}\\[-2pt]{\scriptsize \(\pm\)#2}\end{tabular}%
}

\usepackage{pifont}
\newcommand{\cmark}{\ding{51}} 
\newcommand{\xmark}{\ding{55}} 

\usepackage{enumitem}

\newcommand{\ie}{\textit{i}.\textit{e}., }

\newcommand{\ours}{\scalebox{0.8}[0.8]{\textbf{GAIN}}\xspace}

\definecolor{mycolor}{cmyk}{0.1, 0.1, 0, 0} 
\definecolor{sourcecolor}{gray}{0.95}

\title{Not Every Correction Helps:\\Gain-Guided Continual Test-Time Adaptation}

\author{
\textbf{Youjia Zhang}\textsuperscript{1}\textsuperscript{$\ast$} \
\textbf{Huiling Liu}\textsuperscript{1}\textsuperscript{$\ast$} \
\textbf{Soyun Choi}\textsuperscript{1} \
\textbf{Jaehong Yoon}\textsuperscript{2} \
\textbf{Sungeun Hong}\textsuperscript{1}\textsuperscript{$\dagger$} \\ [0.2cm]
\textsuperscript{1}Sungkyunkwan University \quad \textsuperscript {2} Nanyang Technological University
}

\iclrfinalcopy 

\begin{document}
\maketitle

\begingroup
\renewcommand{\thefootnote}{}
\footnotetext{$\ast$ Equal contribution.}
\footnotetext{$\dagger$ Corresponding author: Sungeun Hong (csehong@skku.edu)}
\endgroup

\begin{abstract}

Continual test-time adaptation (CTTA) adapts a source model to an unlabeled test stream whose distribution may change over time.
Existing TTA methods often assess prediction reliability using confidence or entropy, which primarily reflect the model's self-certainty for the current sample.
In CTTA, accumulated target observations can provide complementary evidence for correcting the source prediction, but this history may become misaligned as the target distribution changes.
The key question is therefore not how much the correction differs from the source prediction, but whether and how strongly it should be applied.
This paper proposes \textbf{G}ain-\textbf{A}ware \textbf{IN}tervention (\textbf{\ours}), a backpropagation-free CTTA framework guided by a simple principle: \textit{history proposes, gain decides}.
\ours maintains compact target statistics to form a correction proposal and a posterior-predictive evaluator that accounts for estimation uncertainty.
The resulting source-relative gain estimates the proposal's benefit and determines a sample-specific intervention strength along a continuous path through efficient one-dimensional optimization.
Gain-controlled predictions then update the target statistics online, limiting the propagation of unreliable corrections, all without backpropagation, sample storage, or replay.
Across five benchmarks, our method achieves strong predictive performance, with favorable accuracy--calibration--efficiency trade-offs in continual adaptation. On ImageNet-C, for example, \ours achieves 61.9\% accuracy with near-source calibration. It remains stable under diverse and challenging continual shifts while running $15.9\times$ faster than a representative optimization-based CTTA baseline.


\end{abstract}    
\section{Introduction}
Continual test-time adaptation (CTTA)~\citep{CoTTA} adapts a source model online to an unlabeled target stream whose distribution evolves over time. A central challenge is deciding when to retain the source prediction and when to correct it.
Existing methods assess adaptation reliability from the current sample through entropy-based objectives~\citep{Tent, EcoTTA, REM}, reliability-aware sample selection~\citep{EATA, DeYO, Sotta}, or region-level confidence modeling~\citep{ReCAP}.
Such measures can be useful indicators of prediction reliability, yet they primarily characterize the model's \emph{self-certainty}. They do not explicitly capture how the same prediction may be supported differently by the evolving target context. This observation motivates us to look beyond the current output and treat accumulated target observations as additional predictive evidence.

Recent CTTA methods exploit historical target information through feature statistics or distribution estimation~\citep{DPCore, DOTA, ADAPT}, allowing accumulated target information to guide subsequent predictions. However, this creates a distinct challenge in continual adaptation. As the target distribution evolves, historical statistics may become misaligned with the current domain, while erroneous corrections can accumulate in the target state and affect subsequent adaptation.
Consequently, a history-induced correction is not necessarily beneficial simply because it differs from the source prediction.
This raises a more fundamental question for CTTA: not whether target history suggests a correction, but whether that correction is worth applying.

\begin{figure*}[!t]
\vspace{-5mm}
    \centering
    \setlength{\abovecaptionskip}{0.3cm}
    \includegraphics[width=0.9\linewidth]{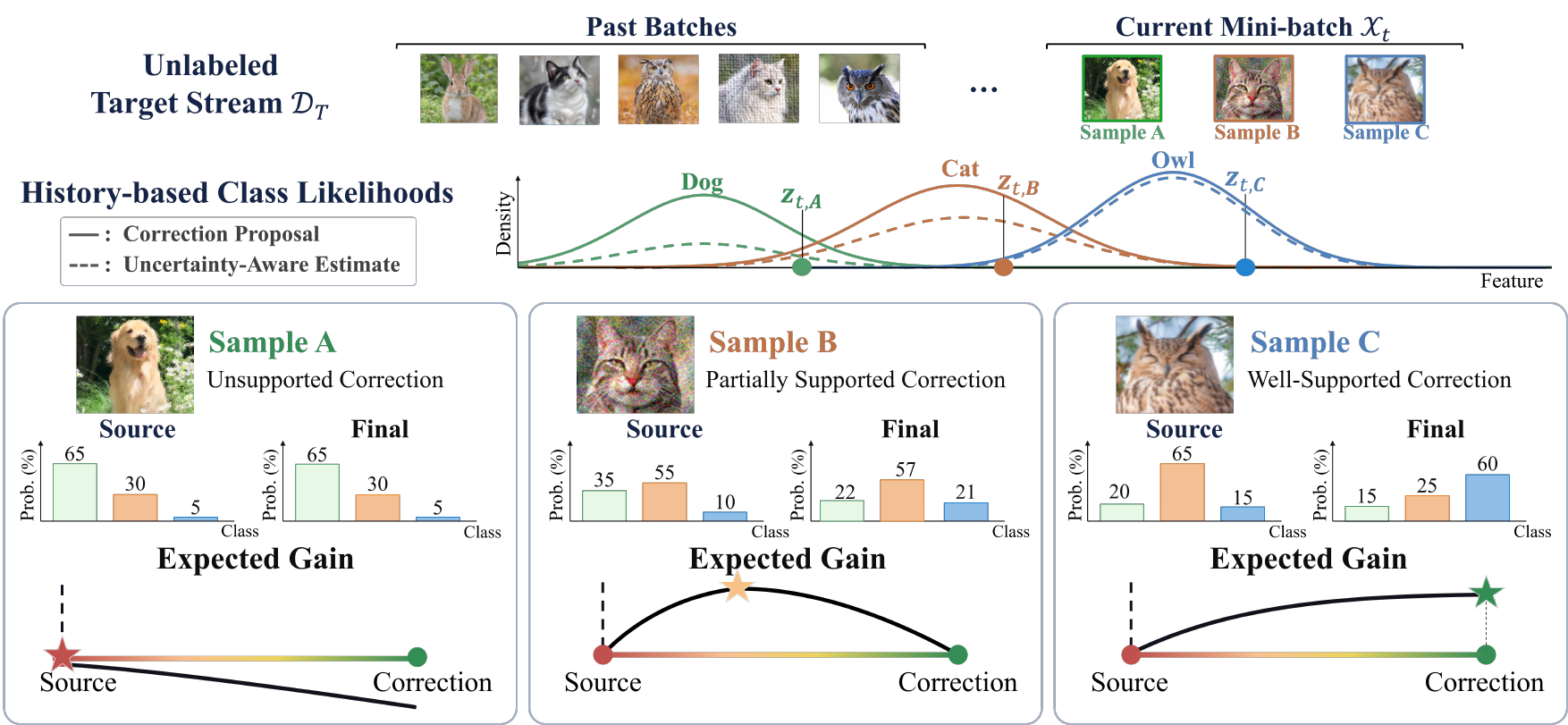}
\caption{
\textbf{Gain-Guided Sample-wise Intervention.}
Past target batches are summarized by compact statistics to construct a history-based correction proposal (solid) and an uncertainty-aware posterior-predictive estimate (dashed) that evaluates it. The resulting source-relative gain determines a sample-specific strength along the source-to-correction path: unsupported corrections are rejected (Sample A), partially supported corrections are applied conservatively (Sample B), and well-supported corrections are fully adopted (Sample C).
}

\label{fig:main_figure1}
    \vspace{-3mm}
\end{figure*}

Motivated by this perspective, we introduce \textbf{G}ain-\textbf{A}ware \textbf{IN}tervention (\textbf{\ours}), following the principle that \emph{history proposes, gain decides}. \ours uses accumulated target statistics to propose a target-side correction and evaluates its source-relative gain while accounting for uncertainty in the evolving target statistics. The resulting gain determines whether and how strongly the correction should influence the current prediction.

To leverage correction gain for reliable continual adaptation, \ours combines sample-specific intervention with online target-state maintenance. As shown in Fig.~\ref{fig:main_figure1}, the intervention strength is selected along a continuous path from source retention to full target correction through a concave one-dimensional objective with an efficient global solution.
The resulting predictions then update the target state causally, limiting the propagation of unsupported history-induced corrections to subsequent adaptation.
Throughout adaptation, the source model remains frozen, requiring no backpropagation, parameter updates, or replay of past target samples.
Across diverse continual-shift settings, \ours improves accuracy while maintaining strong calibration and long-horizon stability.

\begin{itemize}[leftmargin=20pt]


    \item We introduce a \emph{source-relative correction-gain} perspective for CTTA, shifting the focus from the apparent reliability of the current prediction to whether a history-derived correction improves upon source retention.
    
    \item \ours analytically marginalizes class-center uncertainty and selects a sample-specific intervention through a concave one-dimensional evidence-path objective, while updating compact target statistics online without sample storage or replay.

    \item Extensive experiments demonstrate that \ours consistently achieves a strong accuracy--calibration--efficiency trade-off across structured, dynamic, mixed-domain, and long-horizon continual shifts.
\end{itemize}

\vspace{-3mm}
\section{Related Work}
\label{sec:related}

\vspace{-2mm}
\paragraph{Reliability and Stability in Test-Time Adaptation.}
CTTA extends test-time adaptation to evolving unlabeled streams, where repeated self-adaptation can amplify prediction errors and lead to long-term instability~\citep{CoTTA}.
Prior work improves reliability through entropy-based adaptation and sample filtering or reweighting~\citep{Tent,EATA,DeYO,DSS}, stabilized or accelerated optimization~\citep{EcoTTA,SAR,REM,LCoTTA,AEA}, and representation, structural, geometric, or subspace adaptation~\citep{C-MAE,BGD,PAID,TCA,SaTeen,NEO,GOLD}.
CAS~\citep{CAS2026} uses cross-augmentation similarity to make a binary adapt-or-skip decision when adaptation may be harmful.
Beyond this binary decision, \ours evaluates whether a history-induced correction improves over retaining the source prediction and continuously controls its intervention strength through source-relative gain.

\vspace{-1mm}
\paragraph{Knowledge Preservation in CTTA.}
Under continual shifts, preserving useful knowledge is important for mitigating forgetting and cross-domain interference. Existing methods preserve or reuse information through source-weight restoration or ensembling~\citep{CoTTA,ROID}, sample storage~\citep{RoTTA}, domain-specific modules or experts~\citep{ViDA,BECoTTA,FreqCTTA}, and compact prompt or knowledge pools~\citep{FOA,DPCore,KFF}. More recently, DO-ALL~\citep{DOALL2026} improves long-term stability by distilling synthetic source anchors for replay during adaptation. In contrast, \ours keeps the source model frozen and maintains only compact target statistics, using gain-guided intervention to regulate how history-induced corrections influence current predictions and subsequent target-state updates without sample storage or replay.

\vspace{-1mm}
\paragraph{Distributional Modeling in CTTA.}
Distributional modeling has been increasingly explored for continual adaptation. PETAL~\citep{PETAL} formulates lifelong TTA probabilistically, while BayesTTA~\citep{BayesTTA} incrementally models class-conditional distributions under temporal shifts. DOTA~\citep{DOTA} further estimates evolving test-time feature distributions and derives posterior predictions from accumulated target statistics. Related statistical TTA methods incorporate source-informed priors or analytic inference~\citep{StatA,ADAPT,MDAA}. While these approaches improve target-side estimation, accumulated statistics may become inaccurate or stale as the target distribution changes. Rather than directly treating the target estimate as the final prediction, \ours uses it as a correction proposal and evaluates its source-relative gain to determine whether and how strongly it should influence the source prediction, thereby limiting the propagation of unreliable corrections through subsequent adaptation.

\vspace{-1mm}
\section{Preliminaries}
\vspace{-1mm}
\subsection{Continual Test-Time Adaptation}
\label{sec:preliminaries}
\vspace{-1mm}
\paragraph{Problem Setup.}
Given a source model $f_{\theta}$ pre-trained on a labeled source domain $\mathcal{D}_S$, continual test-time adaptation (CTTA) considers an unlabeled target stream $\mathcal{D}_T\!=\!\{\mathcal{X}_t\}_{t=1}^{T}$ whose distribution may change over time. We decompose the source model as $f_{\theta}\!=\!g_{\theta}\circ\phi_{\theta}$, where $\phi_{\theta}$ and $g_{\theta}$ denote the feature extractor and classifier head, respectively. At time $t$, the model observes a target mini-batch $\mathcal{X}_t$ with $|\mathcal{X}_t|\!=\!B_t$. For each sample $\mathbf{x}_t\in\mathcal{X}_t$, the frozen source model produces
\vspace{-1mm}
\begin{equation}
\mathbf{s}_t=f_{\theta}(\mathbf{x}_t)\in\Delta^{K-1},
\qquad
\hat{y}_t^s=\arg\max_k s_{t,k},
\label{eq:source_prediction}
\end{equation}

\vspace{-3mm}
where $\mathbf{s}_t$ denotes the source predictive distribution and $\hat y_t^s$ is the corresponding predicted class.
Preceding target observations $\mathcal{H}_t\!=\!\bigcup_{\tau<t}\mathcal{X}_{\tau}$ form the accumulated target context. We keep $f_{\theta}$ frozen and exploit $\mathcal{H}_t$ for single-pass adaptation without gradient-based model updates.

\vspace{-2mm}
\paragraph{Beyond Source Confidence.}
Since target labels are unavailable, CTTA commonly relies on source-output proxies such as confidence or entropy to assess prediction reliability~\citep{Tent,EATA,REM}. However, these signals reflect only the model's self-certainty and ignore accumulated target context, under which the same source prediction may have different reliability (Appendix~\ref{app:source_reliability}). Historical observations can therefore provide complementary predictive information, but their relevance may diminish as the target distribution changes. We thus ask whether conditioning on $\mathcal H_t$ provides useful evidence for the current sample.

\vspace{-2mm}
\subsection{Accumulated Target Context as Predictive Evidence}
\label{sec:historical_evidence}
\vspace{-1mm}
\paragraph{Historical Target Evidence.}
Let $q_{t,k}^{0}\!\triangleq\! P_T(Y_t\!=\!k\mid\mathbf{x}_t)$ denote the target posterior probability for class $k$ given the current observation alone, and let $q_{t,k}^{H}\!\triangleq \!P_T(Y_t\!\!=\!\!k\!\mid\!\mathbf{x}_t,\mathcal{H}_t)$ denote the corresponding probability additionally conditioned on the accumulated target context.
By Bayes' rule,
\vspace{-1mm}
\begin{equation}
q_{t,k}^{H}
=
q_{t,k}^{0}
\frac{
P_T(\mathcal{H}_t\mid Y_t=k,\mathbf{x}_t)
}{
P_T(\mathcal{H}_t\mid\mathbf{x}_t)
},
\label{eq:history_conditioned_posterior}
\end{equation}
showing that historical context contributes class-dependent evidence beyond the current observation.
In log-probability space, this contribution is
\begin{equation}
\log q_{t,k}^{H}
=
\log q_{t,k}^{0}+\xi_{t,k},
\qquad
\xi_{t,k}
\triangleq
\log({q_{t,k}^{H}}/{q_{t,k}^{0}})
=
i(Y_t=k;\mathcal{H}_t\mid\mathbf{x}_t),
\label{eq:historical_evidence}
\end{equation}
where $\xi_{t,k}$ is the conditional pointwise mutual information (C-PMI)~\citep{PMI,C-PMI}. Positive and negative values indicate that the accumulated target context provides additional evidence for and against class $k$, respectively. Details are provided in Appendix~\ref{app:CPMI}.



\vspace{-2mm}
\begin{proposition}[Non-Negative Predictive Value of Target History]
\label{prop:nonnegative_history_value}
For any predictive distribution $\mathbf r\in\Delta^{K-1}$, define the logarithmic risk under the history-conditioned target posterior as $\mathcal R_t(\mathbf r)\triangleq\mathbb E_{Y_t\sim\mathbf q_t^H}[-\log r_{Y_t}]$. Conditioning on the accumulated target context $\mathcal H_t$ then yields non-negative predictive value under logarithmic loss:
\begin{equation}
\mathcal R_t(\mathbf q_t^0)
-
\mathcal R_t(\mathbf q_t^H)
=
D_{\mathrm{KL}}
(\mathbf q_t^H\Vert\mathbf q_t^0)
=
\mathbb E_{Y_t\sim\mathbf q_t^H}
[\xi_{t,Y_t}]
\ge 0.
\label{eq:nonnegative_history_value}
\end{equation}
Equality holds if and only if
$\mathbf q_t^H=\mathbf q_t^0$.
\end{proposition}
\vspace{-2mm}
Proposition~\ref{prop:nonnegative_history_value} establishes that target history is predictively useful in the oracle setting (proof in Appendix~\ref{app:Oracle_risk}). In practice, however, CTTA only has access to an estimate of $\mathbf q_t^H$ constructed from finite, unlabeled, and potentially stale observations. Thus, informative history does not guarantee that the resulting estimated correction is beneficial, motivating our source-relative gain formulation.

\begin{figure*}[!t]
    \centering
    \setlength{\abovecaptionskip}{2mm}
    \includegraphics[width=1\linewidth]{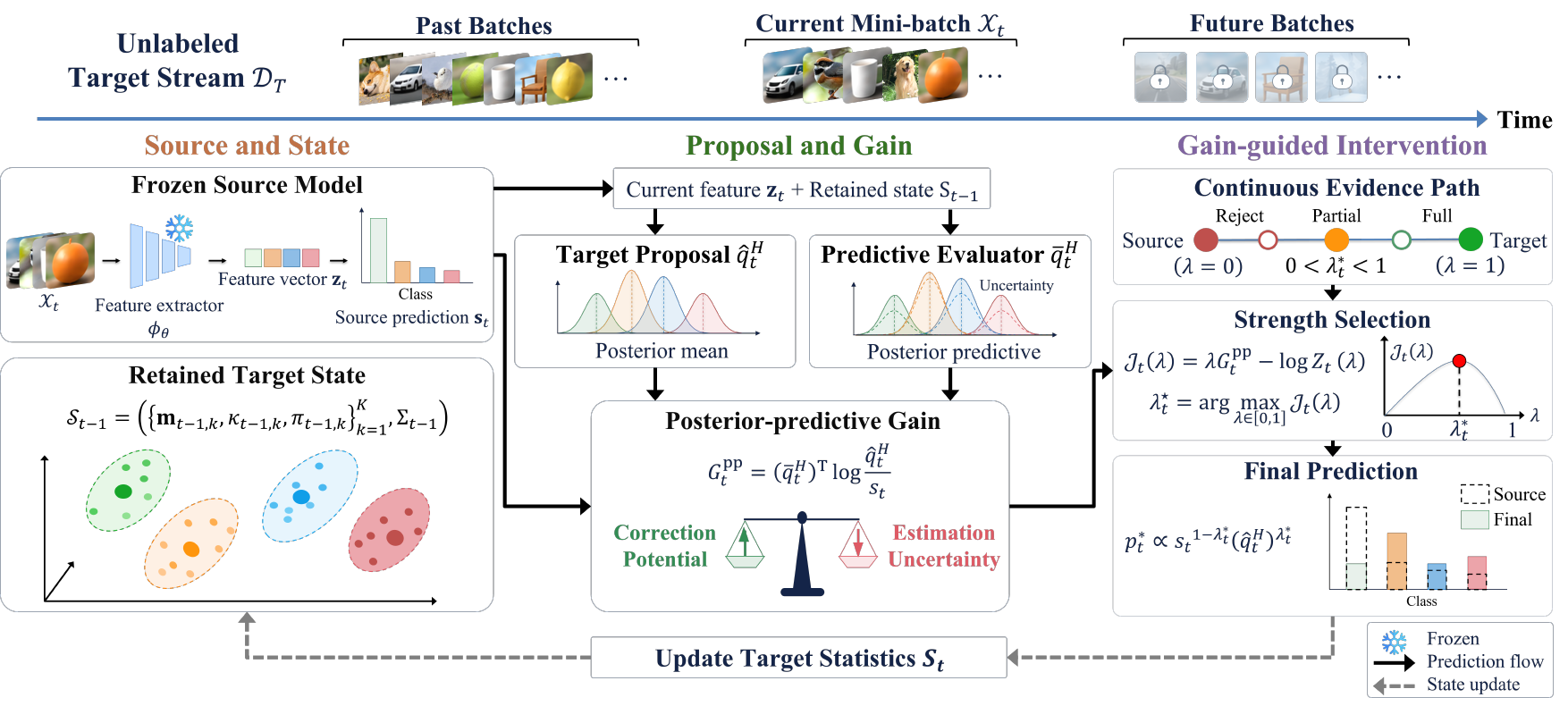}
\caption{\textbf{Overview of \ours: history proposes, gain decides.}
\textbf{(i)} The frozen source model produces the source prediction $\mathbf s_t$ and current feature $\mathbf z_t$, while the retained target state $\mathcal S_{t-1}$ summarizes past batches.
\textbf{(ii)} Together, $\mathbf z_t$ and $\mathcal S_{t-1}$ form the target proposal $\hat{\mathbf q}_t^H$ and predictive evaluator $\bar{\mathbf q}_t^H$. The evaluator accounts for uncertainty to estimate the proposal's source-relative gain $G_t^{\mathrm{pp}}$.
\textbf{(iii)} This source-relative gain determines the intervention strength $\lambda_t^\star$ along the continuous evidence path from $\mathbf s_t$ to $\hat{\mathbf q}_t^H$. The final prediction $\mathbf p_t^\star$ then updates the target statistics $\mathcal S_t$ for future batches.}
\label{fig:overview}
    \vspace{-1mm}
\end{figure*}

\section{\ours: Gain-Aware Intervention}
\label{sec:method}
\vspace{-2mm}
For a current sample $\mathbf{x}_t\!\in\!\mathcal{X}_t$, we denote its representation by $\mathbf{z}_t\!=\!\phi_\theta(\mathbf{x}_t)\in\mathbb{R}^{D}$.
Conceptually, let $\mathcal{Z}_t^H=\left\{\mathbf{z}_i^H\mid\mathbf{z}_i^H=\phi_\theta(\mathbf{x}_i^H), \mathbf{x}_i^H\in\mathcal{H}_t\right\}$ with $N_t^H=|\mathcal{H}_t|$ denote the representations associated with preceding target observations. We introduce $\mathcal Z_t^H$ only for notational convenience. In practice, we do not store or replay these historical representations, but maintain their aggregate effect through recursive sufficient statistics.

\subsection{Reliable Target-Side Gain Estimation}
\label{sec:gain_estimation}

\vspace{-1mm}
\paragraph{From Historical Evidence to Correction Gain.}
\label{sec:correction_gain}
\vspace{-1mm}
Preliminaries establish that accumulated target context has non-negative predictive value in the oracle setting. At test time, however, the history-conditioned posterior $\mathbf q_t^H$ is unavailable and must be approximated by an estimate $\hat{\mathbf q}_t^H$. If the source prediction $\mathbf s_t$ is fully replaced by this target estimate, the resulting conditional log-risk reduction is
\vspace{-1mm}
\begin{equation}
\Delta_t^{\mathrm{full}}
\triangleq
\mathcal{R}_t(\mathbf{s}_t)
-
\mathcal{R}_t(\hat{\mathbf{q}}_t^{H})
=
D_{\mathrm{KL}}
\big(
\mathbf{q}_t^{H}\Vert\mathbf{s}_t
\big)
-
D_{\mathrm{KL}}
\big(
\mathbf{q}_t^{H}\Vert\hat{\mathbf{q}}_t^{H}
\big).
\label{eq:practical_history_reliability}
\end{equation}

\vspace{-3mm}
Eq.~\ref{eq:practical_history_reliability} shows that informative target history does not necessarily yield a beneficial correction, since estimation error can offset its correction potential. This motivates two practical requirements: constructing a target-side correction proposal from accumulated history and evaluating whether that proposal improves upon retaining the source prediction.


\vspace{-2mm}
\paragraph{Source-Anchored Probabilistic Target Estimation.}
As illustrated in Fig.~\ref{fig:overview}, we address both requirements through a probabilistic model of accumulated target evidence. Its posterior mean defines a class-specific target correction proposal, while the posterior-predictive distribution accounts for estimation uncertainty when evaluating the source-relative utility of that proposal.
Specifically, for each class $k$, we model target features with a class-conditional Gaussian distribution and place a source-centered prior on its unknown class center $\boldsymbol{\mu}_k$ to stabilize estimation when target evidence is limited or noisy:
\vspace{-2mm}
\begin{equation}
p_T(\mathbf z\mid Y=k,\boldsymbol\mu_k)
=
\mathcal N(\mathbf z;\boldsymbol\mu_k,\Sigma_{t-1}),
\qquad
\boldsymbol\mu_k
\sim
\mathcal N
(
\mathbf c_k,
\frac{\Sigma_{t-1}}{\kappa_0}
),
\label{eq:target_gaussian_model}
\end{equation}
where $\mathbf c_k$ is the source-derived class prototype (\ie the k-th classifier-head weight) and $\kappa_0$ controls the strength of the source-centered prior.
We initialize $\Sigma_0=\mathbf I_D$ and update the shared diagonal covariance causally from preceding target observations.

Using the recursively maintained sufficient statistics induced by soft class assignments of preceding target observations, together with the pre-$t$ covariance estimate, we obtain the following fractional Gaussian posterior over the class center:
\vspace{-2mm}
\begin{equation}
\boldsymbol\mu_k\mid\mathcal H_t
\approx
\mathcal N
(
\mathbf m_{t-1,k},
\frac{\Sigma_{t-1}}{\kappa_{t-1,k}}
),
\qquad
\kappa_{t-1,k}
=
\kappa_0+n_{t-1,k},
\label{eq:center_posterior}
\end{equation}
where $n_{t-1,k}$ and $\mathbf m_{t-1,k}$ denote the retained target support and the source-anchored center.
The complete fractional-posterior derivation and recursive updates are given in Appendices~\ref{app:source_anchored_estimation} and~\ref{app:continual_statistics}.

Using the posterior mean geometry, we define
\vspace{-2mm}
\begin{equation}
d_{t,k}
=
(\mathbf z_t-\mathbf m_{t-1,k})^\top
\Sigma_{t-1}^{-1}
(\mathbf z_t-\mathbf m_{t-1,k}),
\qquad
\hat\ell_{t,k}^{H}
=
\log\pi_{t-1,k}
-\frac12d_{t,k},
\label{eq:target_score}
\end{equation}
and obtain the posterior-mean target proposal:
\vspace{-2mm}
\begin{equation}
\hat q_{t,k}^{H}
=
\frac{
\pi_{t-1,k}\exp(-d_{t,k}/2)
}{
\sum_j
\pi_{t-1,j}\exp(-d_{t,j}/2)
}.
\label{eq:target_posterior}
\end{equation}
The distribution $\hat{\mathbf q}_t^H\!=\![\hat q_{t,1}^H,\ldots,\hat q_{t,K}^H]^\top$ specifies the target-side correction proposed by the posterior-mean geometry, where $\pi_{t-1,k}$ is the pre-$t$ target class prior.
When $\mathcal H_t\!=\!\varnothing$, we set $\hat{\mathbf q}_t^H\!=\!\mathbf s_t$.

\vspace{-2mm}
\paragraph{Posterior-Predictive Gain Evaluation.}
The proposal $\hat{\mathbf q}_t^H$ is constructed from the posterior-mean target geometry and therefore does not account for the remaining uncertainty in the estimated class centers. To incorporate this uncertainty, we analytically marginalize the latent class centers.
From Eq.~\ref{eq:center_posterior} and the class-conditional observation model in Eq.~\ref{eq:target_gaussian_model}, the posterior-predictive likelihood for class $k$ is $p_T^{\mathrm{pp}}(\mathbf z_t\!\!\mid\!\! Y_t\!=\!k,\mathcal H_t)\!=\!\mathcal N\left(\mathbf z_t;\mathbf m_{t-1,k},h_{t,k}\Sigma_{t-1}\right)$, $h_{t,k}\!\!\triangleq\!\!1\!\!+\!\!\kappa_{t-1,k}^{-1}$.
Thus, classes with less precisely estimated centers induce broader posterior-predictive distributions. Normalizing these predictive likelihoods gives the posterior-predictive evaluator
\vspace{-2mm}
\begin{equation}
\bar q_{t,k}^{H}
\propto
\pi_{t-1,k}
h_{t,k}^{-D/2}
\exp
(
-\frac{d_{t,k}}{2h_{t,k}}
),
\qquad
h_{t,k}
=
1+\kappa_{t-1,k}^{-1}.
\label{eq:posterior_predictive_target}
\end{equation}

\vspace{-3mm}
Here, $\hat{\mathbf q}_t^H$ and $\bar{\mathbf q}_t^H$ serve distinct roles: the former specifies the correction proposed by the estimated target geometry, whereas the latter evaluates that correction after accounting for class-center uncertainty under the posterior-predictive working model.
Define the source-relative correction evidence $\hat\xi_{t,k}^{s}\!\triangleq\! \log \frac{\hat q_{t,k}^{H}}{s_{t,k}}$.
Then the oracle gain in Eq.~\ref{eq:practical_history_reliability} can be written as $\Delta_t^{\mathrm{full}}\!=\!(\mathbf q_t^H)^\top\hat{\boldsymbol\xi}_t^s$.
Since $\mathbf q_t^H$ is unavailable at test time, we evaluate the same correction under the posterior-predictive distribution:
%
\begin{equation}
\begin{aligned}
G_t^{\mathrm{pp}}
&\triangleq
(\bar{\mathbf q}_t^H)^\top
\hat{\boldsymbol\xi}_t^{s}
=
D_{\mathrm{KL}}
\left(
\bar{\mathbf q}_t^H
\Vert
\mathbf s_t
\right)
-
D_{\mathrm{KL}}
\left(
\bar{\mathbf q}_t^H
\Vert
\hat{\mathbf q}_t^H
\right).
\end{aligned}
\label{eq:posterior_predictive_gain}
\end{equation}
The first KL term captures the potential benefit of correcting the source prediction, while the second KL measures the mismatch between the proposed correction and its posterior-predictive evaluation. Thus, source--target disagreement alone does not justify correction; the proposed correction must also remain supported after accounting for uncertainty in the estimated target geometry.
Accordingly, $G_t^{\mathrm{pp}}$ is the expected gain of the proposed
correction under the posterior-predictive working model, rather than a
lower bound on the unknown oracle gain $\Delta_t^{\mathrm{full}}$.
See Appendix~\ref{app:posterior_predictive_gain} for details.

\subsection{Posterior-Predictive Evidence Intervention}
\label{sec:evidence_intervention}
\vspace{-2mm}
The posterior-predictive gain in Eq.~\ref{eq:posterior_predictive_gain} quantifies how strongly
the proposed correction is supported after accounting for uncertainty in the target geometry. We now translate this quantity into the extent of intervention on the frozen source prediction.

\vspace{-3mm}
\paragraph{Continuous Evidence Path.}
Rather than directly replacing $\mathbf s_t$ with the target estimate, we continuously scale the source-relative correction evidence $\hat{\boldsymbol\xi}_t^s$ by an intervention coefficient
$\lambda\in[0,1]$:
\vspace{-2mm}
\begin{equation}
p_{t,k}^{(\lambda)}
=
\frac{
s_{t,k}\exp(\lambda\hat\xi_{t,k}^{s})
}{
Z_t(\lambda)
}
=
\frac{
s_{t,k}^{1-\lambda}
(\hat q_{t,k}^{H})^\lambda
}{
\sum_{j=1}^{K}
s_{t,j}^{1-\lambda}
(\hat q_{t,j}^{H})^\lambda
},
\label{eq:evidence_path}
\end{equation}
where
$Z_t(\lambda)\!\!=\!\!\sum_j s_{t,j}\!\exp(\lambda\hat\xi_{t,j}^{s})$.
The two endpoints satisfy
$\mathbf p_t^{(0)}\!\!\!=\!\!\mathbf s_t$ and
$\mathbf p_t^{(1)}\!\!\!=\!\!\hat{\mathbf q}_t^H$.
Thus, $\lambda$ controls the amount of target-side correction introduced relative to the source prediction.
Importantly, for the current prediction, the posterior-predictive distribution $\bar q_t^H$ serves as an evaluator rather than as an additional replacement prediction.
Define its conditional logarithmic risk as
$\bar{\mathcal R}_t(\mathbf p)
\!\!\triangleq\!\!
-\!\sum_{k}\!
\bar q_{t,k}^{H}\log p_k$.
Then, the reduction in posterior-predictive risk relative to the source prediction is
\vspace{-1mm}
\begin{equation}
\begin{aligned}
\mathcal J_t(\lambda)
&\triangleq
\bar{\mathcal R}_t(\mathbf s_t)
-
\bar{\mathcal R}_t(\mathbf p_t^{(\lambda)})
\\
&=
\lambda
(\bar{\mathbf q}_t^H)^\top
\hat{\boldsymbol\xi}_t^s
-
\log Z_t(\lambda)
\\
&=
\lambda G_t^{\mathrm{pp}}
-
\log Z_t(\lambda),
\qquad
\lambda\in[0,1].
\end{aligned}
\label{eq:predictive_objective}
\end{equation}
Hence, intervention is determined by the gain of the same target correction after accounting for uncertainty in the estimated target geometry, while the normalization term follows exactly from the source-relative evidence path.

\begin{theorem}[Globally Optimal Posterior-Predictive Intervention]
\label{thm:risk_calibrated_intervention}
For the evidence path $\mathbf{p}_t^{(\lambda)}$ in Eq.~\ref{eq:evidence_path}, the posterior-predictive objective $\mathcal{J}_t(\lambda)$ is concave over $\lambda\in[0,1]$.
Consequently, it admits a globally optimal intervention coefficient $\lambda_t^\star$, yielding the final adapted prediction
\vspace{-1mm}
\begin{equation}
p_{t,k}^{\star}
=
p_{t,k}^{(\lambda_t^\star)}
\propto
s_{t,k}^{\,1-\lambda_t^\star}
\big(\hat q_{t,k}^{H}\big)^{\lambda_t^\star},
\qquad
\lambda_t^\star
=
\arg\max_{\lambda\in[0,1]}
\mathcal{J}_t(\lambda).
\label{eq:risk_calibrated_intervention}
\end{equation}
\end{theorem}
\vspace{-3mm}
The proof and the efficient one-dimensional solution for
$\lambda_t^\star$ are provided in Appendix~\ref{app:intervention}.

\vspace{-2mm}
\paragraph{Continual Target Update.}
In CTTA, historical statistics can become mismatched with the current distribution, while unreliable corrections may accumulate through subsequent state updates. \ours mitigates this propagation by updating the target state only from the gain-controlled predictions. Specifically, we define the reliability-weighted assignment
$\omega_{t,b,k}=\zeta_{t,b}p^\star_{t,b,k}$, where
$\zeta_{t,b}=s_{t,b,\hat y^\star_{t,b}}$
measures frozen-source support for the adapted prediction $\hat y^\star_{t,b}\!=\!\arg\max_k p^\star_{t,b,k} $.
The reliability-weighted class support $\kappa_{t,k}$ and predictive class mass $\hat\kappa_{t,k}$ are then accumulated as:
\vspace{-2mm}
\begin{equation}
\kappa_{t,k}
=\kappa_{t-1,k}
+
\sum\nolimits_{b=1}^{B_t}\omega_{t,b,k},
\qquad
\hat\kappa_{t,k}=\hat\kappa_{t-1,k}
+
\sum\nolimits_{b=1}^{B_t}p^\star_{t,b,k},
\label{eq:class_support_main}
\end{equation}

\vspace{-2mm}
with $\kappa_{0,k}=\hat\kappa_{0,k}=\kappa_0$. While $\hat\kappa_{t,k}$ reflects how frequently class $k$ is predicted, $\kappa_{t,k}$ measures how strongly these assignments are supported.
To further prevent frequently predicted classes from being progressively reinforced, we define the historical class prior by combining reliability-normalized support with inverse-support balancing:
\vspace{-2mm}
\begin{equation}
\pi_{t,k}
\propto
\frac{\kappa_{t,k}}{\hat{\kappa}_{t,k}}\cdot
\frac{1}{\hat{\kappa}_{t,k}}
=
\frac{\kappa_{t,k}}{\hat{\kappa}_{t,k}^{2}}.
\label{eq:historical_prior_main}
\end{equation}

\vspace{-4mm}
The same reliability-weighted evidence recursively updates the class centers and shared covariance, yielding
$\mathcal S_t\!=\!\left(\{\mathbf m_{t,k},\kappa_{t,k},\pi_{t,k}\}_{k=1}^{K},\Sigma_t\right)$.
The updated state is used only from time $t\!+\!1$ onward, limiting the repeated reinforcement of unreliable history-induced corrections without storing or replaying previous target samples. Full recursive updates are provided in Appendix~\ref{app:continual_statistics}.

\vspace{-1mm}
\section{Experiments}
\label{sec:exp}
\vspace{-2mm}
\subsection{Experimental Setup}
\vspace{-1mm}
\noindent\textbf{Datasets and Metrics.}
We evaluate \ours on ImageNet-C~\citep{imagenet-c}, ImageNet-3DCC~\citep{3DCC}, ImageNet-R~\citep{imagenet-r}, ImageNet-V2~\citep{imagenet-v}, and ImageNet-Sketch~\citep{imagenet-sketch} to assess robustness under diverse distribution shifts.
For ImageNet-C and ImageNet-3DCC, we use corruption severity 5 unless otherwise specified and perform continual adaptation without reset across the stream.
We consider four complementary stream settings: continual structured change (CSC)~\citep{CoTTA}, continual dynamic change (CDC)~\citep{DPCore}, mixed-domain shift (MDS)~\citep{SAR, ReCAP}, and long-horizon adaptation (LHA)~\citep{ViDA} over 10 repeated corruption cycles. We report top-1 accuracy (Acc.) and expected calibration error (ECE)~\citep{ECE}.

\begin{table}[!t]
\vspace{-3mm}
\caption{
\textbf{CSC results on ImageNet-C.} Accuracy (Acc., \%) and expected calibration error (ECE, \%) with ViT-Base at severity level~5. BP-free denotes backpropagation-free adaptation.
Bold indicates the best results; Source is shown for reference only.
}
\label{tab:imagenetc-csc}
\vspace{-3mm}
\begin{center}
\begingroup
\small
\setlength{\tabcolsep}{1.5pt}
\renewcommand{\arraystretch}{1.0}

\resizebox{\linewidth}{!}{%
\begin{tabular}{lcc|*{15}{c}|c}
\toprule
\multirow{2}{*}{Method} &
\multirow{2}{*}{\shortstack{BP-free}} &
\multirow{2}{*}{Metric} &
\multicolumn{3}{c}{Noise} &
\multicolumn{4}{c}{Blur} &
\multicolumn{4}{c}{Weather} &
\multicolumn{4}{c}{Digital} &
\multicolumn{1}{|c}{\multirow{2}{*}{\textbf{Avg.}}} \\

\cmidrule(lr){4-6}
\cmidrule(lr){7-10}
\cmidrule(lr){11-14}
\cmidrule(lr){15-18}

& & &
\multicolumn{1}{c}{Gauss.} &
\multicolumn{1}{c}{Shot} &
\multicolumn{1}{c}{Impu.} &
\multicolumn{1}{c}{Defo.} &
\multicolumn{1}{c}{Glas.} &
\multicolumn{1}{c}{Moti.} &
\multicolumn{1}{c}{Zoom} &
\multicolumn{1}{c}{Snow} &
\multicolumn{1}{c}{Fros.} &
\multicolumn{1}{c}{Fog} &
\multicolumn{1}{c}{Brig.} &
\multicolumn{1}{c}{Cont.} &
\multicolumn{1}{c}{Elas.} &
\multicolumn{1}{c}{Pix.} &
\multicolumn{1}{c}{JPEG} &
\multicolumn{1}{|c}{} \\
\midrule

\rowcolor{sourcecolor}
& & Acc. \(\uparrow\)
& 47.0 & 48.2 & 47.9 & 31.5 & 21.2 & 41.5 & 36.7
& 50.1 & 45.8 & 42.3 & 73.6 & 8.6 & 42.5 & 62.0
& 63.8 & 44.2 \\
\rowcolor{sourcecolor}
\multirow{-2}{*}{Source}
& \multirow{-2}{*}{--}
& ECE \(\downarrow\)
& 3.6 & 4.1 & 3.7 & 4.3 & 5.4 & 3.9 & 9.1
& 2.3 & 4.9 & 17.4 & 3.1 & 3.9 & 9.0 & 3.3
& 2.7 & 5.4 \\
\midrule

\multirow{2}{*}{Tent~(\hyperlink{cite.Tent}{ICLR 2021})}
& \multirow{2}{*}{\xmark}
& Acc. \(\uparrow\)
& 47.8 & 51.1 & 50.8 & 34.2 & 27.0 & 45.5 & 41.6
& 56.0 & 52.3 & 49.7 & 76.1 & 27.2 & 44.3 & 65.6
& 66.1 & 49.0 \\
& & ECE \(\downarrow\)
& 5.8 & 8.3 & 10.5 & 12.1 & 18.0 & 13.9 & 18.2
& 12.4 & 14.6 & 13.3 & 7.0 & 19.1 & 20.2 & 10.3
& 8.7 & 12.8 \\
\addlinespace[1.8pt]

\multirow{2}{*}{CoTTA~(\hyperlink{cite.CoTTA}{CVPR 2022})}
& \multirow{2}{*}{\xmark}
& Acc. \(\uparrow\)
& 47.1 & 48.4 & 48.6 & 31.7 & 21.9 & 42.9 & 38.0
& 51.8 & 47.3 & 44.7 & 74.1 & 10.0 & 43.6 & 63.6
& 64.8 & 45.2 \\
& & ECE \(\downarrow\)
& \textbf{4.0} & 5.3 & 5.6 & \textbf{3.1} & 8.4 & 7.2 & 13.0
& 6.6 & 10.9 & 11.6 & 4.7 & \textbf{2.7} & 15.7 & 8.3
& 5.5 & 7.5 \\
\addlinespace[1.8pt]

\multirow{2}{*}{SAR~(\hyperlink{cite.SAR}{ICLR 2023})}
& \multirow{2}{*}{\xmark}
& Acc. \(\uparrow\)
& 54.2 & 54.1 & 52.3 & 47.7 & 36.3 & 53.8 & 49.1
& 59.7 & 57.6 & 58.2 & 75.6 & 46.6 & 46.4 & 61.6
& 63.4 & 54.4 \\
& & ECE \(\downarrow\)
& 4.8 & 5.7 & 7.6 & 4.9 & 12.9 & 9.6 & 13.5
& 9.8 & 10.2 & 9.7 & 5.1 & 14.8 & 14.0 & 7.0
& 6.5 & 9.1 \\
\addlinespace[1.8pt]

\multirow{2}{*}{ROID~(\hyperlink{cite.ROID}{WACV 2024})}
& \multirow{2}{*}{\xmark}
& Acc. \(\uparrow\)
& 56.3 & 62.3 & 60.7 & 51.0 & 50.6 & \textbf{59.2} & 54.8
& 63.9 & 62.2 & 64.0 & 78.7 & 50.7 & 58.8 & 69.5
& 70.0 & 60.8 \\
& & ECE \(\downarrow\)
& 56.4 & 61.9 & 60.5 & 51.3 & 51.3 & 58.9 & 54.6
& 63.4 & 62.3 & 64.5 & 78.3 & 49.1 & 58.8 & 69.9
& 69.5 & 60.7 \\
\addlinespace[1.8pt]

\multirow{2}{*}{ViDA~(\hyperlink{cite.ViDA}{ICLR 2024})}
& \multirow{2}{*}{\xmark}
& Acc. \(\uparrow\)
& 52.3 & 57.5 & 57.1 & 47.8 & 43.1 & 54.5 & 51.1
& 61.1 & 57.3 & 59.3 & 75.7 & 47.2 & 50.9 & 66.5
& 66.9 & 56.6 \\
& & ECE \(\downarrow\)
& 6.8 & 11.0 & 13.7 & 10.8 & 20.4 & 14.7 & 19.5
& 13.9 & 16.5 & 15.3 & 8.1 & 24.0 & 23.3 & 11.2
& 11.0 & 14.7 \\
\addlinespace[1.8pt]

\multirow{2}{*}{DeYO~(\hyperlink{cite.DeYO}{ICLR 2024})}
& \multirow{2}{*}{\xmark}
& Acc. \(\uparrow\)
& 52.8 & 60.6 & 59.8 & 45.3 & 46.2 & 57.5 & 50.9
& 60.7 & 60.0 & 60.2 & 77.2 & 51.1 & 55.1 & 67.7
& 69.5 & 58.3 \\
& & ECE \(\downarrow\)
& 5.7 & 6.3 & 8.3 & 6.9 & 13.3 & 10.1 & 15.5
& 11.0 & 11.1 & 10.7 & 5.5 & 14.6 & 13.2 & 8.3
& 7.8 & 9.9 \\
\addlinespace[1.8pt]

\multirow{2}{*}{AEA~(\hyperlink{cite.AEA}{ICLR 2025})}
& \multirow{2}{*}{\xmark}
& Acc. \(\uparrow\)
& 54.1 & 55.1 & 55.6 & \textbf{55.5} & \textbf{51.9} & 58.6 & 49.3
& 11.4 & 25.7 & 69.2 & 73.7 & \textbf{64.5} & 59.1 & 70.1
& 66.9 & 54.7 \\
& & ECE \(\downarrow\)
& 18.6 & 19.1 & 20.0 & 23.1 & 25.0 & 22.1 & 26.6
& 23.3 & 22.2 & 19.4 & 15.8 & 22.0 & 27.9 & 20.4
& 21.1 & 21.8 \\
\addlinespace[1.8pt]

\multirow{2}{*}{ReCAP~(\hyperlink{cite.ReCAP}{ICML 2025})}
& \multirow{2}{*}{\xmark}
& Acc. \(\uparrow\)
& 37.9 & 47.8 & 52.9 & 49.1 & 50.9 & 56.3 & 53.1
& 58.4 & 61.5 & 65.8 & 76.3 & 62.8 & 57.9 & 67.3
& 67.2 & 57.7 \\
& & ECE \(\downarrow\)
& 9.1 & 9.4 & 9.8 & 7.9 & 11.1 & 9.2 & 12.8
& 10.4 & 9.5 & 8.5 & 5.0 & 10.8 & 12.6 & 8.1
& 7.8 & 9.5 \\
\addlinespace[1.8pt]

\multirow{2}{*}{REM~(\hyperlink{cite.REM}{ICML 2025})}
& \multirow{2}{*}{\xmark}
& Acc. \(\uparrow\)
& 56.5 & 61.9 & 60.8 & 46.8 & 51.0 & 56.5 & \textbf{57.2}
& 62.5 & 64.8 & 64.6 & 76.8 & 53.2 & 58.4 & 71.1
& 69.8 & 60.8 \\
& & ECE \(\downarrow\)
& 5.5 & 6.0 & 7.2 & 10.4 & 13.1 & 10.8 & 11.9
& 8.6 & \textbf{7.0} & 8.6 & 5.2 & 11.0 & 11.5 & 6.3
& \textbf{4.9} & 8.5 \\
\addlinespace[1.8pt]

\multirow{2}{*}{DPCore~(\hyperlink{cite.DPCore}{ICML 2025})}
& \multirow{2}{*}{\xmark}
& Acc. \(\uparrow\)
& 57.8 & 61.3 & 60.7 & 52.8 & 48.6 & 52.3 & 53.1
& 60.7 & 63.1 & 62.6 & 78.0 & 55.6 & 54.9 & 69.1
& 70.4 & 60.1 \\
& & ECE \(\downarrow\)
& 7.4 & 8.1 & 8.0 & 6.4 & \textbf{4.7} & 7.2 & 6.6
& 10.9 & 9.6 & 8.3 & 9.7 & 7.7 & 8.4 & 10.8
& 10.0 & 8.2 \\
\addlinespace[1.8pt]

\multirow{2}{*}{PAID~(\hyperlink{cite.PAID}{NeurIPS 2025})}
& \multirow{2}{*}{\xmark}
& Acc. \(\uparrow\)
& 51.2 & 56.3 & 55.6 & 50.6 & 50.4 & 52.7 & 55.8
& 62.5 & 60.6 & 57.9 & 74.8 & 50.0 & \textbf{60.7} & 64.5
& 63.5 & 57.8 \\
& & ECE \(\downarrow\)
& 8.7 & 8.6 & 8.6 & 8.8 & 11.0 & 10.6 & 8.2
& 7.2 & 7.9 & 9.5 & \textbf{4.0} & 13.1 & 8.4 & 7.3
& 6.6 & 8.6 \\
\addlinespace[1.8pt]

\multirow{2}{*}{DOTA~(\hyperlink{cite.DOTA}{NeurIPS 2025})}
& \multirow{2}{*}{\cmark}
& Acc. \(\uparrow\)
& 57.2 & 57.6 & 59.1 & 48.9 & 38.0 & 55.2 & 47.4
& 63.6 & 65.1 & 68.5 & 78.5 & 33.1 & 47.5 & 68.5
& 69.9 & 57.2 \\
& & ECE \(\downarrow\)
& 35.5 & 36.8 & 36.3 & 44.1 & 55.5 & 40.5 & 47.9
& 33.7 & 31.7 & 27.6 & 20.2 & 57.8 & 48.5 & 29.3
& 28.0 & 38.2 \\
\addlinespace[1.8pt]

\multirow{2}{*}{FreqCTTA~(\hyperlink{cite.FreqCTTA}{AAAI 2026})}
& \multirow{2}{*}{\xmark}
& Acc. \(\uparrow\)
& 52.3 & 54.9 & 57.8 & 53.4	& 50.3 & 57.2 & 53.5 & 65.0 & 62.0 & 64.8 & 78.4 & 48.3 & 56.5 & \textbf{73.1} & 69.0 & 59.8\\
& & ECE \(\downarrow\)
& 4.6 & 7.0 & 7.8 & 7.5 & 9.7 & 9.0	& 13.1 & 8.9 & 9.5 & 11.0 & 5.5 & 14.7 & 14.2 & 7.8	& 8.1 & 9.2 \\
\addlinespace[1.8pt]

\multirow{2}{*}{NEO~(\hyperlink{cite.NEO}{ICLR 2026})}
& \multirow{2}{*}{\cmark}
& Acc. \(\uparrow\)
& 56.7 & 57.1 & 57.4 & 46.7 & 35.6 & 52.8 & 45.5
& 62.7 & 63.7 & 68.6 & 78.0 & 36.4 & 45.5 & 66.9
& 67.1 & 56.0 \\
& & ECE \(\downarrow\)
& 10.6 & 7.1 & 9.7 & 6.4 & 5.7 & \textbf{3.6} & \textbf{4.9}
& 5.0 & 20.7 & 51.0 & 8.9 & 23.9 & \textbf{5.9} & \textbf{5.8}
& 6.8 & 11.7 \\
\addlinespace[1.8pt]

\multirow{2}{*}{GOLD~(\hyperlink{cite.GOLD}{CVPR 2026})}
& \multirow{2}{*}{\xmark}
& Acc. \(\uparrow\)
& \textbf{59.9}	& \textbf{64.4}	& \textbf{64.3}	& 42.7	& 44.6	& 56.4	& 48.2	& 64.7	& 64.1	& 63.8	& 77.9	& 30.6	& 55.2	& 68.8	& 70.2	& 58.4   \\
& & ECE \(\downarrow\)
& 27.9	& 25.6	& 26.1	& 40.3	& 42.3	& 33.1	& 40.8	& 27.5	& 27.5	& 25.7	& 17.1	& 48.7	& 35.4	& 24.3	& 23.6	& 31.1  \\

\midrule
\rowcolor{mycolor}
& & Acc. \(\uparrow\)
& 57.7 & 59.3 & 60.0 & 53.2 & 44.1 & 58.3 & 53.2 & \textbf{65.6} & \textbf{67.0} & \textbf{72.6} & \textbf{78.7} & 61.9 & 55.6 & 69.7 & \textbf{71.3} & \textbf{61.9} \\
\rowcolor{mycolor}
\multirow{-2}{*}{\ours (Ours)}
& \multirow{-2}{*}{\cmark}
& ECE \(\downarrow\)
& 4.6 & \textbf{4.8} & \textbf{5.3} & 5.6 & 5.1 & 5.8 & 5.5 & \textbf{4.9} & 9.2 & \textbf{5.4} & 5.3 & 9.0 & 7.0 & 6.6 & 5.9 & \textbf{6.0} \\

\bottomrule
\end{tabular}%
}
\vspace{-5mm}
\endgroup
\end{center}
\end{table}

\vspace{-2mm}
\noindent{\textbf{Implementation Details.}}
All experiments use an ImageNet-pretrained ViT-B/16 with a batch size of 64 on a single NVIDIA RTX A6000 GPU. The source model remains frozen, while \ours updates only compact target statistics without backpropagation or parameter updates. We set $\kappa_0\!=\!3$ and use the same hyperparameters across CSC, CDC, MDS, and long-horizon evaluation. Unless otherwise specified, main-paper experiments are conducted on ImageNet-C. ECE is computed from the final outputs of each method's official implementation. Further details are provided in Appendix~\ref{app:setup}.

\vspace{-1mm}
\subsection{Main Results on ImageNet-C}
\vspace{-1mm}
\paragraph{CSC Scenario.}
Continual structured change (CSC) introduces abrupt domain transitions while carrying historical information across corruptions, making error accumulation a key challenge. As shown in Table~\ref{tab:imagenetc-csc}, \ours achieves 61.9\% accuracy with 6.0\% ECE, improving accuracy over the frozen source by 17.7 points with only a 0.6-point increase in ECE. In contrast, ROID reaches 60.8\% accuracy with 60.7\% ECE, while the BP-free DOTA obtains 57.2\% accuracy with 38.2\% ECE. These results show that \ours enables accurate and well-calibrated adaptation without backpropagation, supporting gain-guided source-relative intervention.

\begin{figure*}[t]
    \centering
    \vspace{-3mm}
    \includegraphics[width=\textwidth]{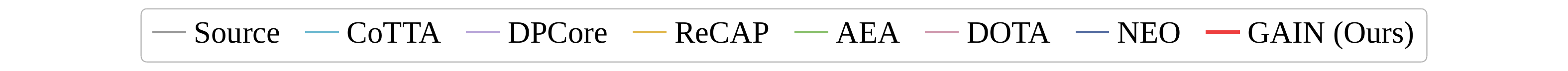}
    \par\vspace{-4mm}

    \makebox[\textwidth][c]{%
        \hspace{4mm} 

        \hspace{-6mm}
        \begin{subfigure}[t]{0.36\textwidth}
            \centering
            \vspace{0pt}
            \includegraphics[width=\linewidth]{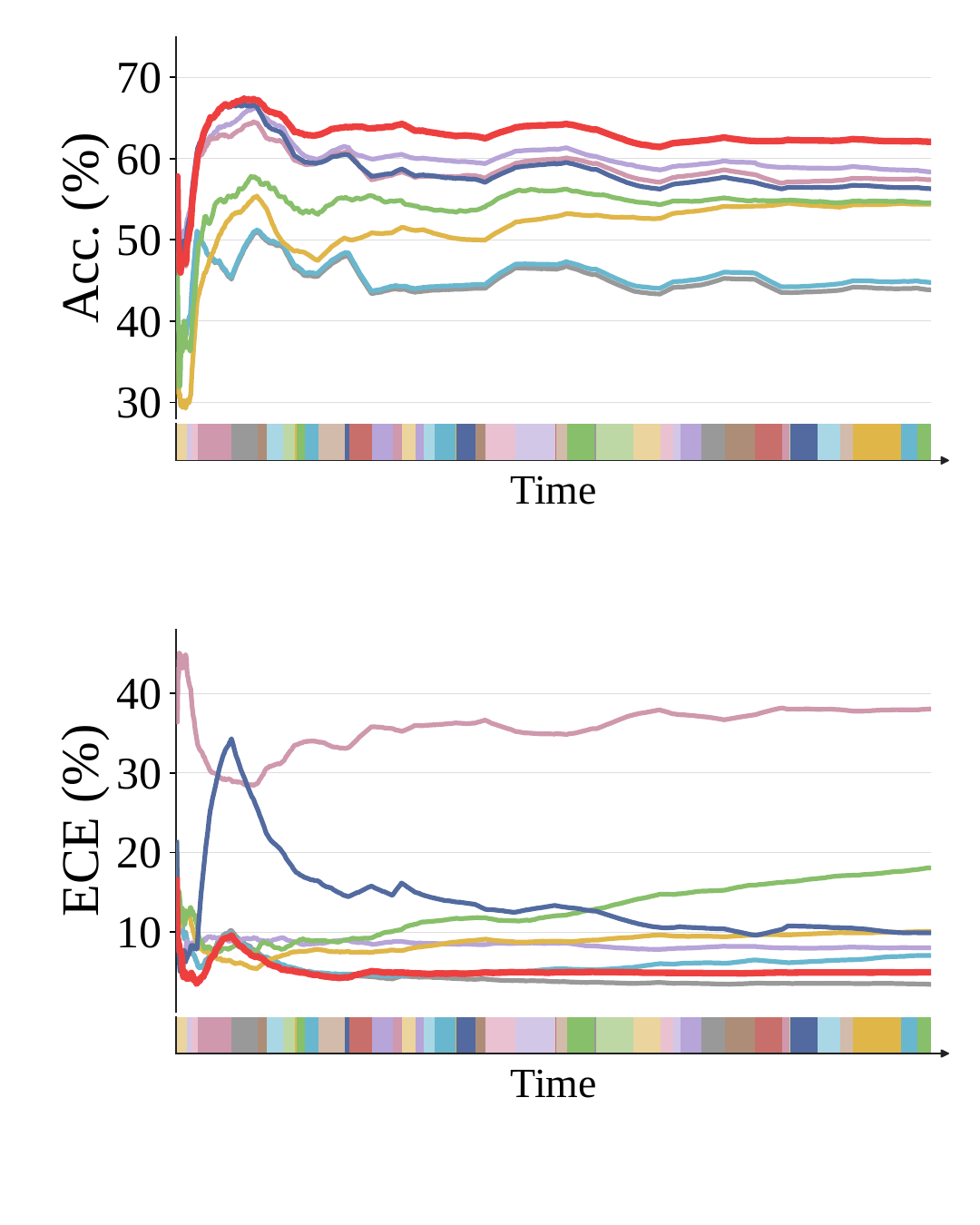}
            \setlength{\abovecaptionskip}{-6mm}
            \caption{CDC Scenario}
            \label{fig:cdc}
        \end{subfigure}\hspace{-6mm}
        \begin{subfigure}[t]{0.36\textwidth}
            \centering
            \vspace{0pt}
            \includegraphics[width=\linewidth]{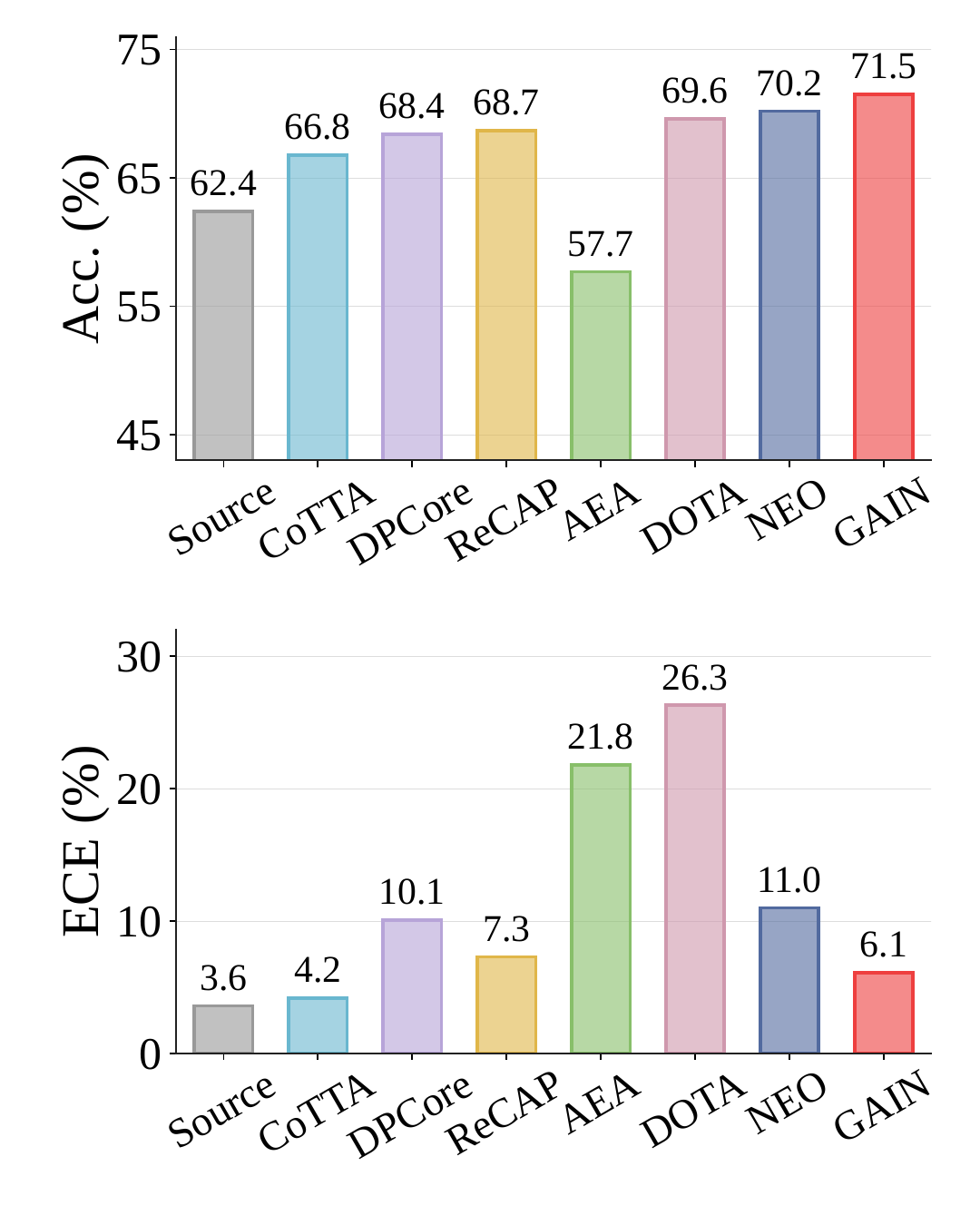}
            \setlength{\abovecaptionskip}{-6mm}
            \caption{MDS Scenario}
            \label{fig:mds}
        \end{subfigure}\hspace{-3.5mm}
        \begin{subfigure}[t]{0.36\textwidth}
            \centering
            \vspace{0pt}
            \includegraphics[width=\linewidth]{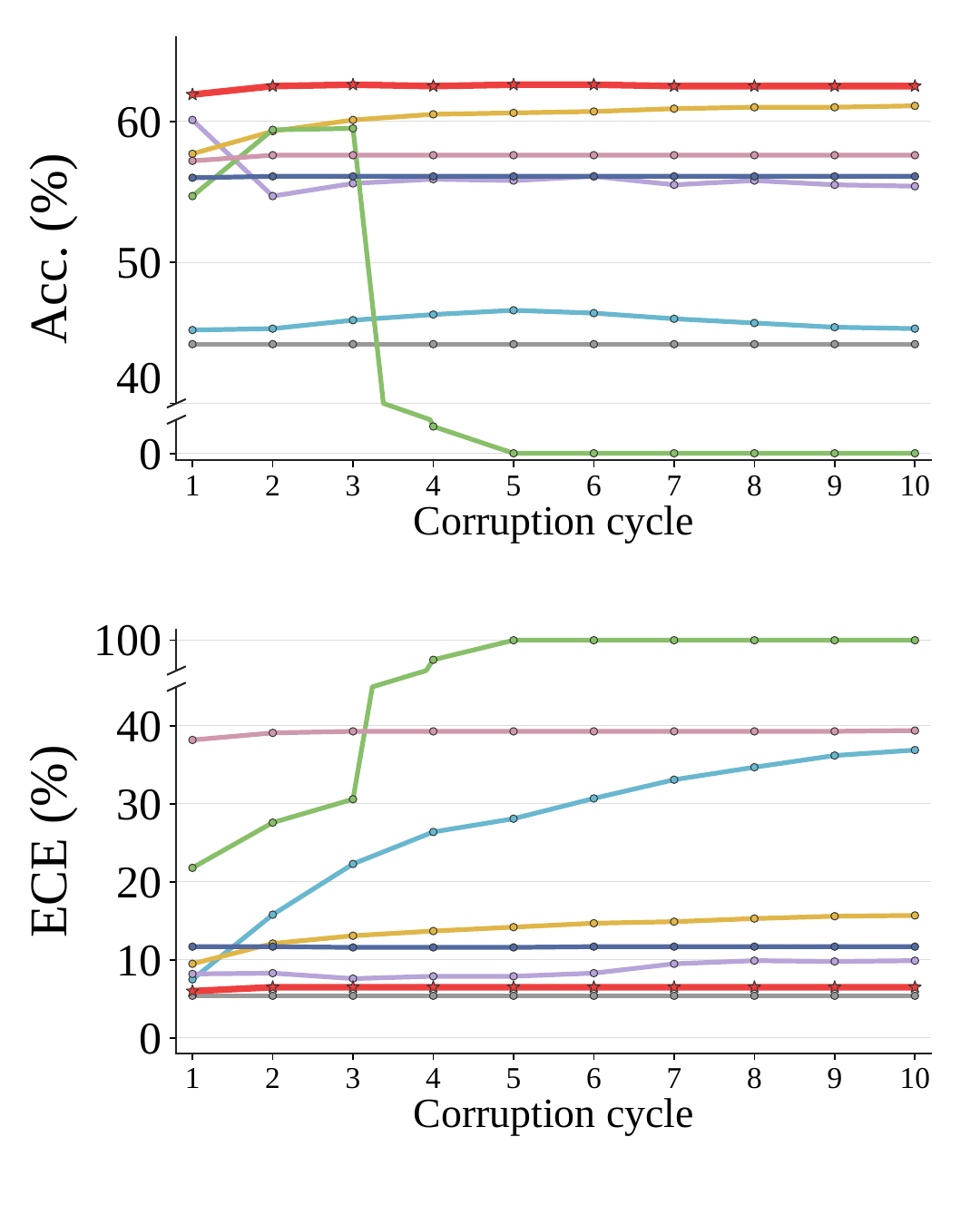}
            \setlength{\abovecaptionskip}{-6mm}
            \caption{LHA Scenario}
            \label{fig:lha}
        \end{subfigure}
    }

    \setlength{\abovecaptionskip}{1mm}
    \caption{
\textbf{Accuracy and calibration under three challenging CTTA scenarios on ImageNet-C.}
\textbf{(a)} CDC evaluates recurring corruptions with irregular durations.
\textbf{(b)} MDS interleaves samples from multiple corruption domains, with results averaged over severity levels 1–5.
\textbf{(c)} LHA evaluates error accumulation and long-term stability over 10 repeated corruption cycles.
}
    \label{fig:3scenario}
    \vspace{-3mm}
\end{figure*}

\vspace{-2mm}
\paragraph{CDC Scenario.}
Continual dynamic change (CDC) further challenges adaptation through recurring corruptions with irregular durations and frequencies, making historical target statistics less consistently aligned with the current distribution. As shown in Fig.~\ref{fig:cdc}, \ours maintains high accuracy and low calibration error throughout the dynamic stream, achieving 61.8\% mean accuracy and 6.2\% ECE. In contrast, several baselines exhibit either accuracy degradation or substantial miscalibration. These results show that gain-guided intervention remains reliable under irregular distribution changes by evaluating the source-relative utility of history-induced corrections.

    

\vspace{-2mm}
\paragraph{MDS Scenario.}
Mixed-domain shift interleaves samples from heterogeneous corruption domains, making historical target statistics less specific to the current sample. As shown in Fig.~\ref{fig:mds}, \ours achieves the highest accuracy averaged across severity levels of 71.5\% with a low ECE of 6.1\%, outperforming the BP-free NEO and DOTA in the accuracy--calibration trade-off. These results show that gain-guided intervention remains effective under heterogeneous target shifts.

\vspace{-3mm}
\paragraph{LHA Scenario.}
The long-horizon setting evaluates adaptation stability under repeated exposure, where small errors may accumulate over time. As shown in Fig.~\ref{fig:lha}, \ours remains stable across all 10 rounds: accuracy increases from 61.9\% in R1 to 62.5--62.6\% thereafter, while ECE stays around 6.5\%. In contrast, CoTTA suffers severe calibration drift, DPCore exhibits noticeable accuracy degradation, ReCAP becomes increasingly miscalibrated, and AEA eventually collapses. DOTA remains persistently miscalibrated, while NEO is stable but substantially less accurate. These results demonstrate that \ours maintains stable accuracy and calibration over long horizons by limiting the propagation of unreliable history-induced corrections.

\begin{wrapfigure}{r}{0.32\textwidth} 
    \vspace{-2.7em}
    \centering
    \includegraphics[width=0.95\linewidth]{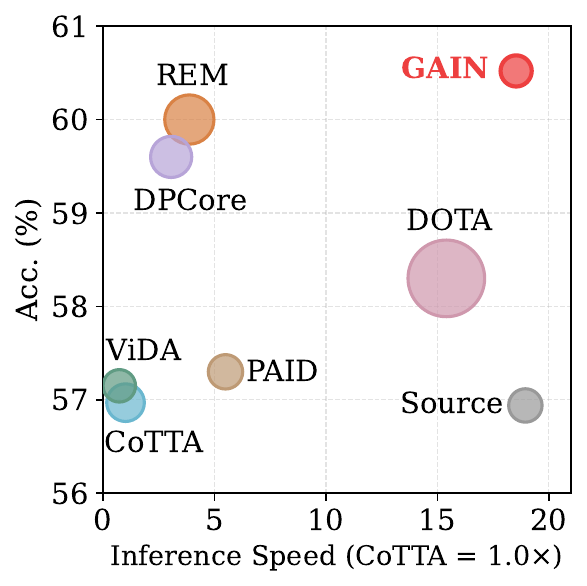}
        \vspace{-3mm}
        \caption{\textbf{Efficiency–accuracy trade-off on ImageNet-3DCC.} 
       {\small Inference speed is normalized to CoTTA; bubble size indicates ECE.}
        }
    \label{fig:3DCC}
    \vspace{-1.5em}
\end{wrapfigure}

\vspace{-2mm}
\subsection{Experiments on  ImageNet-3DCC}
\vspace{-2mm}
To evaluate robustness under more realistic shifts, we further consider ImageNet-3DCC~\citep{3DCC}, which covers diverse geometry- and imaging-related corruptions and provides a complementary test beyond conventional 2D corruptions. As shown in Fig.~\ref{fig:3DCC}, \ours achieves the highest classification accuracy while maintaining near-source inference speed, reaching $18.5\times$ the inference speed of CoTTA, with low calibration error under CSC. In contrast, REM and DPCore achieve competitive accuracy at substantially lower inference speeds, while DOTA remains less accurate and more poorly calibrated. These results show that gain-guided intervention preserves a strong accuracy--calibration--efficiency trade-off beyond ImageNet-C under more diverse corruption shifts.
Additional experiments and analyses are provided in Appendix~\ref{app:experiment}.


\vspace{-2mm}
\subsection{Ablation Studies and Further Analysis}
\vspace{-2mm}
\paragraph{Ablation Studies.}
Table~\ref{tab:ablation_main} compares different intervention rules under the same target-side estimation framework. Source retention does not exploit target evidence, whereas full correction substantially improves accuracy but leads to poor calibration. Fixed, entropy-based, and disagreement-based interventions partially alleviate this trade-off, but remain inferior to \ours. In contrast, \ours determines the intervention strength from the source-relative posterior-predictive gain, achieving the best overall performance with 61.9\% accuracy, 6.0\% ECE, and 1.9 NLL. These results reinforce our principle: history proposes, while gain decides whether and how strongly to intervene.

\vspace{-3mm}
\paragraph{Sensitivity to $\kappa_0$.}
We study the sensitivity of \ours to the prior strength $\kappa_0$, which controls the influence of the source prior on target-side estimation. As shown in Fig.~\ref{fig:sensitivity_kappa}, increasing $\kappa_0$ from $0.5$ to $2$ improves both accuracy and calibration, with accuracy rising from $61.6\%$ to $61.8\%$ and ECE decreasing from $7.9\%$ to $6.1\%$. Performance remains stable for $\kappa_0\in[2,4]$, with the highest accuracy of $61.9\%$ achieved at $\kappa_0\!=\!3$. Larger values slightly improve ECE but gradually reduce accuracy as the source prior becomes more dominant. We therefore set $\kappa_0\!=\!3$ by default, which offers a favorable accuracy--calibration trade-off without careful tuning.

\begin{table}[t]
\vspace{-3mm}
  \centering
  \begin{minipage}[t]{0.43\textwidth}
    \centering
    \setlength{\abovecaptionskip}{1mm}
    \caption{\textbf{Ablation of intervention rules under CSC.} {\small Signal denotes the criterion for setting the intervention strength $\lambda_t$.}}
     \label{tab:ablation_main}
\begingroup
\small
\setlength{\tabcolsep}{1.3pt}
\renewcommand{\arraystretch}{1.24}
\resizebox{\columnwidth}{!}{
\begin{tabular}{lcccccc}
\toprule
Intervention Rule
& Signal
& $\lambda_t$
& Acc. $\uparrow$
& ECE $\downarrow$
& NLL $\downarrow$ \\
\midrule


Source Retention
& -- 
& $0$
& 44.2 & 5.4  & 3.0 \\

Full Correction
& --
& $1$
& 58.6 & 11.2 & 2.5 \\

Fixed Intervention
& --
& $0.5$
& 59.7 & 8.9 & 2.1 \\

Entropy-based
& $H(\mathbf s_t)$
& $\lambda_t^{\mathrm{ent}}$
& 58.9 & 8.4 & 2.1 \\

Disagreement-based
& $\mathrm{JS}(\mathbf s_t,\hat{\mathbf q}_t^H)$
& $\lambda_t^{\mathrm{dis}}$
& 58.9 & 9.2 & 2.1 \\

\midrule
{\ours (Ours)}
& $G_t^{\mathrm{pp}}$
& $\lambda_t^\star$
& \textbf{61.9} & 6.0 & \textbf{1.9} \\
\bottomrule
\end{tabular}}
\endgroup








  \end{minipage}
  \hspace{0mm}
  \begin{minipage}[t]{0.48\textwidth}
    \centering
    \setlength{\abovecaptionskip}{1mm}
    \caption{
   \textbf{ Efficiency analysis on ImageNet-C.}
    {\small BP/FP: backward/forward propagation counts. 
    Speed is normalized to CoTTA (1.0\(\times\)); higher is faster.
    }
    }
    \label{tab:Comp}
\begingroup
\small
\renewcommand{\arraystretch}{0.92}
\setlength{\tabcolsep}{1.0pt}
\renewcommand{\arraystretch}{1.16}
\resizebox{\columnwidth}{!}{
\begin{tabular}{lccccccc}
\toprule
Method
& \#BP
& \#FP
& Param.(M) \(\downarrow\)
& Mem.(GB) \(\downarrow\)
& Speed $\uparrow$
& Acc. \(\uparrow\)
& ECE \(\downarrow\)
\\
\midrule


CoTTA
& 1 & 5.1 & 86.42 & 23.01 & 1.0\(\times\) & 45.2 & 7.5
\\

AEA
& 1 & 1 & 0.04 & 6.05 & 6.3\(\times\) & 54.7 & 21.8
\\

REM
& 1 & 3 & 0.03 & 26.77 & 3.3\(\times\) & 60.8 & 8.5
\\

DPCore
& 7.9 & 9.9 & 1.03 & 8.59 & 0.6\(\times\) & 60.1 & 8.2
\\

PAID
& 1 & 1 & 0.81 & 11.63 & 4.6\(\times\) & 57.8 & 8.6
\\

DOTA
& 0 & 1 & 0 & 9.54 & 12.7\(\times\) & 57.2 & 38.2
\\

\midrule
\ours (Ours)
& \textbf{0} & \textbf{1} & \textbf{0} & \textbf{0.81} & \textbf{15.9\(\times\)} & \textbf{61.9} & \textbf{6.0}
\\
\bottomrule
\end{tabular}}
\endgroup









  \end{minipage}
  \vspace{-2mm}
\end{table}

\begin{figure*}[t]
    \centering
    \setlength{\abovecaptionskip}{1mm}
    \begin{subfigure}[t]{0.49\textwidth}
        \centering
        \includegraphics[width=0.498\linewidth]{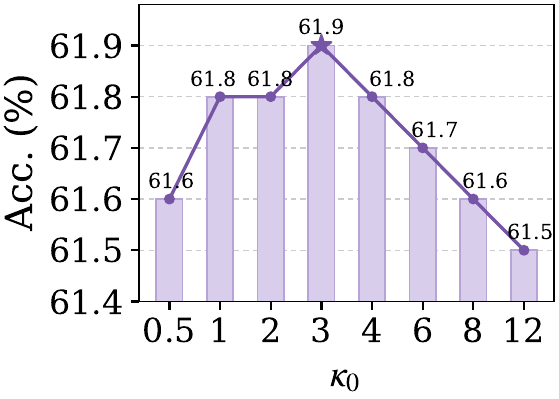}
        \hspace{-1mm}
        \includegraphics[width=0.478\linewidth]{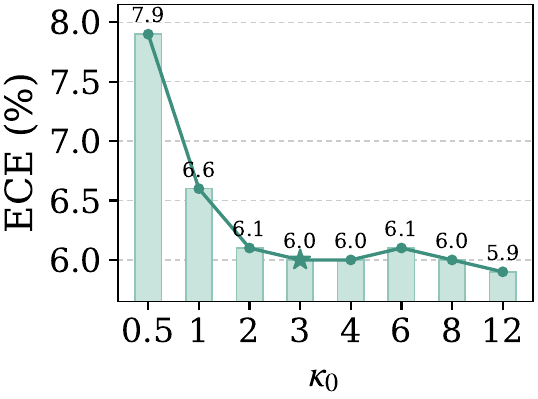}
        
        \vspace{-2mm}
        \caption{Source-centered Prior Strength $\kappa_0$.}
        \label{fig:sensitivity_kappa}
    \end{subfigure}
    \hspace{0mm}
    \begin{subfigure}[t]{0.48\textwidth}
        \centering
        \includegraphics[width=0.48\linewidth]{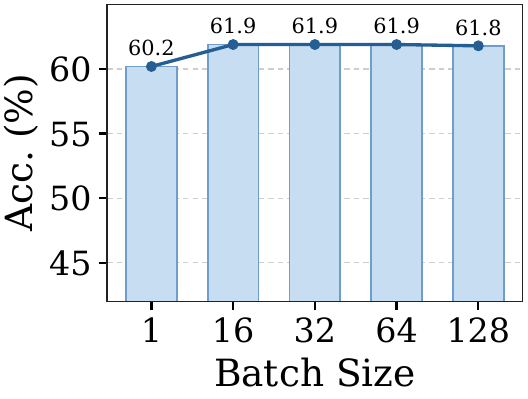}
        \hspace{-1mm}
        \includegraphics[width=0.48\linewidth]{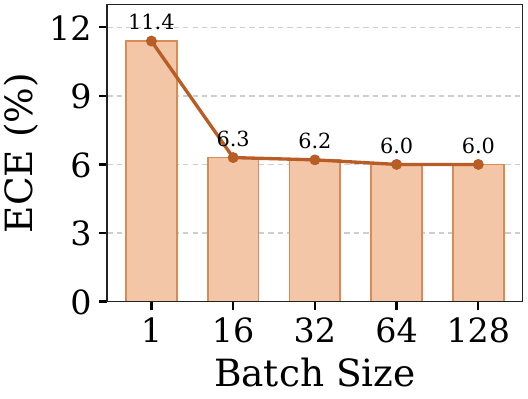}

        \vspace{-2mm}
        \caption{Test-Time Batch Size}
        \label{fig:sensitivity_batch}
    \end{subfigure}

    \caption{
    \textbf{Sensitivity analysis.}
       We study the sensitivity of \ours to (a) the source-centered prior strength $\kappa_0$ and (b) the test-time batch size. For each setting, we report both accuracy and ECE.
    }
    \label{fig:hyperparameter_sensitivity}
    \vspace{-2mm}
\end{figure*}

\vspace{-2mm}
\paragraph{Effect of Test-Time Batch Size.}
We evaluate the effect of test-time batch size on adaptation performance. As shown in Fig.~\ref{fig:sensitivity_batch}, single-sample updates yield less reliable target statistics, with $60.2\%$ accuracy and $11.4\%$ ECE. Increasing the batch size to $16$ improves accuracy to $61.9\%$ and sharply reduces ECE to $6.3\%$. Beyond $16$ samples, performance largely saturates: accuracy remains within $61.8$--$61.9\%$, while ECE only gradually decreases to $6.0\%$ at a batch size of $128$. This indicates that \ours does not require large test-time batches. For a fair comparison, we use a default test-time batch size of $64$ in all main experiments.

\vspace{-2mm}
\paragraph{Computational Efficiency.}
Table~\ref{tab:Comp} compares the computational efficiency of different CTTA methods on ImageNet-C. \ours requires only a single forward pass, without backpropagation or trainable parameter updates, and uses only $0.81$ GB of memory. With computational speed normalized to CoTTA ($1.0\times$), \ours achieves the highest relative speed of $15.9\times$, while also attaining the best accuracy of $61.9\%$ with only $6.0\%$ ECE. Compared with the BP-free DOTA, \ours is also faster while improving accuracy by $4.7$ points and reducing ECE from $ 38.2\%$ to $ 6.0\%$. These results demonstrate that \ours achieves a favorable accuracy--calibration--efficiency trade-off with a lightweight, forward-only adaptation pipeline.
\section{Conclusion}
\vspace{-2mm}
We presented \ours, a gain-guided framework for continual test-time adaptation following the principle that \emph{history proposes, gain decides}. Rather than directly trusting history-derived corrections, \ours evaluates their source-relative utility with a posterior-predictive evaluator and adaptively controls the intervention strength along a continuous evidence path.
 Combined with causal target-statistic updates, \ours limits the propagation of unreliable corrections while enabling efficient adaptation without backpropagation or replay. Across structured, dynamic, mixed-domain, and long-horizon shifts, \ours achieves strong accuracy, calibration, and stability, highlighting the effectiveness of gain-guided intervention for reliable continual adaptation.


\vspace{-3mm}
\paragraph{Limitations and future work.}
Our study follows the standard closed-set CTTA setting, where the source and target domains share the same label space. Extending gain-guided intervention to open-set adaptation and broader prediction tasks presents a promising direction for future work.

\bibliography{iclr2027_conference}
\bibliographystyle{iclr2027_conference}

\clearpage
\appendix
\appendix
\noindent{\Large \textbf{Appendix}} \\

\noindent
This Appendix provides additional theoretical derivations, implementation
details, and experimental results supporting our method.
The contents are organized as follows:
\vspace{-0.3em}
\begin{itemize}
    \item {Appendix~\ref{app:source_reliability}}:
    Why source-only reliability is insufficient under continual target shift;
    \item {Appendix~\ref{app:historical_evidence}}:
    Accumulated target context as additional predictive evidence;
    \item {Appendix~\ref{app:target_estimator}}:
    Posterior-predictive target-side gain estimation;
    \item {Appendix~\ref{app:intervention}}:
    Posterior-predictive evidence intervention;
    \item {Appendix~\ref{app:statistics_algori}}:
    Causal continual target-statistic updates and algorithmic implementation;
    \item {Appendix~\ref{app:setup}}:
    Experimental setup and implementation details;
    \item {Appendix~\ref{app:experiment}}:
    Additional experimental results and analyses.
\end{itemize}

\section{Source-Only Reliability Is Insufficient}
\label{app:source_reliability}

The main paper argues that source confidence alone is generally insufficient for deciding whether the current source prediction should be modified. We formalize this observation below.

Let
\begin{equation}
\hat y_t^s=\arg\max_k s_{t,k},
\qquad
C_t^s=\mathbb{I}[Y_t=\hat y_t^s]
\end{equation}
denote the source prediction and its correctness indicator.

\begin{proposition}[Insufficiency of source-only reliability]
\label{prop:source_only_appendix}
Suppose there exist a source predictive distribution $\mathbf{s}$ and two target histories $h$ and $h'$ with positive probability such that
\begin{equation}
P_T(C_t^s=1
\mid
\mathbf{s}_t=\mathbf{s},\mathcal{H}_t=h)
\neq
P_T(C_t^s=1
\mid
\mathbf{s}_t=\mathbf{s},\mathcal{H}_t=h').
\label{eq:source_reliability_condition}
\end{equation}
Then no function depending only on $\mathbf{s}_t$ can recover the conditional correctness probability under both histories.
\end{proposition}

\paragraph{Proof.}
Assume that there exists a function $g$ such that  $g(\mathbf{s}_t) = P_T(C_t^s=1\mid\mathbf{s}_t,\mathcal{H}_t)$ for every admissible history.
For the same source distribution $\mathbf{s}$ in Eq.~\ref{eq:source_reliability_condition}, this would require simultaneously
\begin{align}
g(\mathbf{s})
&=
P_T(C_t^s=1
\mid
\mathbf{s}_t=\mathbf{s},\mathcal{H}_t=h)=a,
\\
g(\mathbf{s})
&=
P_T(C_t^s=1
\mid
\mathbf{s}_t=\mathbf{s},\mathcal{H}_t=h')=b
\end{align}
with $a \neq b$. This is a contradiction.
\hfill$\square$

This result does not imply that confidence is uninformative. Rather, it shows that confidence is not a sufficient statistic for adaptation reliability when the correctness of an identical source prediction depends on the evolving target context.


\section{Accumulated Target Context as Predictive Evidence}
\label{app:historical_evidence}

This section provides the derivations underlying the C-PMI interpretation in Eq.~\ref{eq:historical_evidence}, the non-negative oracle predictive value in Eq.~\ref{eq:nonnegative_history_value}, and the practical correction gain in Eq.~\ref{eq:practical_history_reliability}.

\subsection{Conditional Pointwise Mutual Information Representation}
\label{app:CPMI}

For notational clarity, consider a fixed realization
$X_t=\mathbf{x}_t$ and $\mathcal{H}_t=h$, and define
\begin{equation}
q_k^0
\triangleq
P_T(Y_t=k\mid X_t=\mathbf{x}_t),
\qquad
q_k^H
\triangleq
P_T(Y_t=k\mid X_t=\mathbf{x}_t,\mathcal{H}_t=h).
\label{eq:app_oracle_posteriors}
\end{equation}
Conditioning additionally on the maintained target context and applying
Bayes' rule gives
\begin{align}
q_k^H
&=
\frac{
P_T(\mathcal{H}_t=h\mid Y_t=k,X_t=\mathbf{x}_t)
P_T(Y_t=k\mid X_t=\mathbf{x}_t)
}{
P_T(\mathcal{H}_t=h\mid X_t=\mathbf{x}_t)
}
\nonumber\\
&=
q_k^0
\frac{
P_T(\mathcal{H}_t=h\mid Y_t=k,X_t=\mathbf{x}_t)
}{
P_T(\mathcal{H}_t=h\mid X_t=\mathbf{x}_t)
}.
\label{eq:app_history_conditioning}
\end{align}
Thus, historical conditioning reweights the current-sample posterior by a class-dependent evidence term. The conditional pointwise mutual information (C-PMI)~\citep{PMI, C-PMI} associated with the realization is
\begin{align}
i(Y_t=k;\mathcal{H}_t=h\mid X_t=\mathbf{x}_t)
&\triangleq
\log
\frac{
P_T(Y_t=k\mid \mathcal{H}_t=h,X_t=\mathbf{x}_t)
}{
P_T(Y_t=k\mid X_t=\mathbf{x}_t)
}
\nonumber\\
&=
\log
\frac{
P_T(Y_t=k,\mathcal{H}_t=h\mid X_t=\mathbf{x}_t)
}{
P_T(Y_t=k\mid X_t=\mathbf{x}_t)
P_T(\mathcal{H}_t=h\mid X_t=\mathbf{x}_t)
}
\nonumber\\
&=
\log
\frac{
P_T(\mathcal{H}_t=h\mid Y_t=k,X_t=\mathbf{x}_t)
}{
P_T(\mathcal{H}_t=h\mid X_t=\mathbf{x}_t)
}
\nonumber\\
&=
\log\frac{q_k^H}{q_k^0}.
\label{eq:app_cpmi}
\end{align}
We therefore obtain
\begin{equation}
\xi_{t,k}
=
\log\frac{q_{t,k}^H}{q_{t,k}^0}
=
i(Y_t=k;\mathcal{H}_t\mid\mathbf{x}_t).
\label{eq:app_historical_evidence}
\end{equation}
Equivalently,
\begin{equation}
\log q_{t,k}^H
=
\log q_{t,k}^0
+
\xi_{t,k}.
\label{eq:app_log_evidence}
\end{equation}
Hence, the class-wise log-posterior correction induced by accumulated target
context is exactly a conditional pointwise mutual-information quantity,
rather than an ad hoc calibration score.

\subsection{Non-Negative Oracle Predictive Value}
\label{app:Oracle_risk}

\paragraph{Proposition~\ref{prop:nonnegative_history_value}}[Non-Negative Predictive Value of Target History] Conditioning on the accumulated target context $\mathcal{H}_t$ yields non-negative oracle predictive value under logarithmic loss:
\begin{equation}
\mathcal{R}_t(\mathbf{q}_t^0)
-
\mathcal{R}_t(\mathbf{q}_t^H)
=
D_{\mathrm{KL}}
\big(
\mathbf{q}_t^H\Vert\mathbf{q}_t^0
\big)
=
\mathbb{E}_{Y_t\sim\mathbf{q}_t^H}
\big[
\xi_{t,Y_t}
\big]
\ge 0.
\label{eq:nonnegative_history_value}
\end{equation}
The inequality is strict whenever
$\mathbf{q}_t^H\neq\mathbf{q}_t^0$.

\paragraph{Proof of Proposition~\ref{prop:nonnegative_history_value}} For any predictive distribution
$\mathbf{r}\in\Delta^{K-1}$, define the conditional logarithmic risk under the oracle history-conditioned target posterior as
\begin{equation}
\mathcal{R}_t(\mathbf{r})
\triangleq
\mathbb{E}_{Y_t\sim\mathbf{q}_t^H}
\left[-\log r_{Y_t}\right]
=
-\sum_{k=1}^{K} q_{t,k}^H \log r_k.
\label{eq:appendix_conditional_risk}
\end{equation}
The predictive value of conditioning on the maintained target context is therefore
\begin{align}
\mathcal{R}_t(\mathbf{q}_t^0)
-
\mathcal{R}_t(\mathbf{q}_t^H)
&=
-\sum_{k=1}^{K} q_{t,k}^H \log q_{t,k}^0
+
\sum_{k=1}^{K} q_{t,k}^H \log q_{t,k}^H
\nonumber\\
&=
\sum_{k=1}^{K}
q_{t,k}^H
\log
\frac{q_{t,k}^H}{q_{t,k}^0}
\nonumber\\
&=
D_{\mathrm{KL}}
\big(
\mathbf{q}_t^H
\Vert
\mathbf{q}_t^0
\big)
\nonumber\\
&=
\mathbb{E}_{Y_t\sim\mathbf{q}_t^H}
\left[
\xi_{t,Y_t}
\right]
\ge 0,
\label{eq:appendix_nonnegative_value}
\end{align}
where the last equality follows from the C-PMI $\xi_{t,k}=\log(q_{t,k}^H/q_{t,k}^0)$, and the inequality follows from the non-negativity of KL divergence. Equality holds if and only if $\mathbf{q}_t^H=\mathbf{q}_t^0$.
\hfill$\square$

This result characterizes the value of conditioning on history under the oracle target posterior; it does not guarantee that an estimate constructed from accumulated unlabeled statistics improves upon the source prediction. The latter depends on the estimation gap, as shown in the practical correction-gain decomposition below.


\vspace{-3mm}
\subsection{Practical Correction Gain}
\label{app:Practical_risk}

\vspace{-1mm}
The oracle result above assumes access to the true history-conditioned posterior $\mathbf{q}_t^H$. In practice, CTTA can only infer an estimate $\hat{\mathbf{q}}_t^H$ from finite, unlabeled target observations accumulated so far. Consider fully replacing the source prediction $\mathbf{s}_t$ with $\hat{\mathbf{q}}_t^H$. The resulting reduction in conditional logarithmic risk is
\vspace{-4mm}
\begin{align}
\Delta_t^{\mathrm{full}}
&\triangleq
\mathcal{R}_t(\mathbf{s}_t)
-
\mathcal{R}_t(\hat{\mathbf{q}}_t^H)
\nonumber 
=
\sum_{k=1}^{K}
q_{t,k}^H
\log
\frac{\hat q_{t,k}^H}{s_{t,k}}
\nonumber\\
&=
\sum_{k=1}^{K}
q_{t,k}^H
\left(
\log\frac{q_{t,k}^H}{s_{t,k}}
-
\log\frac{q_{t,k}^H}{\hat q_{t,k}^H}
\right)
\nonumber\\
&=
\underbrace{
D_{\mathrm{KL}}
\big(
\mathbf{q}_t^H\Vert\mathbf{s}_t
\big)
}_{\text{correction potential}}
-
\underbrace{
D_{\mathrm{KL}}
\big(
\mathbf{q}_t^H\Vert\hat{\mathbf{q}}_t^H
\big)
}_{\text{estimation gap}}.
\label{eq:appendix_practical_gain}
\end{align}

\vspace{-3mm}
Unlike the non-negative oracle value, the practical correction gain $\Delta_t^{\mathrm{full}}$ is not guaranteed to be positive. The first term measures how far the source prediction lies from the oracle history-conditioned posterior, whereas the second measures the remaining gap between its finite-sample estimate and the oracle. Consequently, full correction is beneficial if and only if
\begin{equation}
D_{\mathrm{KL}}
\big(
\mathbf{q}_t^H\Vert\hat{\mathbf{q}}_t^H
\big)
<
D_{\mathrm{KL}}
\big(
\mathbf{q}_t^H\Vert\mathbf{s}_t
\big).
\label{eq:beneficial_full_correction}
\end{equation}
This limitation is particularly relevant to CTTA, where continuously evolving target distributions can render accumulated statistics noisy, biased, or stale. Consequently, methods that directly reuse historical target information for calibration may propagate estimation errors and lead to unstable or degraded adaptation.

\vspace{-3mm}
\section{Reliable Target-Side Gain Estimation}
\label{app:target_estimator}

\vspace{-2mm}
This section provides the derivations underlying the target-side estimator introduced in Sec.~\ref{sec:gain_estimation}. We first derive the source-anchored working posterior over the unknown target class centers, and then marginalize the remaining class-center uncertainty to construct a posterior-predictive evaluator of the proposed correction.

\vspace{-3mm}
\subsection{Source-Anchored Probabilistic Target Estimation}
\label{app:source_anchored_estimation}

\vspace{-1mm}
Before processing the current sample at time $t$, let $\mathcal{Z}_t^H=\{\mathbf z_i^H\}_{i=1}^{N_t^H}$ denote the historical target representations from $\mathcal H_t$, where $\mathbf z_i^H=\phi_\theta(\mathbf x_i^H)\in\mathbb R^D$.
Throughout the following derivation, we condition on the soft responsibilities induced by preceding target predictions and the pre-t covariance estimate $\Sigma_{t-1}$. Thus, the uncertainty derived below characterizes uncertainty in the target class center conditional on the
current historical statistical state.

For each class $k$, we assume the class-conditional Gaussian model
$p_T
\left(
\mathbf{z}_i^H
\!\mid\!
Y_i\!=\!k,\boldsymbol{\mu}_k
\right)
\!\!\!=\!\!\!
\mathcal{N}
\left(
\mathbf{z}_i^H;
\boldsymbol{\mu}_k,
\Sigma_{t-1}
\right)$, 
and place a source-centered prior on the unknown target class center 
\vspace{-2mm}
\begin{equation}
p(\boldsymbol{\mu}_k)
=
\mathcal{N}
\left(
\boldsymbol{\mu}_k;
\mathbf{c}_k,
\frac{\Sigma_{t-1}}{\kappa_0}
\right),
\label{eq:app_source_prior}
\end{equation}
where $\mathbf{c}_k$ denotes the source-derived class prototype and $\kappa_0$ controls the strength of the source-centered prior. At a fixed time $t$, $\Sigma_{t-1}$ is shared across classes, while its estimate is updated causally as new target observations become available.

Since historical target labels are unavailable, each historical representation $\mathbf z_i^H$ contributes to class $k$ through a soft responsibility $\omega_{i,k}\in[0,1]$ in Eq.~\ref{eq:fractional_class_weight}, determined by the reliability-weighted prediction when the observation is incorporated into the target state. These responsibilities are used only to derive the corresponding sufficient statistics and need not be stored explicitly.
Conditional on the soft responsibilities, we define the class-$k$ fractional likelihood~\citep{bissiri2016general}:
\begin{equation}
\widetilde{\mathcal{L}}_{t,k}(\boldsymbol{\mu}_k)
\triangleq
\prod_{i=1}^{N_t^H}
\mathcal{N}
\left(
\mathbf{z}_i^H;
\boldsymbol{\mu}_k,
\Sigma_{t-1}
\right)^{\omega_{i,k}}.
\label{eq:app_fractional_likelihood}
\end{equation}
Combining the fractional likelihood with the source-centered prior gives
\begin{equation}
\widetilde p
\left(
\boldsymbol{\mu}_k
\mid
\mathcal{Z}_t^H,
\{\omega_{i,k}\}_{i=1}^{N_t^H},
\Sigma_{t-1}
\right)
\propto
p(\boldsymbol{\mu}_k)
\widetilde{\mathcal{L}}_{t,k}(\boldsymbol{\mu}_k).
\label{eq:app_fractional_posterior}
\end{equation}
For brevity, we suppress the fixed conditioning on the soft responsibilities and $\Sigma_{t-1}$ below. Taking the negative logarithm of Eq.~\ref{eq:app_fractional_posterior} and multiplying by $2$, we obtain the following expression, where $\doteq$ denotes equality up to additive terms independent of $\boldsymbol{\mu}_k$:
\begin{align}
-2\log \widetilde p
\left(
\boldsymbol{\mu}_k
\mid
\mathcal{Z}_t^H
\right)
\doteq\;&
-2\log p(\boldsymbol{\mu}_k)
-
2\log
\widetilde{\mathcal L}_{t,k}
(\boldsymbol{\mu}_k)
\nonumber\\
\doteq\;&
\kappa_0
(\boldsymbol{\mu}_k-\mathbf{c}_k)^\top
\Sigma_{t-1}^{-1}
(\boldsymbol{\mu}_k-\mathbf{c}_k)
+
\sum\nolimits_{i=1}^{N_t^H}
\omega_{i,k}
(\mathbf{z}_i^H-\boldsymbol{\mu}_k)^\top
\Sigma_{t-1}^{-1}
(\mathbf{z}_i^H-\boldsymbol{\mu}_k)
\nonumber\\[1mm]
=\;&
\kappa_0
\Big[
\boldsymbol{\mu}_k^\top
\Sigma_{t-1}^{-1}
\boldsymbol{\mu}_k
-
2\mathbf{c}_k^\top
\Sigma_{t-1}^{-1}
\boldsymbol{\mu}_k
+
\mathbf{c}_k^\top
\Sigma_{t-1}^{-1}
\mathbf{c}_k
\Big]
\nonumber\\
&+
\sum\nolimits_{i=1}^{N_t^H}
\omega_{i,k}
\Big[
(\mathbf{z}_i^H)^\top
\Sigma_{t-1}^{-1}
\mathbf{z}_i^H
-
2(\mathbf{z}_i^H)^\top
\Sigma_{t-1}^{-1}
\boldsymbol{\mu}_k
+
\boldsymbol{\mu}_k^\top
\Sigma_{t-1}^{-1}
\boldsymbol{\mu}_k
\Big]
\nonumber\\[1mm]
=\;&
\left(
\kappa_0
+
\sum\nolimits_{i=1}^{N_t^H}\omega_{i,k}
\right)
\boldsymbol{\mu}_k^\top
\Sigma_{t-1}^{-1}
\boldsymbol{\mu}_k-
2
\left(
\kappa_0\mathbf{c}_k
+
\sum\nolimits_{i=1}^{N_t^H}
\omega_{i,k}\mathbf{z}_i^H
\right)^\top
\Sigma_{t-1}^{-1}
\boldsymbol{\mu}_k
\nonumber\\
&+
\underbrace{
\kappa_0
\mathbf{c}_k^\top
\Sigma_{t-1}^{-1}
\mathbf{c}_k
+
\sum\nolimits_{i=1}^{N_t^H}
\omega_{i,k}
(\mathbf{z}_i^H)^\top
\Sigma_{t-1}^{-1}
\mathbf{z}_i^H
}_{\text{independent of }\boldsymbol{\mu}_k}
\nonumber\\
\doteq\;&
\left(
\kappa_0
+
\sum_{i=1}^{N_t^H}\omega_{i,k}
\right)
\boldsymbol{\mu}_k^\top
\Sigma_{t-1}^{-1}
\boldsymbol{\mu}_k
-
2
\left(
\kappa_0\mathbf{c}_k
+
\sum_{i=1}^{N_t^H}
\omega_{i,k}\mathbf{z}_i^H
\right)^\top
\Sigma_{t-1}^{-1}
\boldsymbol{\mu}_k .
\label{eq:app_fractional_expansion}
\end{align}

Define the posterior precision parameter and the corresponding
source-anchored posterior-mean center as
\begin{equation}
\kappa_{t-1,k}
=
\kappa_0
+
\sum\nolimits_{i=1}^{N_t^H}\omega_{i,k},
\qquad
\mathbf{m}_{t-1,k}
=
\frac{
\kappa_0\mathbf{c}_k
+
\sum\nolimits_{i=1}^{N_t^H}
\omega_{i,k}\mathbf{z}_i^H
}{
\kappa_{t-1,k}
}.
\label{eq:app_center_statistics}
\end{equation}

Substituting Eq.~\ref{eq:app_center_statistics} into Eq.~\ref{eq:app_fractional_expansion} and completing the square gives
\begin{align}
-2\log \widetilde p
\left(
\boldsymbol{\mu}_k
\mid
\mathcal{Z}_t^H
\right)
\doteq\;&
\kappa_{t-1,k}
\boldsymbol{\mu}_k^\top
\Sigma_{t-1}^{-1}
\boldsymbol{\mu}_k
-
2\kappa_{t-1,k}
\mathbf{m}_{t-1,k}^\top
\Sigma_{t-1}^{-1}
\boldsymbol{\mu}_k
\nonumber\\
\doteq\;&
\kappa_{t-1,k}
\left(
\boldsymbol{\mu}_k-\mathbf{m}_{t-1,k}
\right)^\top
\Sigma_{t-1}^{-1}
\left(
\boldsymbol{\mu}_k-\mathbf{m}_{t-1,k}
\right).
\label{eq:app_complete_square}
\end{align}
Therefore, conditional on the effective responsibilities and the pre-$t$ covariance estimate, the induced fractional posterior is
\begin{equation}
\widetilde p
\left(
\boldsymbol{\mu}_k
\mid
\mathcal{Z}_t^H,
\{\omega_{i,k}\}_{i=1}^{N_t^H},
\Sigma_{t-1}
\right)
=
\mathcal{N}
\left(
\boldsymbol{\mu}_k;
\mathbf{m}_{t-1,k},
\frac{\Sigma_{t-1}}{\kappa_{t-1,k}}
\right).
\label{eq:app_center_posterior}
\end{equation}
Hence, the posterior mean $\mathbf m_{t-1,k}$ provides the history-conditioned estimate of the target class center, while $\Sigma_{t-1}/\kappa_{t-1,k}$ characterizes the corresponding center-estimation uncertainty. Here, $\kappa_{t-1,k}$ measures the accumulated reliability-weighted support for class $k$, so stronger accumulated support yields a more concentrated estimate of its target center. The soft responsibilities are introduced only for derivation. In practice, their aggregate effect is maintained through recursive sufficient statistics in Appendix~\ref{app:continual_statistics}, without storing or replaying historical representations.

\paragraph{Posterior-Mean Target Proposal.}
Given the class-center posterior in
Eq.~\ref{eq:app_center_posterior}, we first construct a target-side
proposal using its posterior mean $\mathbf m_{t-1,k}$.
Specifically, plugging $\mathbf m_{t-1,k}$ into the class-conditional
Gaussian model gives
\begin{equation}
\hat{p}_T
\left(
\mathbf z_t
\mid
Y_t=k,\mathcal H_t
\right)
\triangleq
\mathcal N
\left(
\mathbf z_t;
\mathbf m_{t-1,k},
\Sigma_{t-1}
\right).
\label{eq:app_pm_likelihood}
\end{equation}
Applying Bayes' rule with the pre-$t$ class prior
$\pi_{t-1,k}$ yields
\begin{equation}
\hat q_{t,k}^{H}
=
\frac{
\pi_{t-1,k}
\hat{p}_T
(\mathbf z_t\mid Y_t=k,\mathcal H_t)
}{
\sum_{j=1}^{K}
\pi_{t-1,j}
\hat{p}_T
(\mathbf z_t\mid Y_t=j,\mathcal H_t)
}.
\label{eq:app_pm_bayes}
\end{equation}
Expanding the Gaussian density,
\begin{align}
\hat{p}_T
(\mathbf z_t\mid Y_t=k,\mathcal H_t)
&=
\frac{
1
}{
(2\pi)^{D/2}
|\Sigma_{t-1}|^{1/2}
}
\exp
\left[
-\frac{1}{2}
(\mathbf z_t-\mathbf m_{t-1,k})^\top
\Sigma_{t-1}^{-1}
(\mathbf z_t-\mathbf m_{t-1,k})
\right].
\label{eq:app_pm_density}
\end{align}
Define the corresponding squared Mahalanobis distance
\begin{equation}
d_{t,k}
\triangleq
(\mathbf z_t-\mathbf m_{t-1,k})^\top
\Sigma_{t-1}^{-1}
(\mathbf z_t-\mathbf m_{t-1,k}).
\label{eq:app_mahalanobis}
\end{equation}
Since $\Sigma_{t-1}$ is shared across classes, the Gaussian normalization
factor
$(2\pi)^{-D/2}|\Sigma_{t-1}|^{-1/2}$
is independent of $k$ and therefore cancels in
Eq.~\ref{eq:app_pm_bayes}. Hence,
\begin{equation}
\hat q_{t,k}^{H}
=
\frac{
\pi_{t-1,k}\exp(-d_{t,k}/2)
}{
\sum_{j=1}^{K}
\pi_{t-1,j}\exp(-d_{t,j}/2)
}
\propto
\pi_{t-1,k}
\exp\left(-\frac{1}{2}d_{t,k}\right).
\label{eq:app_target_proposal}
\end{equation}
Equivalently, the class-wise discriminant score can be written as
\begin{equation}
\hat{\ell}_{t,k}^{H}
=
\log\pi_{t-1,k}
-
\frac{1}{2}d_{t,k},
\qquad
\hat{\mathbf q}_t^H
=
\operatorname{softmax}
(\hat{\boldsymbol\ell}_t^H).
\label{eq:app_target_score}
\end{equation}
Thus, $\hat{\mathbf q}_t^H$ is the target-side correction proposed by the posterior-mean target geometry. Importantly, this is a plug-in estimate: it uses the posterior mean $\mathbf m_{t-1,k}$ but does not yet account for the remaining class-center uncertainty
$\Sigma_{t-1}/\kappa_{t-1,k}$. We marginalize this uncertainty next to
construct the posterior-predictive evaluator.

\subsection{Posterior-Predictive Gain Evaluation}
\label{app:posterior_predictive_gain}

The posterior-mean proposal $\hat{\mathbf q}_t^H$ above is constructed by plugging the posterior mean $\mathbf m_{t-1,k}$ into the class-conditional model.
We next marginalize the remaining uncertainty in the target class centers and use the resulting posterior-predictive distribution to evaluate the same proposed correction.

\paragraph{Posterior-Predictive Target Distribution.}
Recall from Eq.~\ref{eq:app_center_posterior} that
\begin{equation}
\boldsymbol{\mu}_k
\mid
\mathcal H_t
\sim
\mathcal N
\left(
\mathbf m_{t-1,k},
\frac{\Sigma_{t-1}}{\kappa_{t-1,k}}
\right),
\label{eq:app_pp_center_posterior}
\end{equation}
while the class-conditional representation model is
\begin{equation}
\mathbf z_t
\mid
Y_t=k,\boldsymbol{\mu}_k
\sim
\mathcal N
\left(
\boldsymbol{\mu}_k,
\Sigma_{t-1}
\right).
\label{eq:app_pp_observation_model}
\end{equation}
Marginalizing the latent class center gives
\begin{align}
p_T^{\mathrm{pp}}
(\mathbf z_t\mid Y_t=k,\mathcal H_t)
&=
\int
p_T
(\mathbf z_t\mid Y_t=k,\boldsymbol{\mu}_k)
\,
\widetilde p
(\boldsymbol{\mu}_k\mid\mathcal H_t)
\,d\boldsymbol{\mu}_k
\nonumber\\
&=
\mathcal N
\left(
\mathbf z_t;
\mathbf m_{t-1,k},
\Sigma_{t-1}
+
\frac{\Sigma_{t-1}}{\kappa_{t-1,k}}
\right)
\nonumber\\
&=
\mathcal N
\left(
\mathbf z_t;
\mathbf m_{t-1,k},
h_{t,k}\Sigma_{t-1}
\right),
\label{eq:app_pp_likelihood}
\end{align}
where
\begin{equation}
h_{t,k}
\triangleq
1+\kappa_{t-1,k}^{-1}.
\label{eq:app_pp_scale}
\end{equation}
Hence, the posterior-predictive covariance accounts for both the
within-class representation variability and the remaining uncertainty in
the estimated class center.

Applying Bayes' rule with the pre-$t$ class prior
$\pi_{t-1,k}$ gives
\begin{equation}
\bar q_{t,k}^{H}
=
\frac{
\pi_{t-1,k}
p_T^{\mathrm{pp}}
(\mathbf z_t\mid Y_t=k,\mathcal H_t)
}{
\sum_{j=1}^{K}
\pi_{t-1,j}
p_T^{\mathrm{pp}}
(\mathbf z_t\mid Y_t=j,\mathcal H_t)
}.
\label{eq:app_pp_bayes}
\end{equation}
Expanding Eq.~\ref{eq:app_pp_likelihood},
\begin{align}
p_T^{\mathrm{pp}}
(\mathbf z_t\mid Y_t=k,\mathcal H_t)
&=
\frac{
\exp
\left[
-\frac{1}{2}
(\mathbf z_t-\mathbf m_{t-1,k})^\top
(h_{t,k}\Sigma_{t-1})^{-1}
(\mathbf z_t-\mathbf m_{t-1,k})
\right]
}{
(2\pi)^{D/2}
|h_{t,k}\Sigma_{t-1}|^{1/2}
}.
\label{eq:app_pp_density}
\end{align}
Using
\begin{equation}
|h_{t,k}\Sigma_{t-1}|
=
h_{t,k}^{D}|\Sigma_{t-1}|,
\qquad
(h_{t,k}\Sigma_{t-1})^{-1}
=
h_{t,k}^{-1}\Sigma_{t-1}^{-1},
\label{eq:app_pp_matrix_identities}
\end{equation}
together with the Mahalanobis distance
$d_{t,k}$ in Eq.~\ref{eq:app_mahalanobis}, we obtain
\begin{equation}
p_T^{\mathrm{pp}}
(\mathbf z_t\mid Y_t=k,\mathcal H_t)
=
C_t\,
h_{t,k}^{-D/2}
\exp
\left(
-\frac{d_{t,k}}{2h_{t,k}}
\right),
\label{eq:app_pp_simplified}
\end{equation}
where
$C_t=(2\pi)^{-D/2}|\Sigma_{t-1}|^{-1/2}$
is independent of $k$ because $\Sigma_{t-1}$ is shared across classes.
The common factor therefore cancels during class normalization, yielding
\begin{equation}
\bar q_{t,k}^{H}
=
\frac{
\pi_{t-1,k}
h_{t,k}^{-D/2}
\exp\!\left(
-\dfrac{d_{t,k}}{2h_{t,k}}
\right)
}{
\sum_{j=1}^{K}
\pi_{t-1,j}
h_{t,j}^{-D/2}
\exp\!\left(
-\dfrac{d_{t,j}}{2h_{t,j}}
\right)
}.
\label{eq:app_pp_target}
\end{equation}
Equivalently, $\bar{\ell}_{t,k}^{H}=\log\pi_{t-1,k}-\frac{D}{2}\log h_{t,k}-\frac{d_{t,k}}{2h_{t,k}}$,
and $\bar{\mathbf q}_t^H=\operatorname{softmax}(\bar{\boldsymbol\ell}_t^H)$.

%
Compared with the posterior-mean proposal, $h_{t,k}$ attenuates the distance penalty for uncertain class centers, while $(D/2)\log h_{t,k}$ accounts for the corresponding increase in predictive volume. The latter cannot, in general, be removed when $\kappa_{t-1,k}$ varies across classes.

\paragraph{Posterior-Predictive Correction Gain.}
Recall the source-relative correction evidence $\hat{\xi}_{t,k}^{s}\triangleq\log\frac{\hat q_{t,k}^{H}}{s_{t,k}}$. The oracle practical gain in Eq.~\ref{eq:appendix_practical_gain} can be written as 
$\Delta_t^{\mathrm{full}}\!=\!(\mathbf q_t^H)^\top\hat{\boldsymbol\xi}_t^s$, where the unknown $\mathbf q_t^H$ evaluates the correction proposed by
$\hat{\mathbf q}_t^H$. Since $\mathbf q_t^H$ is unavailable at test time, we instead evaluate the same correction under the posterior-predictive target distribution:
\begin{equation}
G_t^{\mathrm{pp}}
\triangleq
(\bar{\mathbf q}_t^H)^\top
\hat{\boldsymbol\xi}_t^s
=
\sum_{k=1}^{K}
\bar q_{t,k}^{H}
\log
\frac{\hat q_{t,k}^{H}}{s_{t,k}}.
\label{eq:app_pp_gain}
\end{equation}
Equivalently, defining the posterior-predictive logarithmic risk
\begin{equation}
\bar{\mathcal R}_t(\mathbf p)
\triangleq
-
\sum_{k=1}^{K}
\bar q_{t,k}^{H}\log p_k,
\end{equation}
we have $G_t^{\mathrm{pp}}=\bar{\mathcal R}_t(\mathbf s_t)-\bar{\mathcal R}_t(\hat{\mathbf q}_t^H)$.  Moreover,
\begin{align}
G_t^{\mathrm{pp}}
&=
\sum_{k=1}^{K}
\bar q_{t,k}^{H}
\left[
\log
\frac{\bar q_{t,k}^{H}}{s_{t,k}}
-
\log
\frac{\bar q_{t,k}^{H}}{\hat q_{t,k}^{H}}
\right]
\nonumber\\
&=
D_{\mathrm{KL}}
\left(
\bar{\mathbf q}_t^H
\Vert
\mathbf s_t
\right)
-
D_{\mathrm{KL}}
\left(
\bar{\mathbf q}_t^H
\Vert
\hat{\mathbf q}_t^H
\right)
.
\label{eq:app_pp_gain_kl}
\end{align}
The first term measures the potential benefit of correcting the source prediction, whereas the second measures the mismatch between the proposed correction and its posterior-predictive evaluation. Hence, source--target disagreement alone does not imply a reliable correction; the correction must remain supported after accounting for uncertainty in the estimated target geometry.

\paragraph{Relation to the Oracle Gain.}
The posterior-predictive gain evaluates the same source-relative
correction as the oracle practical gain, but replaces the unknown
history-conditioned target posterior $\mathbf q_t^H$ with the
posterior-predictive evaluator $\bar{\mathbf q}_t^H$. Their difference is
therefore
\begin{equation}
\Delta_t^{\mathrm{full}}
-
G_t^{\mathrm{pp}}
=
\left(
\mathbf q_t^H-\bar{\mathbf q}_t^H
\right)^\top
\hat{\boldsymbol\xi}_t^s.
\label{eq:app_pp_oracle_gap}
\end{equation}
Let $\operatorname{span}(\hat{\boldsymbol\xi}_t^s)\triangleq\max_k\hat\xi_{t,k}^s-\min_k\hat\xi_{t,k}^s$.
Since both $\mathbf q_t^H$ and $\bar{\mathbf q}_t^H$ are probability distributions, their difference sums to zero. Hence,
\begin{align}
\left|
\Delta_t^{\mathrm{full}}
-
G_t^{\mathrm{pp}}
\right|
&\le
\operatorname{TV}
\left(
\mathbf q_t^H,
\bar{\mathbf q}_t^H
\right)
\operatorname{span}
(\hat{\boldsymbol\xi}_t^s),
\label{eq:app_pp_oracle_tv_bound}
\end{align}
where $\operatorname{TV}(\mathbf p, \!\mathbf q)\!=\!\frac12\|\mathbf p\!-\!\mathbf q\|_1$.
By Pinsker's inequality~\citep{cover1991elements}, this further implies
\begin{equation}
\left|
\Delta_t^{\mathrm{full}}
-
G_t^{\mathrm{pp}}
\right|
\le
\operatorname{span}
(\hat{\boldsymbol\xi}_t^s)
\sqrt{
\frac12
D_{\mathrm{KL}}
\left(
\mathbf q_t^H
\Vert
\bar{\mathbf q}_t^H
\right)
}.
\label{eq:app_pp_oracle_kl_bound}
\end{equation}
Thus, the discrepancy between the posterior-predictive and oracle gains is controlled jointly by the accuracy of the posterior-predictive evaluator and the magnitude of the proposed correction. In particular, $G_t^{\mathrm{pp}}$ approaches the oracle practical gain whenever $\bar{\mathbf q}_t^H$ approaches $\mathbf q_t^H$, while aggressive source-relative corrections amplify errors in the target evaluator. For numerical stability, probabilities entering log-ratios are lower bounded by $\epsilon_p$ before renormalization.

\vspace{-3mm}
\section{Posterior-Predictive Evidence Intervention}
\label{app:intervention}

\vspace{-1mm}
This section provides the derivations underlying the posterior-predictive
evidence intervention introduced in Sec.~\ref{sec:evidence_intervention}.
We first derive the continuous source-to-target evidence path and its exact
posterior-predictive risk reduction. We then establish the concavity of the
resulting objective, characterize its boundary behavior, and derive an
efficient global solution for the intervention strength.

\vspace{-3mm}
\subsection{Continuous Evidence Path and Predictive Objective}
\label{app:evidence_path}

\vspace{-2mm}
\paragraph{Continuous Evidence Path.}
To control the extent of target-side intervention, we construct a continuous evidence path by scaling the source-relative evidence with $\lambda\in[0,1]$. Recall that $\hat{\xi}_{t,k}^{s}=\log(\hat q_{t,k}^{H}/s_{t,k})$. Scaling this evidence and normalizing across classes gives
\begin{equation}
p_{t,k}^{(\lambda)}
=
\frac{
s_{t,k}\exp(\lambda\hat{\xi}_{t,k}^{s})
}{
Z_t(\lambda)
}
=
\frac{
s_{t,k}^{\,1-\lambda}
(\hat q_{t,k}^{H})^{\lambda}
}{
\sum_{j=1}^{K}
s_{t,j}^{\,1-\lambda}
(\hat q_{t,j}^{H})^{\lambda}
},
\label{eq:app_evidence_path}
\end{equation}
where $Z_t(\lambda)\!=\!\sum\nolimits_{j=1}^{K}s_{t,j}\exp(\lambda\hat{\xi}_{t,j}^{s})$ is the normalization factor.
The two endpoints satisfy $\mathbf p_t^{(0)}\!=\!\mathbf s_t$ and $\mathbf p_t^{(1)}\!=\!\hat{\mathbf q}_t^H$.
Hence, $\lambda$ continuously controls the amount of target-side evidence introduced into the source prediction.

\vspace{-3mm}
\paragraph{Posterior-Predictive Path-Wise Gain.}
As established in Sec.~\ref{sec:gain_estimation}, the posterior-predictive
distribution $\bar{\mathbf q}_t^H$ evaluates the correction proposed by
$\hat{\mathbf q}_t^H$ after accounting for uncertainty in the estimated
target geometry. Define the corresponding conditional logarithmic risk as
\vspace{-2mm}
\begin{equation}
\bar{\mathcal R}_t(\mathbf p)
\triangleq
-
\sum_{k=1}^{K}
\bar q_{t,k}^{H}\log p_k.
\label{eq:app_predictive_risk}
\end{equation}

\vspace{-5mm}
From Eq.~\ref{eq:app_evidence_path}, $\log\frac{p_{t,k}^{(\lambda)}}{s_{t,k}}\!=\!\lambda\hat{\xi}_{t,k}^{s}-\log Z_t(\lambda)$. 
Therefore, the posterior-predictive risk reduction relative to the source prediction is
\vspace{-2mm}
\begin{align}
\mathcal J_t(\lambda)
&\triangleq
\bar{\mathcal R}_t(\mathbf s_t)
-
\bar{\mathcal R}_t
\bigl(
\mathbf p_t^{(\lambda)}
\bigr)
\nonumber
=
\sum_{k=1}^{K}
\bar q_{t,k}^{H}
\log
\frac{
p_{t,k}^{(\lambda)}
}{
s_{t,k}
}
\nonumber\\
&=
\lambda
(\bar{\mathbf q}_t^H)^\top
\hat{\boldsymbol{\xi}}_t^s
-
\log Z_t(\lambda)
\nonumber\\
&=
\lambda G_t^{\mathrm{pp}}
-
\log Z_t(\lambda),
\qquad
\lambda\in[0,1].
\label{eq:app_predictive_objective}
\end{align}
Here, $G_t^{\mathrm{pp}}$ is the posterior-predictive correction gain introduced in
Sec.~\ref{sec:gain_estimation}.
Thus, Eq.~\ref{eq:app_predictive_objective} gives the exact reduction in posterior-predictive logarithmic risk along the evidence path.
The same proposed correction $\hat{\mathbf q}_t^H$ determines the path, whereas $\bar{\mathbf q}_t^H$ determines how strongly that correction is supported under the uncertainty-aware target working model.

Equivalently, since
$\bar{\mathcal R}_t(\mathbf p)=H(\bar{\mathbf q}_t^H)+D_{\mathrm{KL}}\left(\bar{\mathbf q}_t^H\Vert\mathbf p\right)$, maximizing $\mathcal J_t(\lambda)$ is equivalent to finding the point on the source--target evidence path that minimizes $D_{\mathrm{KL}}\left(\bar{\mathbf q}_t^H\Vert\mathbf  p_t^{(\lambda)}\right)$.
Hence, the intervention coefficient can also be interpreted as the forward-KL projection of the posterior-predictive evaluator onto the continuous source--target evidence path.

\vspace{-1mm}
\subsection{Globally Optimal Posterior-Predictive Intervention}
\label{app:optimal_intervention}
\vspace{-1mm}
We now establish the global solution of
Eq.~\ref{eq:app_predictive_objective} and prove Theorem~\ref{thm:risk_calibrated_intervention}.

\vspace{-1mm}
\paragraph{Concavity and Global Optimality.}
Since $G_t^{\mathrm{pp}}$ is independent of $\lambda$, differentiating the log-normalizer gives
\begin{align}
\frac{\mathrm d}{\mathrm d\lambda}
\log Z_t(\lambda)
&=
\frac{
\sum_{k=1}^{K}
s_{t,k}
\exp(\lambda\hat{\xi}_{t,k}^{s})
\hat{\xi}_{t,k}^{s}
}{
Z_t(\lambda)
}=
\sum\nolimits_{k=1}^{K}
p_{t,k}^{(\lambda)}
\hat{\xi}_{t,k}^{s}
=
\mathbb E_{k\sim\mathbf p_t^{(\lambda)}}
\left[
\hat{\xi}_{t,k}^{s}
\right].
\label{eq:app_log_partition_gradient}
\end{align}
Therefore, $\mathcal J_t'(\lambda)=G_t^{\mathrm{pp}}-\mathbb E_{k\sim\mathbf p_t^{(\lambda)}}\left[\hat{\xi}_{t,k}^{s}\right]$.
Differentiating once more yields
\begin{align}
\mathcal J_t''(\lambda)
&=
-
\frac{\mathrm d^2}{\mathrm d\lambda^2}
\log Z_t(\lambda)
=
-
\left\{
\mathbb E_{k\sim\mathbf p_t^{(\lambda)}}
\left[
(\hat{\xi}_{t,k}^{s})^2
\right]
-
\left(
\mathbb E_{k\sim\mathbf p_t^{(\lambda)}}
\left[
\hat{\xi}_{t,k}^{s}
\right]
\right)^2
\right\}
\nonumber\\
&=
-
\operatorname{Var}_{k\sim\mathbf p_t^{(\lambda)}}
\left[
\hat{\xi}_{t,k}^{s}
\right]
\leq 0.
\label{eq:app_objective_concavity}
\end{align}
Hence, $\mathcal J_t(\lambda)$ is concave over $\lambda\in[0,1]$.
If $\hat{\mathbf q}_t^H\neq\mathbf s_t$, the source-relative evidence is not constant across classes and the variance is strictly positive, yielding strict concavity. Therefore, the maximizer is unique except in the degenerate case $\hat{\mathbf q}_t^H=\mathbf s_t$.

\vspace{-2mm}
\paragraph{Boundary Behavior.}
The derivatives at the two endpoints further characterize when intervention is suppressed or fully applied.
At $\lambda=0$, $\mathbf p_t^{(0)}=\mathbf s_t$, and therefore
\vspace{-2mm}
\begin{align}
\mathcal J_t'(0)
&=
G_t^{\mathrm{pp}}
-
\sum\nolimits_{k=1}^{K}
s_{t,k}
\log
\frac{
\hat q_{t,k}^{H}
}{
s_{t,k}
}
\nonumber\\
&=
G_t^{\mathrm{pp}}
+
D_{\mathrm{KL}}
\left(
\mathbf s_t
\Vert
\hat{\mathbf q}_t^H
\right).
\label{eq:app_source_boundary}
\end{align}
Since $\mathcal J_t$ is concave, $\mathcal J_t'(0)\leq0$ implies that the objective is non-increasing from the source endpoint, and hence
\vspace{-2mm}
\begin{equation}
G_t^{\mathrm{pp}}
\leq
-
D_{\mathrm{KL}}
\left(
\mathbf s_t
\Vert
\hat{\mathbf q}_t^H
\right)
\quad\Longrightarrow\quad
\lambda_t^\star=0.
\label{eq:app_source_retention}
\end{equation}
Thus, the posterior-predictive evaluator assigns sufficiently negative gain to the proposed correction that even an infinitesimal target-side intervention is rejected.

At $\lambda=1$, $\mathbf p_t^{(1)}=\hat{\mathbf q}_t^H$, giving
\vspace{-3mm}
\begin{align}
\mathcal J_t'(1)
&=
G_t^{\mathrm{pp}}
-
\sum\nolimits_{k=1}^{K}
\hat q_{t,k}^{H}
\log
\frac{
\hat q_{t,k}^{H}
}{
s_{t,k}
}
\nonumber\\
&=
G_t^{\mathrm{pp}}
-
D_{\mathrm{KL}}
\left(
\hat{\mathbf q}_t^H
\Vert
\mathbf s_t
\right).
\label{eq:app_target_boundary}
\end{align}

\vspace{-2mm}
Since the objective remains non-decreasing up to the target endpoint,
\begin{equation}
G_t^{\mathrm{pp}}
\geq
D_{\mathrm{KL}}
\left(
\hat{\mathbf q}_t^H
\Vert
\mathbf s_t
\right)
\quad\Longrightarrow\quad
\lambda_t^\star=1.
\label{eq:app_full_intervention}
\end{equation}
In this regime, the posterior-predictive evaluator sufficiently supports the full target-side correction.

Finally, when
\begin{equation}
\mathcal J_t'(0)>0
\qquad\text{and}\qquad
\mathcal J_t'(1)<0,
\label{eq:app_partial_condition}
\end{equation}
strict concavity guarantees a unique interior optimum
$\lambda_t^\star\in(0,1)$ satisfying $\mathbb E_{k\sim\mathbf p_t^{(\lambda_t^\star)}}\left[\hat{\xi}_{t,k}^{s}\right]=G_t^{\mathrm{pp}}$.
Thus, the optimal partial intervention is the point along the evidence
path at which the path-wise expected source-relative evidence matches the
gain supported by the posterior-predictive evaluator.

\paragraph{Efficient Global Solution.}
Combining the three regimes gives
\begin{equation}
\lambda_t^\star
=
\begin{cases}
0,
&
G_t^{\mathrm{pp}}
\leq
-
D_{\mathrm{KL}}
\left(
\mathbf s_t
\Vert
\hat{\mathbf q}_t^H
\right),
\\[2mm]
1,
&
G_t^{\mathrm{pp}}
\geq
D_{\mathrm{KL}}
\left(
\hat{\mathbf q}_t^H
\Vert
\mathbf s_t
\right),
\\[2mm]
\text{the unique root of }
\mathcal J_t'(\lambda)=0,
&
\text{otherwise}.
\end{cases}
\label{eq:app_optimal_lambda}
\end{equation}
Equivalently, the interior case corresponds to $-D_{\mathrm{KL}}\left(\mathbf s_t\Vert\hat{\mathbf q}_t^H\right)<G_t^{\mathrm{pp}}<D_{\mathrm{KL}}\left(\hat{\mathbf q}_t^H\Vert\mathbf s_t\right)$.

Since $\mathcal J_t'(\lambda)$ is monotone non-increasing,
the interior root can be found efficiently by bisection.
Reaching a tolerance $\epsilon_\lambda$ requires
$O(\log(1/\epsilon_\lambda))$ iterations, with
$O(K)$ computation per iteration.

Substituting the optimal intervention strength into the evidence path gives
\begin{equation}
p_{t,k}^{\star}
=
p_{t,k}^{(\lambda_t^\star)}
=
\frac{
s_{t,k}^{\,1-\lambda_t^\star}
(\hat q_{t,k}^{H})^{\lambda_t^\star}
}{
\sum_{j=1}^{K}
s_{t,j}^{\,1-\lambda_t^\star}
(\hat q_{t,j}^{H})^{\lambda_t^\star}
}
\propto
s_{t,k}^{\,1-\lambda_t^\star}
\bigl(\hat q_{t,k}^{H}\bigr)^{\lambda_t^\star}.
\label{eq:app_final_intervention}
\end{equation}
If $\hat{\mathbf q}_t^H=\mathbf s_t$, the evidence path collapses to a
single prediction and $\mathcal J_t(\lambda)\equiv0$. In this degenerate
case, we set $\lambda_t^\star=0$ as a conservative tie-breaking rule.

This completes the proof of Theorem~\ref{thm:risk_calibrated_intervention}.

\subsection{Scope of Optimality and Relation to Target Risk}
\label{app:optimality_scope}

\vspace{-2mm}
\paragraph{Optimality under the Working Evaluator.}
Fix the current representation and the pre-$t$ statistical state, so that $\mathbf s_t$, $\hat{\mathbf q}_t^H$, and $\bar{\mathbf q}_t^H$ remain fixed while optimizing $\lambda$. Assume
$s_{t,k},\hat q_{t,k}^H>0$ for every class.
Theorem~\ref{thm:risk_calibrated_intervention} establishes global optimality
with respect to the posterior-predictive logarithmic risk
$\bar{\mathcal R}_t(\mathbf p)
=-\sum_k\bar q_{t,k}^H\log p_k$
on the prescribed evidence path. In particular,
\begin{equation}
\lambda_t^\star
\in
\arg\min_{\lambda\in[0,1]}
\bar{\mathcal R}_t
\bigl(\mathbf p_t^{(\lambda)}\bigr).
\label{eq:scope_predictive_optimality}
\end{equation}
Since both the source prediction and the target proposal belong to
this path, the selected prediction satisfies
\begin{equation}
\bar{\mathcal R}_t(\mathbf p_t^\star)
\le
\min\left\{
\bar{\mathcal R}_t(\mathbf s_t),
\bar{\mathcal R}_t(\hat{\mathbf q}_t^H)
\right\}.
\label{eq:scope_endpoint_dominance}
\end{equation}
Indeed, optimality implies
$\mathcal J_t(\lambda_t^\star)
\ge
\max\{\mathcal J_t(0),\mathcal J_t(1)\}
=
\max\{0,G_t^{\mathrm{pp}}\}$.
This is optimality over the fixed evidence path, not over all
predictive distributions.

\subsection{Why Use the Posterior-Predictive Distribution as an Evaluator?}
\label{app:pp_evaluator}

A natural alternative is to directly use the posterior-predictive distribution
$\bar{\mathbf q}_t^H$ as the adapted prediction. However, its role in \ours is deliberately different.
The posterior-mean distribution $\hat{\mathbf q}_t^H$ specifies the correction suggested by the estimated target geometry, whereas $\bar{\mathbf q}_t^H$ marginalizes class-center uncertainty and is used to evaluate whether this correction remains beneficial relative to the frozen source prediction.

This separation is important because using the same distribution both to propose and evaluate a correction leads to a degenerate intervention. Consider an arbitrary target distribution $\mathbf r_t$ and the evidence path
\begin{equation}
p_{t,k}^{(\lambda)}
=
\frac{
s_{t,k}^{1-\lambda} r_{t,k}^{\lambda}
}{
\sum_j s_{t,j}^{1-\lambda} r_{t,j}^{\lambda}
}.
\end{equation}

\vspace{-3mm}
If $\mathbf r_t$ is also used as the evaluator, the corresponding objective is  $J_t^{r}(\lambda)=\mathbb E_{\mathbf r_t}\left[\log\frac{\mathbf p_t^{(\lambda)}}{\mathbf s_t}\right]$.
At $\lambda=1$, $\mathbf p_t^{(1)}=\mathbf r_t$ and
$
\left.\frac{\partial J_t^{r}(\lambda)}
{\partial \lambda}\right|_{\lambda=1}
=
D_{\mathrm{KL}}(\mathbf r_t\Vert\mathbf s_t)
-
\mathbb E_{\mathbf r_t}
\left[
\log\frac{\mathbf r_t}{\mathbf s_t}
\right]
=0$.

Since $J_t^{r}(\lambda)$ is concave, its optimum is attained at $\lambda^\star=1$ (except for the degenerate case $\mathbf r_t=\mathbf s_t$). Therefore, self-evaluation simply reduces to directly adopting the target distribution and provides no meaningful mechanism for deciding whether the proposed correction should be applied.

\ours instead separates the two roles: $\hat{\mathbf q}_t^H$ proposes the correction, and $\bar{\mathbf q}_t^H$ evaluates its source-relative utility.
This asymmetric proposal--evaluation design enables $\lambda_t^\star$ to reject, partially apply, or fully accept the history-derived correction rather than automatically trusting the target estimate.

\vspace{-1mm}
\paragraph{Direct posterior-predictive prediction.}
To empirically examine whether the posterior-predictive evaluator should instead be used directly as the prediction, we compare \ours with a variant that sets $\mathbf p_t^\star=\bar{\mathbf q}_t^H$.
As shown in Table~\ref{tab:ablation_role}, directly predicting with $\bar{\mathbf q}_t^H$ is inferior to using it as an evaluator of $\hat{\mathbf q}_t^H$.
This confirms that accounting for target-statistic uncertainty is most effective for assessing the utility of a proposed correction rather than indiscriminately replacing the source prediction.

\section{Continual Target Update and Algorithm}
\label{app:statistics_algori}

\vspace{-2mm}
\subsection{Continual Target Statistics}
\label{app:continual_statistics}

Our target-side estimator is maintained through compact sufficient statistics
rather than explicit replay of preceding target representations. At time
step $t$, the state constructed from
$\mathcal H_t=\bigcup_{\tau<t}\mathcal X_\tau$
is used to estimate $\hat{\mathbf q}_t^H$ in
Sec.~\ref{sec:gain_estimation}. After the current mini-batch is predicted,
the statistics are updated once and carried forward to the next time step.
This section details the resulting causal recursion.

\vspace{-1mm}
\paragraph{Causal Update Protocol.}
Let
$\mathcal X_t=\{\mathbf x_{t,b}\}_{b=1}^{B_t}$
denote the target mini-batch at time step $t$, with frozen representations
$\mathbf z_{t,b}=\phi_\theta(\mathbf x_{t,b})$.
The predictive state available before processing $\mathcal X_t$ is
\begin{equation}
\mathcal S_{t-1}
=
\left(
\{\mathbf m_{t-1,k},\kappa_{t-1,k},\pi_{t-1,k}\}_{k=1}^{K},
\Sigma_{t-1}
\right).
\label{eq:app_target_state}
\end{equation}
This state remains fixed while predicting all samples in $\mathcal X_t$. After obtaining the gain-controlled predictions $\mathbf p^\star_{t,b}$, we assign each sample to class $k$ with reliability-weighted responsibility
\begin{equation}
    \qquad
    \omega_{t,b,k}
    =
    \zeta_{t,b}p^\star_{t,b,k}, 
    \qquad
    \text{where }
     \zeta_{t,b}
    =
    s_{t,b,\hat y^\star_{t,b}},
    \hat y^\star_{t,b}
    =
    \arg\max_k p^\star_{t,b,k} 
   .
    \label{eq:fractional_class_weight}
\end{equation}
The resulting weights satisfy
$\sum_{k=1}^{K}\omega_{t,b,k}=\zeta_{t,b}$.
Here, $\mathbf p^\star_{t,b}$ determines the class-wise allocation, while $\zeta_{t,b}$ measures how strongly the frozen source model supports the adapted prediction. Hence, corrections weakly supported by the source contribute less to the accumulated target statistics.

All predictions in $\mathcal X_t$ are computed before any state update. Therefore, each prediction depends only on preceding target observations and the current sample, while the current mini-batch can affect only subsequent adaptation. This batch-causal predict-then-update protocol avoids within-batch feedback and limits the propagation of unreliable corrections through the continual target state.

\vspace{-2mm}
\paragraph{Cold-Start Initialization.}
When $\mathcal H_1\!\!=\!\!\varnothing$, no historical target evidence is available. As specified in Sec.~\ref{sec:gain_estimation}, we set $\hat{\mathbf q}_1^H\!=\!\mathbf s_1$, which yields the conservative choice $\lambda_1^\star\!=\!0$ and $\mathbf p_{1,b}^\star\!=\!\mathbf s_{1,b}$. We initialize
\begin{equation}
n_{0,k}=0,\qquad
\mathbf U_{0,k}=\mathbf 0,\qquad
\mathbf V_{0,k}=\mathbf 0,\qquad
Q_{0,k}=0,
\label{eq:app_statistics_initialization}
\end{equation}
where $n_{t,k}$ denotes the accumulated reliability-weighted class support, $\mathbf U_{t,k}$ and $\mathbf V_{t,k}$ are the corresponding weighted first- and second-moment statistics, and $Q_{t,k}$ accumulates squared weights for covariance estimation. Together with $\mathbf m_{0,k}=\mathbf c_k$, $\kappa_{0,k}=\kappa_0$, and $\Sigma_0=\mathbf I_D$, these quantities initialize the target state. The source predictions of $\mathcal X_1$ then provide the initial reliability-weighted assignments for subsequent target-state updates.

\vspace{-2mm}
\paragraph{Effective Class Support and Center Update.}
For class $k$, the current mini-batch contributes reliability-weighted soft support $\Delta n_{t,k}=\sum_{b=1}^{B_t}\omega_{t,b,k}$. We accumulate this support as
\begin{equation}
n_{t,k}
=n_{t-1,k}
+
\Delta n_{t,k},
\qquad
\kappa_{t,k}
=\kappa_0+n_{t,k},
\label{eq:app_kappa_update}
\end{equation}
where $\kappa_0$ controls the strength of the source-centered prior and $\kappa_{t,k}$ denotes the resulting effective class support. We further maintain the responsibility-weighted first moment
\vspace{-2mm}
\begin{equation}
\mathbf U_{t,k}
=\mathbf U_{t-1,k}
+
\sum\nolimits_{b=1}^{B_t}
\omega_{t,b,k}\mathbf z_{t,b},
\label{eq:app_first_moment}
\end{equation}
from which the source-anchored target center is recovered as
\begin{equation}
\mathbf m_{t,k}
=\frac{
\kappa_0\mathbf c_k+\mathbf U_{t,k}
}{
\kappa_0+n_{t,k}
}
=
\frac{
\kappa_0\mathbf c_k+\mathbf U_{t,k}
}{
\kappa_{t,k}
}.
\label{eq:app_center_update}
\end{equation}
Thus, the target center is updated from accumulated reliability-weighted evidence, while the source prior stabilizes the estimate when target support is limited.

\vspace{-2mm}
\paragraph{Effective-Support Shared Covariance Update.}
We instantiate the shared covariance in Sec.~\ref{sec:gain_estimation} with a diagonal estimator. In addition to the first moment, we maintain the responsibility-weighted second moment
\vspace{-2mm}
\begin{equation}
\mathbf V_{t,k}
=\mathbf V_{t-1,k}
+
\sum\nolimits_{b=1}^{B_t}
\omega_{t,b,k}\mathbf z_{t,b}^{\odot2},
\qquad
\mathbf V_{0,k}=\mathbf0.
\label{eq:app_second_moment}
\end{equation}
For $n_{t,k}\!>\!0$, the corresponding target empirical center is
$\bar{\mathbf z}_{t,k}\!\!=\!\!\frac{\mathbf U_{t,k}}{n_{t,k}}$, yielding the within-class scatter
\begin{equation}
\mathbf R_{t,k}
=\mathbf V_{t,k}-
\mathbf U_{t,k}^{\odot2}/n_{t,k}.
\label{eq:app_class_scatter}
\end{equation}
Unlike the source-anchored center $\mathbf m_{t,k}$ used for classification, $\bar{\mathbf z}_{t,k}$ is estimated solely from target evidence, isolating within-class target dispersion from source--target center shift.

Raw assignment mass does not directly quantify the statistical support available for covariance estimation. We therefore additionally maintain the squared effective-weight mass
\begin{equation}
Q_{t,k}
=Q_{t-1,k}
+
\sum\nolimits_{b=1}^{B_t}
\omega_{t,b,k}^{2},
\qquad
Q_{0,k}=0,
\label{eq:app_squared_weight}
\end{equation}
and define the weighted residual degrees of freedom
\begin{equation}
\nu_{t,k}
=n_{t,k}-
\frac{Q_{t,k}}{n_{t,k}},
\qquad
n_{t,k}>0.
\label{eq:app_effective_dof}
\end{equation}
For $n$ unit-weight hard assignments, $\nu_{t,k}=n-1$. In particular, a singleton gives $\nu_{t,k}=0$ and therefore does not spuriously increase the statistical support of the covariance estimate.

Pooling the within-class scatter across classes gives
\begin{equation}
\nu_t=
\sum\nolimits_{k:n_{t,k}>0}\nu_{t,k},
\qquad
\hat{\mathbf v}_t^T=
\frac{
\sum_{k:n_{t,k}>0}\mathbf R_{t,k}
}{
\nu_t
},
\qquad
\nu_t>0.
\label{eq:app_pooled_variance}
\end{equation}
We initialize the shared covariance isotropically with
$\mathbf v_0=\mathbf1$, equivalently $\Sigma_0=\mathbf I_D$, and shrink the empirical target variance toward this initialization:
\begin{equation}
\mathbf v_t=
\begin{cases}
\dfrac{
\kappa_0\mathbf v_0
+
\nu_t\hat{\mathbf v}_t^T
}{
\kappa_0+\nu_t
},
& \nu_t>0,
\\
\mathbf v_0,
& \nu_t=0.
\end{cases}
\label{eq:app_covariance_shrinkage}
\end{equation}
The resulting shared covariance is
\begin{equation}
\Sigma_t
=
\operatorname{Diag}
\left(
\max\{
\mathbf v_t,
\epsilon_\Sigma\mathbf 1
\}
\right),
\label{eq:app_covariance_update}
\end{equation}
where the maximum is applied element-wise and $\epsilon_\Sigma$ is a fixed numerical variance floor. We reuse $\kappa_0$ as the shrinkage pseudo-support to avoid introducing an additional tuning parameter. The effective degrees of freedom prevent weak or singleton class support from prematurely overriding the isotropic initialization.

\paragraph{Reliability-Balanced Historical Class Prior.}
As defined in Eq.~\ref{eq:app_kappa_update}, $n_{t,k}$ denotes the retained target support accumulated from the reliability-weighted class weights $\omega_{t,b,k}=\zeta_{t,b}p^\star_{t,b,k}$, providing the effective support used in the Gaussian target-state update. For the historical class prior, we use this reliability-weighted support together with the corresponding predictive class mass obtained from the same adapted predictions. Specifically, we maintain the accumulated predictive class mass 
$\hat\kappa_{t,k}=\hat n_{t,k} +\kappa_0$ with $\hat n_{t,k}=\hat n_{t-1,k}+\sum_{b=1}^{B_t} p^\star_{t,b,k}$ . 
We then construct the historical class prior by combining reliability-normalized support with inverse-support balancing:
\begin{equation}
    \pi_{t,k}
    \propto
    \underbrace{
    \frac{n_{t,k}+\kappa_0}
         {\hat n_{t,k}+\kappa_0}
    }_{\text{reliability-normalized support}}
    \cdot
    \underbrace{
    \frac{1}
         {\hat n_{t,k}+\kappa_0}
    }_{\text{class balancing}}= \frac{\kappa_{t,k}}{\hat\kappa_{t,k}^2}
    ,
    \label{eq:historical_prior}
\end{equation}
where the proportionality is normalized across classes. The first factor measures how strongly the accumulated predictive mass for class $k$ is supported by the reliability-weighted target statistics, while the second prevents frequently assigned classes from dominating the historical prior. At cold start, $n_{0,k}=\hat n_{0,k}=0$ for all classes, so both factors are class-independent and the normalized prior reduces to $\pi_{0,k}=1/K$. The prior is updated only after the current mini-batch has been predicted; hence, $\pi_{t-1,k}$ is used for samples in $\mathcal X_t$, while the updated $\pi_{t,k}$ is carried forward to time $t+1$.


\paragraph{Carried-Forward State and Efficiency.}
After processing $\mathcal X_t$, the updated predictive state
\begin{equation}
\mathcal S_t
=
\left(
\{\mathbf m_{t,k},\kappa_{t,k},\pi_{t,k}\}_{k=1}^{K},
\Sigma_t
\right)
\label{eq:app_updated_target_state}
\end{equation}
is used only from time step $t+1$ onward.
The quantities
$\{n_{t,k},\hat n_{t,k},\mathbf U_{t,k},\mathbf V_{t,k},Q_{t,k}\}_{k=1}^{K}$
are auxiliary sufficient statistics used only for recursive state updates.
Consequently, the method neither stores preceding target representations
nor revisits earlier samples. With a diagonal shared covariance, the
maintained statistics require $O(KD)$ memory independent of stream length,
and all updates consist only of responsibility-weighted vector operations
without backpropagation or replay.

\subsection{Overall Algorithm}
\label{app:algorithm}
Algorithm~\ref{alg:ours} summarizes the causal implementation of \ours. The pre-$t$ target state is fixed while predicting the entire mini-batch, and all sufficient statistics are updated only after the corresponding predictions have been obtained. The detailed recursions are given in Sec.~\ref{app:continual_statistics}.

\begin{algorithm}[t]
\caption{\textbf{G}ain-\textbf{A}ware \textbf{IN}tervention (GAIN)}
\label{alg:ours}
\begin{algorithmic}[1]

\Require
Unlabeled target stream $\{\mathcal X_t\}_{t=1}^{T}$;
frozen source model $f_\theta$ with feature extractor $\phi_\theta$;
source prototypes $\{\mathbf c_k\}_{k=1}^{K}$; source-centered prior strength $\kappa_0$.


\Ensure
Adapted predictions $\{\mathbf p_{t,b}^{\star}\}$.

\State \textbf{Initialize:} $\mathbf{m}_{0,k}\!=\!\mathbf{c}_k$, $\kappa_{0,k}\!=\!\kappa_0$, $\!\hat{\kappa}_{0,k}\!=\!\kappa_0$, $\pi_{0,k}\!=\!1/K$, and $\mathbf{\Sigma}_0\!=\!\mathbf{I}_D$.




\For{$t=1,\ldots,T$}
    \State Receive
    $\mathcal X_t=\{\mathbf x_{t,b}\}_{b=1}^{B_t}$
    and freeze the pre-$t$ state $\mathcal S_{t-1}$
    in Eq.~\ref{eq:app_target_state}.

    \State Compute features
    $\mathbf z_{t,b}=\phi_\theta(\mathbf x_{t,b})$
    and source predictions $\mathbf s_{t,b}$ using Eq.~\ref{eq:source_prediction}.
    
    \If{$t=1$}
        \State $\hat q^H_{t,b}\leftarrow s_{t,b}$,
        $\bar q^H_{t,b}\leftarrow s_{t,b}$,
        $\lambda^\star_{t,b}\leftarrow0$,
        $p^\star_{t,b}\leftarrow s_{t,b}$.
    \Else
        \For{$b=1,\ldots,B_t$}

            \State Construct the target proposal
            $\hat{\mathbf q}_{t,b}^{H}$
            from $d_{t,b,k}$ and $\hat\ell_{t,b,k}^{H}$
            using
            Eq.~\ref{eq:target_score} and Eq.~\ref{eq:target_posterior}.

            \State Form source-relative evidence
            $\hat\xi_{t,b,k}^{s}
            \gets
            \log
            (\hat q_{t,b,k}^{H}/s_{t,b,k})$.

            \State Construct
            $\bar{\mathbf q}_{t,b}^{H}$
            and evaluate
            $G_{t,b}^{\mathrm{pp}}$
            using
            Eq.~\ref{eq:posterior_predictive_target} and Eq.~\ref{eq:posterior_predictive_gain}.

            \State Determine $\lambda_{t,b}^{\star}$ by Eq.~\ref{eq:app_optimal_lambda}.

            \State Obtain
            $\mathbf p_{t,b}^{\star}$
            from Eq.~\ref{eq:app_final_intervention}.
        \EndFor

    \EndIf

    \State
    Obtain $\omega_{t,b,k}$ by Eq.~\ref{eq:fractional_class_weight}.

    \State
    Update $\{n_{t,k},\kappa_{t,k},\mathbf m_{t,k}\}$
    using Eqs.~\ref{eq:app_kappa_update}--\ref{eq:app_center_update}.

    \State Update
    $\Sigma_t$ using Eqs.~\ref{eq:app_second_moment}--\ref{eq:app_covariance_update}.

    \State
    Update $\{\hat{\kappa}_{t,k},\pi_{t,k}\}$ using Eq.~\ref{eq:historical_prior}
    and carry $\mathcal S_t$ in Eq.~\ref{eq:app_updated_target_state} to time $t+1$.

\EndFor
\State \Return $\{\mathbf p_{t,b}^{\star}\}$.
\end{algorithmic}
\end{algorithm}
\vspace{-3mm}

\section{Detailed Experimental Setup}
\label{app:setup}



\paragraph{Datasets.}
We conduct our main continual adaptation experiments on ImageNet-C~\citep{imagenet-c}, which contains 15 corruption types at five severity levels: Gaussian noise, shot noise, impulse noise, defocus blur, glass blur, motion blur, zoom blur, snow, frost, fog, brightness, contrast, elastic transform, pixelate, and JPEG compression. Following established CTTA protocols~\citep{CoTTA, DPCore, REM}, we evaluate all corruptions at severity level~5 unless otherwise specified. For each corruption, we use a fixed set of 5,000 images following the RobustBench-based evaluation protocol~\citep{DPCore, ViDA}, yielding 75,000 samples per complete corruption cycle. The same per-corruption samples are used across all stream settings, which differ only in their temporal organization and repetition. Target labels are never accessed during adaptation and are used solely for evaluation.

To further evaluate the robustness and generalizability of \ours beyond ImageNet-C, we include additional evaluations on both CTTA and standard TTA settings. For continual adaptation, we consider ImageNet-3DCC~\citep{3DCC}, which contains 12 corruption types at five severity levels and introduces geometry-aware transformations that produce more realistic distribution shifts. We further extend the evaluation beyond CTTA to standard TTA on ImageNet-R~\citep{imagenet-r}, ImageNet-V2~\citep{imagenet-v}, and ImageNet-Sketch~\citep{imagenet-sketch}, assessing whether gain-guided intervention generalizes across diverse forms of domain shift.

\paragraph{Metrics.}
We report top-1 accuracy (Acc.) and expected calibration error (ECE)~\citep{ECE} to evaluate predictive performance and confidence calibration. For each target sample $i$, let $\mathbf p_i$ denote the predicted distribution, $\hat y_i = \arg\max_k p_{i,k}$ the predicted label, and $c_i = \max_k p_{i,k}$ the prediction confidence. Top-1 accuracy is
\vspace{-3mm}
\begin{equation}
\operatorname{Acc}
=\frac{1}{N}
\sum_{i=1}^{N}
\mathbb I[\hat y_i = y_i].
\end{equation}
ECE partitions predictions into $M\!=\!20$ confidence bins $\{\mathcal B_m\}_{m=1}^{M}$ and measures the discrepancy between empirical accuracy and mean confidence:
\begin{equation}
\operatorname{ECE}
=\sum_{m=1}^{M}
\frac{|\mathcal B_m|}{N}
\left|
\operatorname{acc}(\mathcal B_m)
-\operatorname{conf}(\mathcal B_m)
\right|,
\end{equation}

\vspace{-3mm}
where $\operatorname{acc}(\mathcal B_m)\!=\!|\mathcal B_m|^{-1}\sum_{i\in\mathcal B_m}\mathbb I[\hat y_i\!=\!y_i]$ and $\operatorname{conf}(\mathcal B_m) = |\mathcal B_m|^{-1}\sum_{i\in\mathcal B_m} c_i$. 
For CSC and CDC, ECE is computed separately for each corruption and then averaged over corruption types, with the same protocol used within each LHA cycle. For MDS, ECE is computed over the pooled mixed stream at each severity level. We report both metrics in percentage points, with higher Acc. and lower ECE indicating better performance.
For ablation studies, we additionally report negative log-likelihood (NLL), $\operatorname{NLL}=-N^{-1}\sum_i\log p_{i,y_i}$, where lower is better.

\begin{table}[t]
\vspace{-5mm}
\caption{
\textbf{CDC results on ImageNet-C.} Accuracy (Acc., \%) and expected calibration error (ECE, \%) with ViT-Base at severity level~5.
BP-free denotes backpropagation-free adaptation.
Bold indicates the best results; Source is shown for reference only.
Our results are averaged over five runs.
}
\label{tab:imagenetc-cdc}
\vspace{-3mm}
\begin{center}
\begingroup
\small
\setlength{\tabcolsep}{1.6pt}
\renewcommand{\arraystretch}{1.0}

\resizebox{\linewidth}{!}{%
\begin{tabular}{lcc|*{15}{c}|c}
\toprule
\multirow{2}{*}{Method} &
\multirow{2}{*}{\shortstack{BP-free}} &
\multirow{2}{*}{Metric} &
\multicolumn{3}{c}{Noise} &
\multicolumn{4}{c}{Blur} &
\multicolumn{4}{c}{Weather} &
\multicolumn{4}{c}{Digital} &
\multicolumn{1}{|c}{\multirow{2}{*}{\textbf{Avg.}}} \\

\cmidrule(lr){4-6}
\cmidrule(lr){7-10}
\cmidrule(lr){11-14}
\cmidrule(lr){15-18}

& & &
\multicolumn{1}{c}{Gauss.} &
\multicolumn{1}{c}{Shot} &
\multicolumn{1}{c}{Impu.} &
\multicolumn{1}{c}{Defo.} &
\multicolumn{1}{c}{Glas.} &
\multicolumn{1}{c}{Moti.} &
\multicolumn{1}{c}{Zoom} &
\multicolumn{1}{c}{Snow} &
\multicolumn{1}{c}{Fros.} &
\multicolumn{1}{c}{Fog} &
\multicolumn{1}{c}{Brig.} &
\multicolumn{1}{c}{Cont.} &
\multicolumn{1}{c}{Elas.} &
\multicolumn{1}{c}{Pix.} &
\multicolumn{1}{c}{JPEG} &
\multicolumn{1}{|c}{} \\
\midrule

\rowcolor{sourcecolor}
& & Acc. \(\uparrow\)
& 47.0 & 48.2 & 47.9 & 31.5 & 21.2 & 41.5 & 36.7
& 50.1 & 45.8 & 42.3 & 73.6 & 8.6 & 42.5 & 62.0
& 63.8 & 44.2 \\
\rowcolor{sourcecolor}
\multirow{-2}{*}{Source}
& \multirow{-2}{*}{--}
& ECE \(\downarrow\)
& 3.6 & 4.1 & 3.7 & 4.3 & 5.4 & 3.9 & 9.1
& 2.3 & 4.9 & 17.4 & 3.1 & 3.9 & 9.0 & 3.3
& 2.7 & 5.4 \\
\midrule

\multirow{2}{*}{CoTTA~(\hyperlink{cite.CoTTA}{CVPR 2022})}
& \multirow{2}{*}{\xmark}
& Acc. \(\uparrow\)
& 46.1 & 47.4 & 47.5 & 33.3 & 22.3 & 42.7 & 38.6
& 50.3 & 46.1 & 44.2 & 74.0 & 6.6 & 43.3 & 62.4
& 64.6 & 44.6 \\
& & ECE \(\downarrow\)
& 7.0 & 6.9 & 10.3 & \textbf{4.1} & 12.1 & 8.4 & 17.3
& 5.5 & 8.9 & 10.0 & 5.2 & \textbf{2.2} & 12.8 & 5.7
& \textbf{4.1} & 8.0 \\
\addlinespace[1.8pt]

\multirow{2}{*}{SAR~(\hyperlink{cite.SAR}{ICLR 2023})}
& \multirow{2}{*}{\xmark}
& Acc. \(\uparrow\)
& 53.7 & 57.5 & 56.5 & 51.4 & 45.7 & 56.8 & 48.1
& 61.2 & 57.9 & 53.7 & 77.3 & 41.6 & 50.8 & 66.7
& 66.5 & 56.4 \\
& & ECE \(\downarrow\)
& 5.2 & 8.4 & 8.8 & 10.4 & 11.2 & 9.0 & 13.4
& 8.9 & 9.7 & \textbf{7.2} & \textbf{4.4} & 11.6 & 10.1 & \textbf{5.2}
& 4.9 & 8.6 \\
\addlinespace[1.8pt]

\multirow{2}{*}{ROID~(\hyperlink{cite.ROID}{WACV 2024})}
& \multirow{2}{*}{\xmark}
& Acc. \(\uparrow\)
& 56.5 & 58.8 & 56.5 & 50.9 & 47.4 & 54.5 & 52.2
& 62.5 & 59.3 & 60.9 & 78.2 & 43.9 & 56.3 & 67.0
& 67.7 & 58.2 \\
& & ECE \(\downarrow\)
& 56.4 & 58.7 & 56.4 & 50.8 & 47.3 & 54.4 & 52.1
& 62.4 & 59.2 & 60.8 & 78.1 & 43.8 & 56.2 & 66.9
& 67.6 & 58.1 \\
\addlinespace[1.8pt]

\multirow{2}{*}{ViDA~(\hyperlink{cite.ViDA}{ICLR 2024})}
& \multirow{2}{*}{\xmark}
& Acc. \(\uparrow\)
& 53.7 & 56.6 & 55.3 & 51.4 & 43.7 & 54.4 & 50.2
& 61.4 & 57.1 & 59.5 & 76.1 & 40.0 & 49.1 & 67.7
& 67.2 & 56.2 \\
& & ECE \(\downarrow\)
& 11.7 & 10.6 & 15.2 & 15.8 & 19.7 & 11.9 & 20.2
& 10.4 & 13.1 & 14.3 & 7.8 & 22.1 & 18.0 & 7.1
& 6.7 & 13.6 \\
\addlinespace[1.8pt]

\multirow{2}{*}{DeYO~(\hyperlink{cite.DeYO}{ICLR 2024})}
& \multirow{2}{*}{\xmark}
& Acc. \(\uparrow\)
& 55.8 & 58.8 & 57.0 & 50.9 & 47.9 & 54.7 & 48.6
& 60.8 & 59.1 & 61.4 & 77.3 & 41.9 & 53.4 & 67.4
& 68.6 & 57.6 \\
& & ECE \(\downarrow\)
& 9.1 & 7.8 & 10.8 & 12.8 & 13.7 & 10.1 & 16.5
& 9.7 & 9.8 & 11.4 & 6.1 & 13.8 & 12.1 & 6.2
& 5.3 & 10.3 \\
\addlinespace[1.8pt]

\multirow{2}{*}{AEA~(\hyperlink{cite.AEA}{ICLR 2025})}
& \multirow{2}{*}{\xmark}
& Acc. \(\uparrow\)
& 47.9 & 47.7 & 51.9 & 48.5 & 47.6 & 49.7 & 49.8
& 47.0 & 55.2 & 65.1 & 75.0 & 37.4 & 49.1 & 64.4
& 64.5 & 53.4 \\
& & ECE \(\downarrow\)
& 22.4 & 20.5 & 25.2 & 27.8 & 26.1 & 24.7 & 60.6
& 26.0 & 21.8 & 21.0 & 16.6 & 34.4 & 28.2 & 18.9
& 19.6 & 26.2 \\
\addlinespace[1.8pt]

\multirow{2}{*}{ReCAP~(\hyperlink{cite.ReCAP}{ICML 2025})}
& \multirow{2}{*}{\xmark}
& Acc. \(\uparrow\)
& 40.8 & 42.5 & 46.0 & 51.5 & 48.2 & 53.2 & 50.0
& 55.0 & 57.9 & 63.7 & 75.7 & 57.3 & 49.8 & 62.6
& 64.6 & 54.6 \\
& & ECE \(\downarrow\)
& 11.7 & 10.1 & 13.0 & 10.9 & 12.5 & 9.6 & 15.2
& 9.5 & 8.8 & 9.3 & 5.3 & 8.6 & 12.3 & 6.0
& 6.0 & 9.9 \\
\addlinespace[1.8pt]

\multirow{2}{*}{REM~(\hyperlink{cite.REM}{ICML 2025})}
& \multirow{2}{*}{\xmark}
& Acc. \(\uparrow\)
& 55.8 & 58.6 & 56.9 & 50.9 & \textbf{50.8} & 53.0 & \textbf{54.4}
& 61.8 & 63.0 & 63.8 & 77.1 & 46.9 & 57.7 & \textbf{69.9}
& 69.3 & 59.3 \\
& & ECE \(\downarrow\)
& 7.6 & 6.9 & 8.2 & 12.2 & 12.4 & 10.2 & 12.2
& 8.5 & \textbf{7.4} & 8.6 & 4.9 & 11.9 & 10.6 & 5.3
& 4.6 & 8.8 \\
\addlinespace[1.8pt]

\multirow{2}{*}{DPCore~(\hyperlink{cite.DPCore}{ICML 2025})}
& \multirow{2}{*}{\xmark}
& Acc. \(\uparrow\)
& 55.8 & 58.0 & 57.3 & 47.2 & 46.0 & 53.1 & 52.4
& 61.9 & 64.0 & 60.0 & 76.9 & 48.8 & 53.4 & 69.7
& \textbf{70.6} & 58.3 \\
& & ECE \(\downarrow\)
& 10.3 & 10.7 & 10.4 & 7.8 & 6.5 & 10.0 & 8.7
& 12.5 & 12.9 & 10.2 & 12.6 & 8.7 & 9.3 & 12.9
& 12.9 & 10.4 \\
\addlinespace[1.8pt]

\multirow{2}{*}{PAID~(\hyperlink{cite.PAID}{NeurIPS 2025})}
& \multirow{2}{*}{\xmark}
& Acc. \(\uparrow\)
& 49.8 & 54.8 & 49.9 & 45.1 & 48.5 & 49.4 & 51.0
& 61.5 & 59.9 & 55.8 & 71.1 & 44.7 & \textbf{57.9} & 65.5
& 64.6 & 55.3 \\
& & ECE \(\downarrow\)
& 10.9 & 9.1 & 12.0 & 13.3 & 12.2 & 11.9 & 11.7
& 6.6 & 8.3 & 10.7 & 4.6 & 14.2 & 8.1 & 5.4
& 5.4 & 9.6 \\
\addlinespace[1.8pt]

\multirow{2}{*}{DOTA~(\hyperlink{cite.DOTA}{NeurIPS 2025})}
& \multirow{2}{*}{\cmark}
& Acc. \(\uparrow\)
& 59.1 & 59.2 & 60.1 & 50.2 & 38.1 & 56.1 & 46.9
& 64.1 & 64.8 & 66.3 & 78.3 & 32.5 & 47.4 & 68.5
& 69.4 & 57.4 \\
& & ECE \(\downarrow\)
& 37.0 & 36.9 & 36.7 & 44.9 & 55.9 & 40.4 & 47.9
& 33.0 & 31.5 & 25.9 & 20.3 & 56.8 & 48.0 & 28.6
& 27.3 & 38.1 \\
\addlinespace[1.8pt]

\multirow{2}{*}{NEO~(\hyperlink{cite.NEO}{ICLR 2026})}
& \multirow{2}{*}{\cmark}
& Acc. \(\uparrow\)
& 56.2 & 56.5 & 56.9 & 46.9 & 36.0 & 52.6 & 45.4
& 62.9 & 63.9 & 68.9 & 78.2 & 36.4 & 45.8 & 67.0
& 67.1 & 56.1 \\
& & ECE \(\downarrow\)
& 10.0 & 6.7 & 9.1 & 6.8 & \textbf{5.3} & \textbf{4.0} & 4.9
& 5.4 & 20.8 & 51.5 & 9.0 & 23.8 & 5.9 & 5.7
& 6.7 & 11.7 \\

\midrule
\rowcolor{mycolor}
& & Acc. \(\uparrow\)
& \bval{59.9}{0.73} & \bval{59.7}{0.76} & \bval{60.8}{1.19} & \bval{54.4}{0.64} & \val{43.9}{0.98} & \bval{58.7}{0.39} & \val{51.6}{0.91} & \bval{66.1}{0.46} & \bval{67.0}{0.19} & \bval{72.2}{0.46} & \bval{78.4}{0.25} & \bval{61.7}{0.61} & \val{53.1}{2.64} & \val{69.2}{0.64} & \val{70.3}{0.49} & \bval{61.8}{0.28} \\
\rowcolor{mycolor}
\multirow{-2}{*}{\ours (Ours)}
& \multirow{-2}{*}{\cmark}
& ECE \(\downarrow\)
& \bval{5.2}{0.52} & \bval{5.7}{0.70} & \bval{5.9}{1.12} & \val{5.9}{0.35} & \val{5.6}{0.31} & \val{5.4}{0.38} & \bval{4.8}{0.52} & \bval{5.2}{0.36} & \val{8.8}{0.52} & \val{7.3}{3.06} & \val{5.1}{0.75} & \val{10.9}{0.95} & \bval{5.8}{0.40} & \val{5.9}{1.04} & \val{4.9}{0.46} & \bval{6.2}{0.23} \\

\bottomrule
\end{tabular}%
}
\vspace{-3mm}
\endgroup
\end{center}
\end{table}

\paragraph{Considered Settings.}
We evaluate \ours under four complementary forms of continual distribution shift.
\textbf{\textit{Continual Structured Change (CSC)}} follows the conventional CTTA protocol~\citep{CoTTA}, where the model encounters all 15 ImageNet-C corruptions sequentially at severity level 5, with 5,000 consecutive samples per corruption and no reset across domain transitions. We use the standard corruption order adopted in prior CTTA work.
\textbf{\textit{Continual Dynamic Change (CDC)}} follows DPCore~\citep{DPCore} and reorganizes the same corruption domains into a less structured stream, where domains recur with non-uniform durations and frequencies. We use the released CDC construction with a Dirichlet concentration parameter of $\delta=1$.
\textbf{\textit{Mixed-Domain Shift (MDS)}} follows the mixed-domain evaluation protocol used in Wild TTA~\citep{SAR, ReCAP}, where samples from multiple corruption domains are interleaved within the same test stream rather than appearing in locally homogeneous domain segments. This setting introduces concurrent domain heterogeneity and tests whether adaptation remains reliable when accumulated target statistics reflect a mixture of shifts.
Finally, \textbf{\textit{Long-Horizon Adaptation (LHA)}} evaluates stability under repeated exposure to previously observed shifts. Following prior repeating-domain protocols~\citep{ViDA, DPCore}, we repeat the complete 15-corruption stream for 10 rounds (R1--R10) without resetting either the model or the target state. No corruption identity or domain boundary is provided to \ours in any setting.

\vspace{-2mm}
\paragraph{Implementation Details.}
We use an ImageNet-pretrained ViT-B/16  as the source model and set the test-time mini-batch size to 64 for all main experiments. The feature representation $\mathbf z=\phi_\theta(\mathbf x)$ and source prediction $\mathbf s\!=\!f_\theta(\mathbf x)$ are obtained from the frozen source network defined in \Cref{sec:preliminaries}; neither the backbone nor the classifier is updated during adaptation. \ours maintains only the recursive target statistics described in \Cref{app:continual_statistics}. We instantiate the shared covariance with a diagonal estimator, initialize $\Sigma_0=\mathbf I_D$, and set the source-centered prior strength to $\kappa_0\!=\!3$.
The intervention coefficient $\lambda_{t,b}^{\star}$ is determined independently for each test sample by the endpoint conditions in Eq.~\ref{eq:app_optimal_lambda}, with the interior case solved by one-dimensional bisection rather than treated as a tuned mixing coefficient. Unless otherwise specified, all hyperparameters and numerical tolerances are fixed across CSC, CDC, MDS, and long-horizon evaluation.
For consistency, we compute ECE from the final outputs returned by each method's official implementation.
Experiments are implemented in PyTorch and conducted on a single NVIDIA RTX A6000 GPU.

\begin{table*}[t]
\vspace{-5mm}
\caption{
\textbf{MDS results on ImageNet-C.} Accuracy (Acc.,\%) and expected calibration error (ECE,\%) across corruption severity levels 5--1.
BP-free denotes backpropagation-free adaptation. 
Bold indicates the best results; Source is shown for reference only.
Our results are averaged over five runs.
}
\label{tab:imagenetc_mds}
\vspace{-3mm}
\begin{center}
\begingroup
\small
\setlength{\tabcolsep}{6.5pt}
\renewcommand{\arraystretch}{0.8}
\resizebox{0.98\linewidth}{!}{%
\begin{tabular}{lcc|ccccc|c}
\toprule
Method & BP-free & Metric & Level 5 & Level 4 & Level 3 & Level 2 & Level 1 & \textbf{Avg.} \\
\midrule


\rowcolor{sourcecolor}
& & Acc. \(\uparrow\)
& 44.2 & 55.4  & 68.3  & 69.2  & 74.8  & 62.4  \\
\rowcolor{sourcecolor}
\multirow{-2}{*}{Source}
& \multirow{-2}{*}{--}
& ECE \(\downarrow\)
& 3.4 & 4.3 & 4.2 & 3.3 & 2.8 & 3.6 \\

\midrule

\multirow{2}{*}{CoTTA~(\hyperlink{cite.CoTTA}{CVPR 2022})}
& \multirow{2}{*}{\xmark}
& Acc. \(\uparrow\)
& 50.3 & 62.0 & 69.4 & 73.9 & 78.2 & 66.8 \\
& & ECE \(\downarrow\)
& \textbf{5.4} & \textbf{4.6} & 3.9 & 3.9 & 3.4 & 4.2 \\
\addlinespace[1.8pt]

\multirow{2}{*}{SAR~(\hyperlink{cite.SAR}{ICLR 2023})}
& \multirow{2}{*}{\xmark}
& Acc. \(\uparrow\)
& 56.1 & 64.9 & 70.9 & 74.0 & 78.3 & 68.8 \\
& & ECE \(\downarrow\)
& 8.5 & 5.9 & 4.4 & 3.8 & 3.2 & 5.2 \\
\addlinespace[1.8pt]

\multirow{2}{*}{ROID~(\hyperlink{cite.ROID}{WACV 2024})}
& \multirow{2}{*}{\xmark}
& Acc. \(\uparrow\)
& 56.9 & 65.5 & 71.2 & 74.4 & 78.5 & 69.3 \\
& & ECE \(\downarrow\)
& 56.8 & 65.4 & 71.1 & 74.3 & 78.4 & 69.2 \\
\addlinespace[1.8pt]

\multirow{2}{*}{DeYO~(\hyperlink{cite.DeYO}{ICLR 2024})}
& \multirow{2}{*}{\xmark}
& Acc. \(\uparrow\)
& 55.2 & 64.0 & 69.8 & 72.8 & 77.4 & 67.8 \\
& & ECE \(\downarrow\)
& 10.8 & 7.6 & 5.9 & 5.1 & 4.0 & 6.7 \\
\addlinespace[1.8pt]

\multirow{2}{*}{AEA~(\hyperlink{cite.AEA}{ICLR 2025})}
& \multirow{2}{*}{\xmark}
& Acc. \(\uparrow\)
& 52.4 & 60.2 & 67.8 & 52.2 & 55.8 & 57.7 \\
& & ECE \(\downarrow\)
& 19.4 & 16.3 & 12.1 & 33.5 & 27.6 & 21.8 \\
\addlinespace[1.8pt]

\multirow{2}{*}{ReCAP~(\hyperlink{cite.ReCAP}{ICML 2025})}
& \multirow{2}{*}{\xmark}
& Acc. \(\uparrow\)
& 56.7 & 65.0 & 70.6 & 73.4 & 77.7 & 68.7 \\
& & ECE \(\downarrow\)
& 11.0 & 8.2 & 6.5 & 5.9 & 4.8 & 7.3 \\
\addlinespace[1.8pt]

\multirow{2}{*}{REM~(\hyperlink{cite.REM}{ICML 2025})}
& \multirow{2}{*}{\xmark}
& Acc. \(\uparrow\)
& \textbf{59.5} & 66.8 & 72.1 & 74.8 & 78.6 & 70.4 \\
& & ECE \(\downarrow\)
& 8.6 & 6.6 & 5.3 & 4.9 & 4.2 & 5.9 \\
\addlinespace[1.8pt]

\multirow{2}{*}{DPCore~(\hyperlink{cite.DPCore}{ICML 2025})}
& \multirow{2}{*}{\xmark}
& Acc. \(\uparrow\)
& 54.3 & 64.5 & 69.8 & 75.0 & 78.6 & 68.4 \\
& & ECE \(\downarrow\)
& 8.1 & 10.3 & 10.3 & 11.3 & 10.7 & 10.1 \\
\addlinespace[1.8pt]

\multirow{2}{*}{PAID~(\hyperlink{cite.PAID}{NeurIPS 2025})}
& \multirow{2}{*}{\xmark}
& Acc. \(\uparrow\)
& 53.0 & 62.2 & 68.9 & 72.9 & 77.8 & 67.0 \\
& & ECE \(\downarrow\)
& 7.0 & 4.7 & \textbf{3.5} & \textbf{2.8} & \textbf{2.1} & \textbf{4.0} \\
\addlinespace[1.8pt]

\multirow{2}{*}{DOTA~(\hyperlink{cite.DOTA}{NeurIPS 2025})}
& \multirow{2}{*}{\cmark}
& Acc. \(\uparrow\)
& 55.6 & 65.9 & 71.8 & 75.3 & 79.2 & 69.6 \\
& & ECE \(\downarrow\)
& 37.9 & 29.4 & 24.4 & 21.5 & 18.3 & 26.3 \\
\addlinespace[1.8pt]

\multirow{2}{*}{NEO~(\hyperlink{cite.NEO}{ICLR 2026})}
& \multirow{2}{*}{\cmark}
& Acc. \(\uparrow\)
& 56.7 & 66.7 & 72.4 & 75.8 & 79.6 & 70.2 \\
& & ECE \(\downarrow\)
& 9.4 & 12.0 & 11.4 & 11.1 & 11.1 & 11.0 \\

\midrule
\rowcolor{mycolor}
& & Acc. \(\uparrow\)
& \(59.0 \pm 0.07\) & \textbf{68.2} \(\pm\) 0.03 & \textbf{73.6} \(\pm\) 0.04 & \textbf{76.6} \(\pm\) 0.06 & \textbf{80.1} \(\pm\) 0.03 & \textbf{71.5} \(\pm\) 0.02 \\
\rowcolor{mycolor}
\multirow{-2}{*}{\ours (Ours)}
& \multirow{-2}{*}{\cmark}
& ECE \(\downarrow\)
& \(8.1 \pm 0.07\) & \(6.6 \pm 0.15\) & \(5.6 \pm 0.03\) & \(5.1 \pm 0.06\) & \(5.2 \pm 0.04\) & \(6.1 \pm 0.04\) \\


\bottomrule
\end{tabular}%
}
\vspace{-3mm}
\endgroup
\end{center}
\end{table*}
\begin{table}[!t]
\caption{
\textbf{LHA results on ImageNet-C.} Accuracy (Acc., \%) and expected calibration error (ECE, \%) over 10 repeated corruption cycles with ViT-Base at severity~5. BP-free denotes backpropagation-free adaptation.
Bold indicates the best results; Source is shown for reference only.
}
\label{tab:imagenetc-lifelong}
\vspace{-3mm}
\begin{center}
\begingroup
\small
\setlength{\tabcolsep}{6.pt}
\renewcommand{\arraystretch}{0.8}
\resizebox{0.98\linewidth}{!}{%

\begin{tabular}{lcc|*{10}{c}|c}
\toprule
Method & \shortstack{BP-free} & Metric &
\multicolumn{1}{c}{R1} &
\multicolumn{1}{c}{R2} &
\multicolumn{1}{c}{R3} &
\multicolumn{1}{c}{R4} &
\multicolumn{1}{c}{R5} &
\multicolumn{1}{c}{R6} &
\multicolumn{1}{c}{R7} &
\multicolumn{1}{c}{R8} &
\multicolumn{1}{c}{R9} &
\multicolumn{1}{c}{R10} &
\multicolumn{1}{|c}{\textbf{Avg.}} \\
\midrule


\rowcolor{sourcecolor}
& & Acc. \(\uparrow\)
& 44.2 & 44.2 & 44.2 & 44.2 & 44.2 & 44.2
& 44.2 & 44.2 & 44.2 & 44.2 & 44.2 \\
\rowcolor{sourcecolor}
\multirow{-2}{*}{Source}
& \multirow{-2}{*}{--}
& ECE \(\downarrow\)
& 5.4 & 5.4 & 5.4 & 5.4 & 5.4 & 5.4
& 5.4 & 5.4 & 5.4 & 5.4 & 5.4 \\
\midrule

\multirow{2}{*}{CoTTA~(\hyperlink{cite.CoTTA}{CVPR 2022})}
& \multirow{2}{*}{\xmark}
& Acc. \(\uparrow\)
& 45.2 & 45.3 & 45.9 & 46.3 & 46.6 & 46.4
& 46.0 & 45.7 & 45.4 & 45.3 & 45.8 \\
& & ECE \(\downarrow\)
& 7.5 & 15.8 & 22.3 & 26.4 & 28.1 & 30.7
& 33.1 & 34.7 & 36.2 & 36.9 & 27.2 \\
\addlinespace[1.8pt]

\multirow{2}{*}{ViDA~(\hyperlink{cite.ViDA}{ICLR 2024})}
& \multirow{2}{*}{\xmark}
& Acc. \(\uparrow\)
& 56.6 & 57.3 & 57.3 & 57.4 & 57.5 & 57.5
& 57.3 & 54.1 & 54.8 & 54.9 & 56.5 \\
& & ECE \(\downarrow\)
& 14.7 & 21.0 & 24.1 & 26.3 & 28.0 & 29.3
& 29.9 & 31.1 & 31.9 & 32.9 & 26.9 \\
\addlinespace[1.8pt]

\multirow{2}{*}{AEA~(\hyperlink{cite.AEA}{ICLR 2025})}
& \multirow{2}{*}{\xmark}
& Acc. \(\uparrow\)
& 54.7 & 59.4 & 59.5 & 8.0 & 0.1 & 0.1
& 0.1 & 0.1 & 0.1 & 0.1 & 18.2 \\
& & ECE \(\downarrow\)
& 21.8 & 27.6 & 30.6 & 90.3 & 99.9 & 99.9
& 99.9 & 99.9 & 99.9 & 99.9 & 77.0 \\
\addlinespace[1.8pt]

\multirow{2}{*}{ReCAP~(\hyperlink{cite.ReCAP}{ICML 2025})}
& \multirow{2}{*}{\xmark}
& Acc. \(\uparrow\)
& 57.7 & 59.3 & 60.1 & 60.5 & 60.6 & 60.7
& 60.9 & 61.0 & 61.0 & 61.1 & 60.3 \\
& & ECE \(\downarrow\)
& 9.5 & 12.1 & 13.1 & 13.7 & 14.2 & 14.7
& 14.9 & 15.3 & 15.6 & 15.7 & 13.9 \\
\addlinespace[1.8pt]

\multirow{2}{*}{REM~(\hyperlink{cite.REM}{ICML 2025})}
& \multirow{2}{*}{\xmark}
& Acc. \(\uparrow\)

& 60.8	& 61.3	& 61.4	& 62.0	& 62.1	& 62.2	& 61.9	& 61.9	& 61.9	& 61.8	& 61.7 \\
& & ECE \(\downarrow\)
& 8.5	& 10.3	& 11.1	& 11.4	& 11.8	& 12.1	& 12.7	& 13.0	& 13.3	& 13.7	& 11.8 \\
\addlinespace[1.8pt]

\multirow{2}{*}{DPCore~(\hyperlink{cite.DPCore}{ICML 2025})}
& \multirow{2}{*}{\xmark}
& Acc. \(\uparrow\)
& 60.1 & 54.7 & 55.6 & 55.9	& 55.8 & 56.1 & 55.5 & 55.8	& 55.5 & 55.4 & 56.0\\
& & ECE \(\downarrow\)
& 8.2 & 8.3	& 7.6 & 7.9 & 7.9 & 8.3	& 9.5 & 9.9 & 9.8 & 9.9 & 8.7\\
\addlinespace[1.8pt]

\multirow{2}{*}{PAID~(\hyperlink{cite.PAID}{NeurIPS 2025})}
& \multirow{2}{*}{\xmark}
& Acc. \(\uparrow\)
& 57.8 & 53.9 & 50.9 & 48.6 & 46.7 & 45.2
& 43.6 & 42.3 & 41.3 & 40.2 & 47.0 \\
& & ECE \(\downarrow\)
& 8.6 & 10.5 & 11.2 & 12.1 & 13.0 & 13.9
& 14.7 & 15.7 & 16.2 & 16.9 & 13.3 \\
\addlinespace[1.8pt]

\multirow{2}{*}{DOTA~(\hyperlink{cite.DOTA}{NeurIPS 2025})}
& \multirow{2}{*}{\cmark}
& Acc. \(\uparrow\)
& 57.2 & 57.6 & 57.6 & 57.6 & 57.6 & 57.6
& 57.6 & 57.6 & 57.6 & 57.6 & 57.5 \\
& & ECE \(\downarrow\)
& 38.2 & 39.1 & 39.3 & 39.3 & 39.3 & 39.3
& 39.3 & 39.3 & 39.3 & 39.4 & 39.2 \\
\addlinespace[1.8pt]

\multirow{2}{*}{NEO~(\hyperlink{cite.NEO}{ICLR 2026})}
& \multirow{2}{*}{\cmark}
& Acc. \(\uparrow\)
& 56.0 & 56.1 & 56.1 & 56.1 & 56.1 & 56.1
& 56.1 & 56.1 & 56.1 & 56.1 & 56.1 \\
& & ECE \(\downarrow\)
& 11.7 & 11.7 & 11.6 & 11.6 & 11.6 & 11.7
& 11.7 & 11.7 & 11.7 & 11.7 & 11.7 \\

\midrule
\rowcolor{mycolor}
& & Acc. \(\uparrow\)
& \textbf{61.9} & \textbf{62.6} & \textbf{62.6} & \textbf{62.6} & \textbf{62.6} & \textbf{62.6}
& \textbf{62.5} & \textbf{62.5} & \textbf{62.5} & \textbf{62.5} & \textbf{62.5} \\
\rowcolor{mycolor}
\multirow{-2}{*}{\ours (Ours)}
& \multirow{-2}{*}{\cmark}
& ECE \(\downarrow\)
& \textbf{6.0} & \textbf{6.5} & \textbf{6.5} & \textbf{6.5} & \textbf{6.5} & \textbf{6.5}
& \textbf{6.5} & \textbf{6.5} & \textbf{6.5} & \textbf{6.5} & \textbf{6.4} \\

\bottomrule
\end{tabular}%
}
\endgroup
\end{center}
\vspace{-5mm}
\end{table}

\vspace{-2mm}
\section{Additional experimental results}
\label{app:experiment}
\vspace{-3mm}
\subsection{Detailed Experimental Results on ImageNet-C}

\vspace{-1mm}
\paragraph{Continual Dynamic Change (CDC).}
CDC introduces irregular and recurring domain shifts, making accumulated target statistics more susceptible to staleness. As shown in Table~\ref{tab:imagenetc-cdc}, \ours achieves the highest accuracy of 61.8\% with 6.2\% ECE, outperforming the strongest competing baseline REM by 2.5 accuracy points while reducing ECE by 2.6 points. Several methods degrade from CSC to CDC, including DPCore, ReCAP, and PAID, while ROID and DOTA exhibit severe miscalibration. These results show that \ours remains reliable under dynamic shifts by evaluating the source-relative utility of history-induced corrections before intervention.

\vspace{-3mm}
\paragraph{Mixed-Domain Shift (MDS).}
Table~\ref{tab:imagenetc_mds} reports results across corruption severity levels. \ours achieves the highest average accuracy of 71.5\%, outperforming REM (70.4\%) and NEO (70.2\%), and ranks first from Levels~4 to~1. It also maintains a low average ECE of 6.1\%, substantially below other high-accuracy BP-free methods such as NEO (11.0\%) and DOTA (26.3\%).
Overall, \ours maintains a favorable accuracy--calibration trade-off under heterogeneous mixed-domain shifts.

\vspace{-3mm}
\paragraph{Long-Horizon Adaptation (LHA).}
The long-horizon setting evaluates stability over ten repeated corruption cycles without resetting the adaptation state. As shown in Table~\ref{tab:imagenetc-lifelong}, \ours remains stable throughout all rounds, with accuracy increasing from 61.9\% to 62.5--62.6\% and ECE remaining around 6.5\%. In contrast, DPCore shows a marked accuracy decline and PAID progressively deteriorates, while REM and ReCAP become increasingly miscalibrated despite competitive accuracy. These results show that gain-guided intervention limits the reinforcement of unreliable history-induced corrections and maintains stable adaptation over long horizons.

\vspace{-3mm}
\subsection{Further Ablation Study and Analysis}
\begin{table}[t]
\vspace{-3mm}
\centering
\caption{\textbf{Ablation of the proposal–evaluator roles under CSC.} We compare direct target-side prediction, reversed proposal–evaluation roles, and the asymmetric design used by GAIN.}
\label{tab:ablation_role}
\vspace{-1mm}
\begingroup
\small
\setlength{\tabcolsep}{6pt}
\renewcommand{\arraystretch}{0.9}
\resizebox{0.9\columnwidth}{!}{
\begin{tabular}{lccc|ccc}
\toprule
Variant
& Proposal
& Gain evaluation
& Intervention
& Acc. $\uparrow$
& ECE $\downarrow$
& NLL $\downarrow$ \\
\midrule


Proposal as Prediction
& $\hat{\mathbf q}^{H}$
& --
& --
& 58.6
& 11.2
& 2.5
\\

Evaluator as Prediction
& $\bar{\mathbf q}^{H}$
& --
& --
& 58.9
& 12.8
& 2.4
\\


Swapped Roles
& $\bar{\mathbf q}^{H}$
& $\hat{\mathbf q}^{H}$
& $\tilde{\lambda}^{\star}$
& 61.7
& 8.0
& 2.0
\\

\midrule
\ours (Ours)
& $\hat{\mathbf q}^{H}$
& $\bar{\mathbf q}^{H}$
& $\lambda^{\star}$
& \textbf{61.9}
& \textbf{6.0}
& \textbf{1.9}
\\
\bottomrule
\end{tabular}}
\endgroup
\vspace{-2mm}
\end{table}

\vspace{-2mm}
\paragraph{Proposal--Evaluator Roles.}
Table~\ref{tab:ablation_role} examines the asymmetric roles of the posterior-mean proposal $\hat{\mathbf q}_t^H$ and posterior-predictive evaluator $\bar{\mathbf q}_t^H$. Directly using either distribution as the prediction performs substantially worse than \ours, showing that improved target-side estimation alone does not guarantee a reliable correction. In particular, using $\bar{\mathbf q}_t^H$ directly yields 58.9\% accuracy and 12.8\% ECE, supporting its role as an uncertainty-aware evaluator rather than a replacement prediction. Reversing the proposal and evaluator retains competitive accuracy with 61.7\% but degrades ECE from 6.0\% to 8.0\%. These results support the intended asymmetry of \ours: $\hat{\mathbf q}_t^H$ specifies the correction, while $\bar{\mathbf q}_t^H$ evaluates its source-relative utility.

\begin{figure*}[t]
    \centering
    \setlength{\abovecaptionskip}{1mm}
    \includegraphics[width=0.45\textwidth]{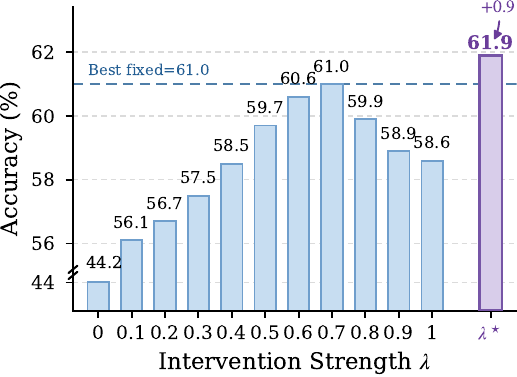}
    \hspace{2mm}
    \includegraphics[width=0.45\textwidth]{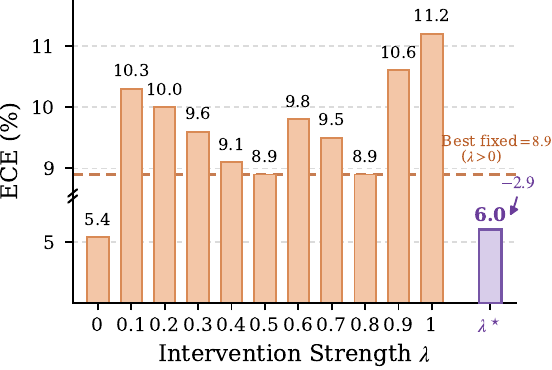}
    \caption{
    \textbf{Fixed vs. adaptive intervention strength.}
    We compare fixed $\lambda \in [0,1]$ with the proposed sample-wise adaptive $\lambda^\star$.
    No single nonzero fixed intervention matches the accuracy–calibration trade-off of \ours: the best fixed accuracy is 61.0\%, while the lowest nonzero fixed ECE is 8.9\%. The adaptive \(\lambda^\star\) achieves 61.9\% accuracy with 6.0\% ECE.
    }
    \label{fig:fixed_lambda}
    \vspace{-3mm}
\end{figure*}

\vspace{-3mm}
\paragraph{Fixed vs. Adaptive Intervention.}
Figure~\ref{fig:fixed_lambda} further compares \ours with fixed intervention strengths $\lambda\in[0,1]$. Increasing $\lambda$ initially improves accuracy by incorporating more target-side evidence, but aggressive correction eventually degrades both accuracy and calibration. No single nonzero fixed value achieves the same trade-off as the sample-wise adaptive intervention: the best fixed accuracy reaches 61.0\%, while the lowest nonzero fixed ECE remains 8.9\%. In contrast, \ours achieves 61.9\% accuracy with 6.0\% ECE, showing that correction strength should adapt to the estimated source-relative gain rather than remain fixed across samples.

\vspace{-3mm}
\paragraph{Continual Target-state Update.}
\begin{table*}[t]
\centering
\vspace{-5mm}
\caption{
\textbf{Ablation of continual target-state updates under CSC.}
RB denotes the reliability-balanced historical prior.
}
\vspace{-1mm}
\label{tab:ablation_state}
\small
\setlength{\tabcolsep}{6.pt}
\renewcommand{\arraystretch}{0.9}
\resizebox{0.9\linewidth}{!}{%
\begin{tabular}{lccc|ccc}
\toprule
\multirow{2}{*}{Variant}
& \multirow{2}{*}{Assignment}
& \multirow{2}{*}{Source Support $\boldsymbol{\zeta}$}
& \multirow{2}{*}{Class Prior}
& \multicolumn{3}{c}{CSC} \\
\cmidrule(lr){5-7}
& & & 
& Acc.$\uparrow$
& ECE$\downarrow$
& NLL$\downarrow$ \\
\midrule

\multicolumn{7}{c}{\textit{State Assignment}} \\
\midrule

\multirow{2}{*}{Source Assignment}
& \multirow{2}{*}{$\mathbf{s}_t$}
& --
& RB
& 58.5 & 13.7 & 2.2\\

& 
& \checkmark
& RB
& 60.8 & 7.2 & 2.0 \\

\multirow{2}{*}{Proposal Assignment}
& \multirow{2}{*}{$\hat{\mathbf q}_t^H$}
& --
& RB
& 59.6 & 12.2 & 2.2\\

&
& \checkmark
& RB
& 61.6 & 6.2 & 2.0 \\

\multirow{2}{*}{\textbf{\ours} (Ours)}
& \multirow{2}{*}{$\mathbf p_t^\star$}
& --
& RB
& 59.9 & 11.9 & 2.1 \\
 
& 
& \checkmark
& RB
& \textbf{61.9} & \textbf{6.0} & \textbf{1.9} \\

\midrule

\multicolumn{7}{c}{\textit{Historical Class Prior}} \\
\midrule

Uniform Prior
& $\mathbf p_t^\star$
& \checkmark
& $1/K$
& 59.9 & 6.5 & 2.0\\

Reliability Only
& $\mathbf p_t^\star$
& \checkmark
& $\kappa_{t,k}/\hat{\kappa}_{t,k}$
& 59.6 & 6.8 & 2.0\\

Balancing Only
& $\mathbf p_t^\star$
& \checkmark
& $1/\hat{\kappa}_{t,k}$
& 60.5 & \textbf{5.5} & 2.0\\

\textbf{\ours} (Ours)
& $\mathbf p_t^\star$
& \checkmark
& RB
& \textbf{61.9} & 6.0 & \textbf{1.9}\\

\bottomrule
\end{tabular}
}
\end{table*}

Table~\ref{tab:ablation_state} first examines the reliability-weighted responsibility used for target-state updates in Eq.~\ref{eq:fractional_class_weight}. Removing source support $\zeta$ consistently degrades all assignment variants, indicating that indiscriminately accumulating current predictions can amplify unreliable evidence over time. In particular, weighting the gain-controlled prediction by $\zeta$ improves accuracy from 59.9\% to 61.9\% and reduces ECE from 11.9\% to 6.0\%. This supports our update $\omega_{t,b,k}=\zeta_{t,b}p^\star_{t,b,k}$: predictions weakly supported by the frozen source contribute less to future target statistics. Moreover, using the intervened prediction $\mathbf p_t^\star$ outperforms updating with either the source prediction or the unfiltered target proposal, showing that gain-guided intervention also provides more reliable evidence for subsequent adaptation.
We further ablate the reliability-balanced historical class prior in Eq.~\ref{eq:historical_prior}, which combines reliability-normalized support with inverse-support balancing. Reliability-normalized support favors classes whose accumulated predictions are better supported, while inverse-support balancing prevents frequently predicted classes from progressively dominating the prior. 
The full prior achieves the best accuracy and NLL while maintaining low ECE, reaching 61.9\% accuracy, 1.9 NLL, and 6.0\% ECE.
Together, the reliability-weighted state update and reliability-balanced historical prior play complementary roles in limiting the reinforcement of unreliable predictions and class bias, thereby mitigating error accumulation and maintaining a stable continual target state.


%

\begin{figure}[t]
    \centering
    \setlength{\abovecaptionskip}{1mm}

    \begin{subfigure}[t]{0.44\columnwidth}
        \centering
        \vspace{-2mm}
        \includegraphics[width=\linewidth]{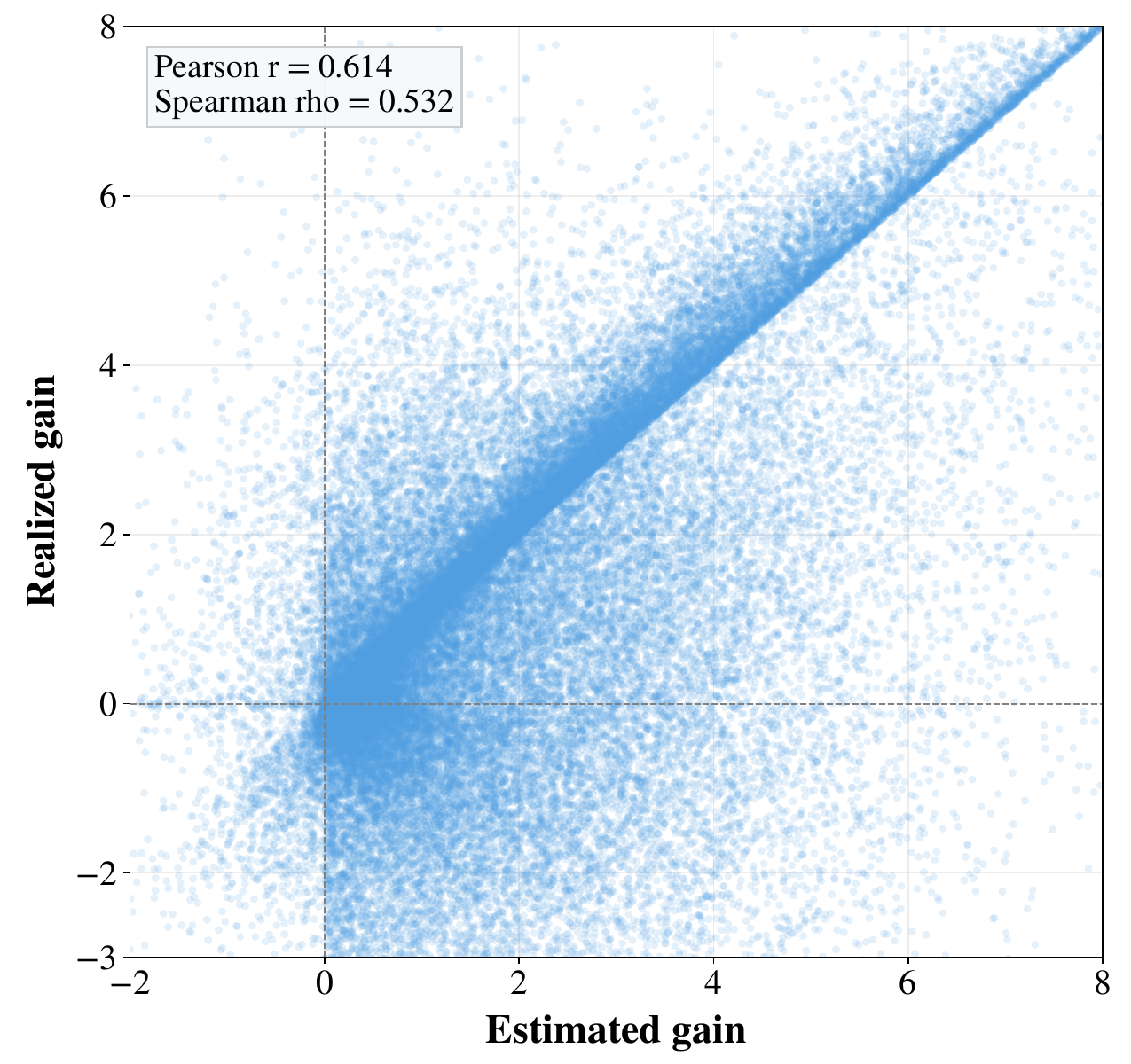}

        \vspace{-2mm}
        \caption{Estimated vs. Realized Correction Gain}
        \label{fig:realgain}
    \end{subfigure}
    \hspace{1mm}
    \begin{subfigure}[t]{0.47\columnwidth}
        \centering
        \vspace{-2mm}
        \includegraphics[width=\linewidth]{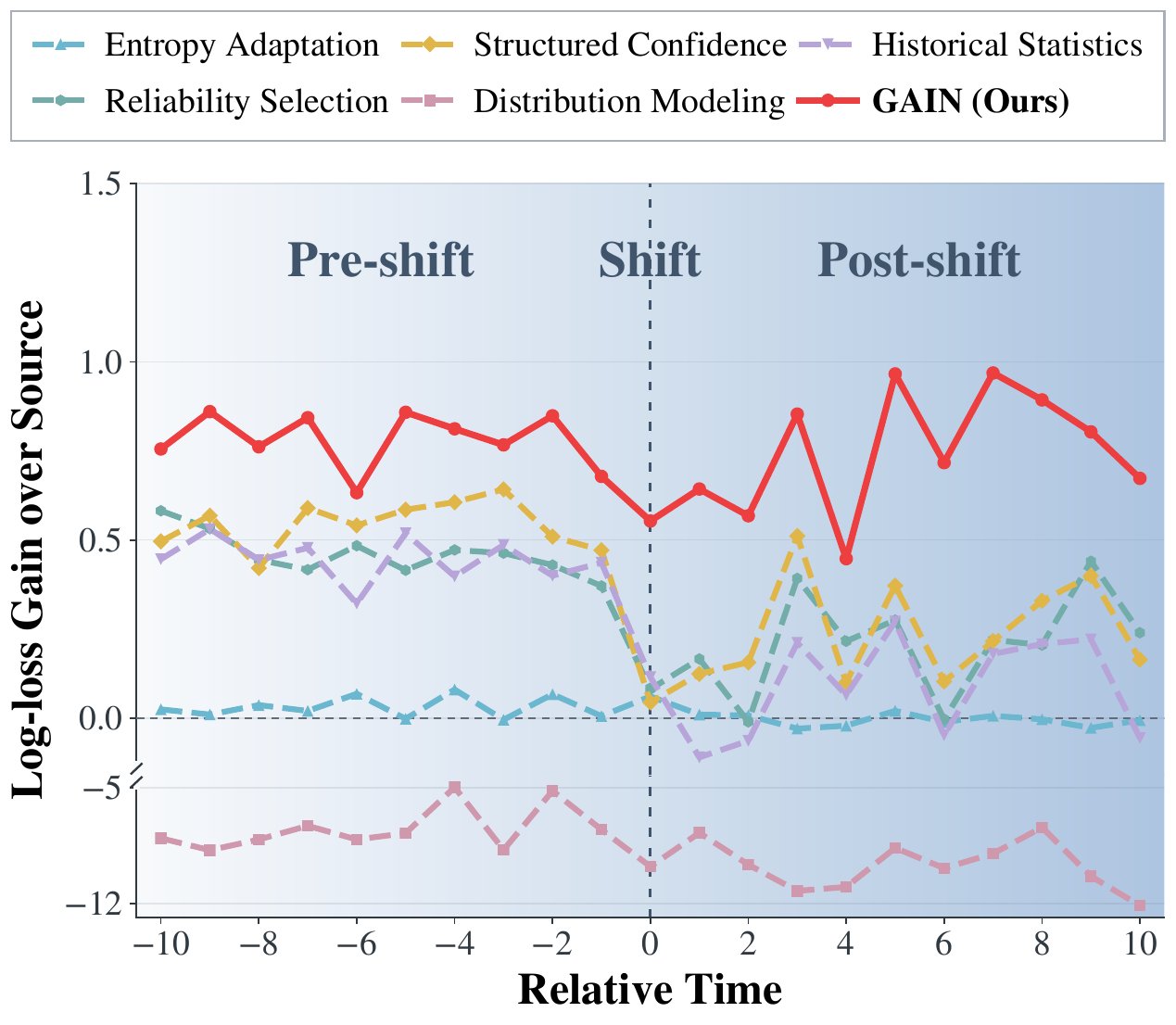}

        \vspace{-2mm}
        \caption{Correction Gain Over Time}
        \label{fig:gain_overtime}
    \end{subfigure}

    \caption{
    \textbf{Correction gain analysis.}
        (a) The estimated posterior-predictive gain is positively associated with the realized source-relative correction gain across test samples.
        (b) Not every correction helps: the source-relative benefit of adaptation varies throughout the stream, and target-driven corrections may yield limited or even negative gain.
    }
    \label{fig:main_figure}
    \vspace{-5mm}
\end{figure}

\vspace{-3mm}
\paragraph{Does Estimated Gain Reflect Correction Utility?}
To assess whether the posterior-predictive gain reflects correction utility, we compare $G_t^{\mathrm{pp}}$ with the realized gain $\log(\hat q_{t,y_t}^{H}/s_{t,y_t})$, measured using ground-truth labels $y_t$ only for retrospective evaluation. This quantifies the full proposal's log-loss improvement over the source prediction, whose conditional expectation corresponds to Eq.~\ref{eq:practical_history_reliability}.
As shown in Fig.~\ref{fig:realgain}, the estimated and realized gains are positively associated across samples (Pearson $r\!=\!0.614$, Spearman $\rho\!=\!0.532$). Despite finite, unlabeled observations and non-stationary target shifts, $G_t^{\mathrm{pp}}$ therefore meaningfully tracks and ranks the utility of a proposed correction.
Unlike a confidence score, \(G_t^{\mathrm{pp}}\) estimates the expected benefit of a specific correction relative to retaining the source prediction.
Fig.~\ref{fig:gain_overtime} further shows why this utility must be evaluated continually. Around distribution shifts, correction benefit changes substantially, with several alternative signals yielding limited or even negative gain, whereas \ours remains consistently positive across the transition. Together, these results directly support our principle: history proposes, while gain determines whether and how strongly to intervene.

\begin{table*}[t]
\centering
\vspace{-3mm}
\caption{
\textbf{CSC results on ImageNet-3DCC.}
Accuracy (Acc., \%) and expected calibration error (ECE, \%) with ViT-Base at severity~5.
BP-free denotes backpropagation-free adaptation.
Bold indicates the best results; Source is shown for reference only.
}
\label{tab:3dcc-csc}
\small
\setlength{\tabcolsep}{1.0pt}
\renewcommand{\arraystretch}{1.05}
\resizebox{\textwidth}{!}{
\begin{tabular}{lcc|cc|ccc|c|c|ccc|cc|c}
\toprule
\multirow{2}{*}{Method}
& \multirow{2}{*}{BP-free}
& \multirow{2}{*}{Metric}
& \multicolumn{2}{c|}{Depth of field}
& \multicolumn{3}{c|}{Noise}
& Lighting
& Weather
& \multicolumn{3}{c|}{Video}
& \multicolumn{2}{c|}{Camera motion}
& \multirow{2}{*}{\textbf{Avg.}} \\
\cmidrule(lr){4-5}
\cmidrule(lr){6-8}
\cmidrule(lr){11-13}
\cmidrule(lr){14-15}
&
&
&
Near foc.
& Far foc.
& Color quant.
& ISO
& Low light
& Flash
& Fog 3D
& Bit err.
& H.265 abr.
& H.265 crf
& XY-mot.
& Z-mot.
& \\
\midrule


\rowcolor{sourcecolor}
& & Acc. \(\uparrow\)
& 71.1 & 62.6 & 55.9 & 62.1 & 61.7 & 45.2 & 44.6 & 36.0 & 71.8 & 77.4 & 45.6 & 49.3 & 56.9 \\
\rowcolor{sourcecolor}
\multirow{-2}{*}{Source}
& \multirow{-2}{*}{--}
& ECE \(\downarrow\)
& 7.0 & 5.0 & 3.9 & 14.9 & 8.2 & 4.3 & 5.5 & 8.6 & 9.7 & 9.7 & 4.1 & 4.9 & 7.2 \\
\midrule

& & Acc. \(\uparrow\)
& 70.8 & 62.4 & 57.1 & 61.1 & 62.2 & 45.4 & 44.9 & 35.7 & 72.0 & 77.3 & 46.1 & 49.6 & 57.0 \\
\multirow{-2}{*}{CoTTA~(\hyperlink{cite.CoTTA}{CVPR 2022})}
& \multirow{-2}{*}{\xmark}
& ECE \(\downarrow\)
& 6.1 & 4.8 & \textbf{3.8} & 8.1 & \textbf{2.7} & 8.0 & \textbf{4.8} & 18.3 & \textbf{3.3} & 3.7 & 15.0 & 18.5 & 8.1 \\
\addlinespace[1.8pt]

& & Acc. \(\uparrow\)
& 70.8 & 62.3 & 57.2 & 60.9 & 62.2 & 45.3 & 44.7 & 35.9 & 72.3 & 77.5 & 46.5 & 50.2 & 57.1 \\
\multirow{-2}{*}{ViDA~(\hyperlink{cite.ViDA}{ICLR 2024})}
& \multirow{-2}{*}{\xmark}
& ECE \(\downarrow\)
& 6.6 & 4.8 & 4.1 & 12.4 & 7.2 & \textbf{4.3} & 4.8 & 10.5 & 7.3 & 6.9 & \textbf{4.9} & \textbf{7.1} & \textbf{6.7} \\
\addlinespace[1.8pt]

& & Acc. \(\uparrow\)
& \textbf{74.9} & \textbf{68.0} & 61.8 & 65.6 & \textbf{70.8} & \textbf{49.7} & \textbf{49.5} & 31.2 & 63.9 & 77.3 & 49.0 & \textbf{57.6} & 60.0 \\
\multirow{-2}{*}{REM~(\hyperlink{cite.REM}{ICML 2025})}
& \multirow{-2}{*}{\xmark}
& ECE \(\downarrow\)
& 3.4 & \textbf{4.5} & 6.1 & \textbf{5.0} & 4.8 & 10.7 & 8.6 & 36.6 & 15.3 & 6.2 & 16.8 & 14.7 & 11.1 \\
\addlinespace[1.8pt]

& & Acc. \(\uparrow\)
& 74.5 & 67.7 & \textbf{62.1} & 63.0 & 69.3 & 48.1 & 41.2 & 34.4 & 74.0 & 78.3 & 48.8 & 53.6 & 59.6 \\
\multirow{-2}{*}{DPCore~(\hyperlink{cite.DPCore}{ICML 2025})}
& \multirow{-2}{*}{\xmark}
& ECE \(\downarrow\)
& 12.6 & 10.9 & 10.4 & 11.5 & 11.6 & 6.7 & 4.8 & \textbf{5.9} & 12.7 & 10.5 & 5.9 & 7.6 & 9.2 \\
\addlinespace[1.8pt]

& & Acc. \(\uparrow\)
& 73.3 & 65.4 & 58.4 & 56.2 & 65.2 & 47.7 & 46.5 & 33.5 & 68.8 & 74.1 & 43.6 & 54.6 & 57.3 \\
\multirow{-2}{*}{PAID~(\hyperlink{cite.PAID}{NeurIPS 2025})}
& \multirow{-2}{*}{\xmark}
& ECE \(\downarrow\)
& \textbf{2.9} & 4.8 & 6.7 & 8.1 & 6.1 & 9.6 & 9.5 & 14.7 & 4.1 & \textbf{3.3} & 13.2 & 8.4 & 7.6 \\
\addlinespace[1.8pt]

& & Acc. \(\uparrow\)
& 71.1 & 63.2 & 57.2 & 63.3 & 63.9 & 46.6 & 45.6 & 36.3 & 74.1 & 78.8 & 48.2 & 51.0 & 58.3 \\
\multirow{-2}{*}{DOTA~(\hyperlink{cite.DOTA}{NeurIPS 2025})}
& \multirow{-2}{*}{\cmark}
& ECE \(\downarrow\)
& 24.4 & 32.0 & 37.7 & 32.4 & 32.5 & 48.0 & 48.8 & 58.3 & 24.1 & 19.9 & 47.8 & 45.5 & 37.6 \\

\midrule
\rowcolor{mycolor}
& & Acc. \(\uparrow\)
& 71.8 & 64.7 & 59.5 & \textbf{65.9} & 66.5 & 48.2 & 46.8 & \textbf{40.3} & \textbf{74.8} & \textbf{78.8} & \textbf{52.4} & 56.5 & \textbf{60.5} \\
\rowcolor{mycolor}
\multirow{-2}{*}{\ours (Ours)}
& \multirow{-2}{*}{\cmark}
& ECE \(\downarrow\)
& 3.1 & 7.2 & 7.8 & 6.4 & 6.5 & 8.4 & 6.9 & 6.2 & 6.8 & 7.0 & 8.2 & 8.3 & 6.9 \\

\bottomrule
\end{tabular}
}
\end{table*}

\begin{table*}[t]
\centering
\caption{
\textbf{CDC results on ImageNet-3DCC.}
Accuracy (Acc.,\%) and expected calibration error (ECE,\%) with ViT-Base at severity~5.
BP-free denotes backpropagation-free adaptation.
Bold indicates the best results; Source is shown for reference only.
Our results are averaged over five runs.
}
\label{tab:3dcc-cdc}
\small
\setlength{\tabcolsep}{1.0pt}
\renewcommand{\arraystretch}{1.05}
\resizebox{0.98\textwidth}{!}{
\begin{tabular}{lcc|cc|ccc|c|c|ccc|cc|c}
\toprule
\multirow{2}{*}{Method}
& \multirow{2}{*}{BP-free}
& \multirow{2}{*}{Metric}
& \multicolumn{2}{c|}{Depth of field}
& \multicolumn{3}{c|}{Noise}
& Lighting
& Weather
& \multicolumn{3}{c|}{Video}
& \multicolumn{2}{c|}{Camera motion}
& \multirow{2}{*}{\textbf{Avg.}} \\
\cmidrule(lr){4-5}
\cmidrule(lr){6-8}
\cmidrule(lr){11-13}
\cmidrule(lr){14-15}
&
&
&
Near foc.
& Far foc.
& Color quant.
& ISO
& Low light
& Flash
& Fog 3D
& Bit err.
& H.265 abr.
& H.265 crf
& XY-mot.
& Z-mot.
& \\
\midrule


\rowcolor{sourcecolor}
& & Acc. \(\uparrow\)
& 71.1 & 62.6 & 55.9 & 62.1 & 61.7 & 45.2 & 44.6 & 36.0 & 71.8 & 77.4 & 45.6 & 49.3 & 56.9 \\
\rowcolor{sourcecolor}
\multirow{-2}{*}{Source}
& \multirow{-2}{*}{--}
& ECE \(\downarrow\)
& 7.0 & 5.0 & 3.9 & 14.9 & 8.2 & 4.3 & 5.5 & 8.6 & 9.7 & 9.7 & 4.1 & 4.9 & 7.2 \\
\midrule

& & Acc. \(\uparrow\)
& 71.1 & 62.4 & 57.1 & 61.9 & 62.4 & 45.0 & 45.2 & 35.7 & 71.9 & 77.4 & 45.5 & 49.6 & 57.1 \\
\multirow{-2}{*}{CoTTA~(\hyperlink{cite.CoTTA}{CVPR 2022})}
& \multirow{-2}{*}{\xmark}
& ECE \(\downarrow\)
& \textbf{3.0} & 5.7 & 6.7 & \textbf{3.4} & \textbf{3.7} & 14.3 & 4.8 & 15.9 & \textbf{3.1} & 5.0 & 9.1 & 10.1 & 7.1 \\
\addlinespace[1.8pt]

& & Acc. \(\uparrow\)
& 71.1 & 62.5 & 57.2 & 61.0 & 62.2 & 45.3 & 44.7 & 36.0 & 72.1 & 77.4 & 45.9 & 49.9 & 57.1 \\
\multirow{-2}{*}{ViDA~(\hyperlink{cite.ViDA}{ICLR 2024})}
& \multirow{-2}{*}{\xmark}
& ECE \(\downarrow\)
& 5.6 & 4.7 & \textbf{3.9} & 10.5 & 7.5 & \textbf{5.2} & \textbf{4.3} & 10.4 & 7.8 & 8.4 & \textbf{4.4} & \textbf{5.7} & \textbf{6.5} \\
\addlinespace[1.8pt]

& & Acc. \(\uparrow\)
& \textbf{74.3} & \textbf{65.2} & 59.2 & 63.8 & \textbf{68.0} & 42.7 & 43.7 & 33.3 & 73.3 & 76.6 & 52.1 & 54.0 & 58.9 \\
\multirow{-2}{*}{REM~(\hyperlink{cite.REM}{ICML 2025})}
& \multirow{-2}{*}{\xmark}
& ECE \(\downarrow\)
& 5.0 & \textbf{3.8} & 9.9 & 7.4 & 6.8 & 31.2 & 30.7 & 28.4 & 7.0 & 6.0 & 14.7 & 15.2 & 13.8 \\
\addlinespace[1.8pt]

& & Acc. \(\uparrow\)
& 72.7 & 63.8 & \textbf{60.2} & 60.2 & 66.7 & \textbf{48.4} & 41.6 & 33.6 & \textbf{74.7} & \textbf{78.9} & 49.6 & 51.9 & 58.5 \\
\multirow{-2}{*}{DPCore~(\hyperlink{cite.DPCore}{ICML 2025})}
& \multirow{-2}{*}{\xmark}
& ECE \(\downarrow\)
& 12.1 & 10.3 & 11.6 & 12.5 & 11.6 & 6.2 & 4.8 & \textbf{7.2} & 12.3 & 12.3 & 6.5 & 8.2 & 9.6 \\
\addlinespace[1.8pt]

& & Acc. \(\uparrow\)
& 70.2 & 61.2 & 54.9 & 54.2 & 64.3 & 45.2 & 45.4 & 34.1 & 71.4 & 77.0 & 46.8 & 53.7 & 56.5 \\
\multirow{-2}{*}{PAID~(\hyperlink{cite.PAID}{NeurIPS 2025})}
& \multirow{-2}{*}{\xmark}
& ECE \(\downarrow\)
& 4.4 & 6.6 & 9.3 & 9.2 & 6.1 & 12.1 & 10.8 & 16.2 & 3.5 & \textbf{2.8} & 12.4 & 9.5 & 8.6 \\
\addlinespace[1.8pt]

& & Acc. \(\uparrow\)
& 72.2 & 63.9 & 57.8 & 63.7 & 63.6 & 46.6 & 45.4 & 36.4 & 73.8 & 77.8 & 47.6 & 50.2 & 58.3 \\
\multirow{-2}{*}{DOTA~(\hyperlink{cite.DOTA}{NeurIPS 2025})}
& \multirow{-2}{*}{\cmark}
& ECE \(\downarrow\)
& 24.8 & 32.9 & 38.6 & 32.6 & 32.0 & 48.4 & 49.7 & 57.1 & 24.1 & 19.7 & 47.0 & 44.3 & 37.6 \\

\midrule
\rowcolor{mycolor}
& & Acc. \(\uparrow\)
& \val{72.9}{0.26} & \val{64.9}{0.86} & \val{59.2}{1.39} & \bval{65.6}{0.44} & \val{66.2}{0.33} & \val{47.4}{0.77} & \bval{46.5}{0.38} & \bval{39.9}{0.27} & \val{73.8}{0.57} & \val{78.3}{0.29} & \bval{52.2}{0.38} & \bval{54.4}{0.70} & \bval{60.1}{0.09}  \\

\rowcolor{mycolor}
\multirow{-2}{*}{\ours (Ours)}
& \multirow{-2}{*}{\cmark}
& ECE \(\downarrow\)
& \val{6.7}{0.51} & \val{6.9}{1.84} & \val{7.5}{0.93} & \val{6.7}{0.52} & \val{6.3}{0.28} & \val{8.0}{1.00} & \val{7.5}{0.25} & \val{7.4}{0.81} & \val{5.7}{0.97} & \val{6.7}{0.51} & \val{7.6}{0.13} & \val{8.1}{0.42} & \val{7.1}{0.10}  \\


\bottomrule
\end{tabular}
}
\vspace{-3mm}
\end{table*}

\begin{table}[t]
\centering
\vspace{-3mm}
\caption{
\textbf{MDS results on ImageNet-3DCC.}
Accuracy (Acc., \%) and expected calibration error (ECE, \%) across severity levels~5--1.
BP-free denotes backpropagation-free adaptation.
Bold indicates the best results; Source is shown for reference only.
Our results are averaged over five runs.
}
\label{tab:3dcc-mds}
\small
\setlength{\tabcolsep}{6.5pt}
\renewcommand{\arraystretch}{1.1}
\resizebox{0.98\linewidth}{!}{%
\begin{tabular}{lcc|ccccc|c}
\toprule
Method
& BP-free
& Metric
& Level 5
& Level 4
& Level 3
& Level 2
& Level 1
& \textbf{Avg.} \\
\midrule


\rowcolor{sourcecolor}
& & Acc. \(\uparrow\)
& 56.9 & 64.1 & 69.4 & 73.6 & 76.6 & 68.1 \\
\rowcolor{sourcecolor}
\multirow{-2}{*}{Source}
& \multirow{-2}{*}{--}
& ECE \(\downarrow\)
& 4.3 & 5.9 & 6.7 & 7.7 & 8.3 & 6.6 \\
\midrule

\multirow{2}{*}{CoTTA~(\hyperlink{cite.CoTTA}{CVPR 2022})}
& \multirow{2}{*}{\xmark}
& Acc. \(\uparrow\)
& 57.1 & 64.3 & 69.6 & 73.8 & 76.7 & 68.3 \\
&
& ECE \(\downarrow\)
& 4.8 & \textbf{3.3} & \textbf{2.7} & \textbf{2.2} & \textbf{2.2} & \textbf{3.0} \\
\addlinespace[1.8pt]

\multirow{2}{*}{ViDA~(\hyperlink{cite.ViDA}{ICLR 2024})}
& \multirow{2}{*}{\xmark}
& Acc. \(\uparrow\)
& 57.2 & 64.3 & 69.6 & 73.7 & 76.7 & 68.3 \\
&
& ECE \(\downarrow\)
& \textbf{3.5} & 4.2 & 5.0 & 5.9 & 6.4 & 5.0 \\
\addlinespace[1.8pt]

\multirow{2}{*}{REM~(\hyperlink{cite.REM}{ICML 2025})}
& \multirow{2}{*}{\xmark}
& Acc. \(\uparrow\)
& 58.3 & 65.4 & 70.7 & 74.8 & \textbf{77.6} & 69.4 \\
&
& ECE \(\downarrow\)
& 6.4 & 6.8 & 7.1 & 7.7 & 8.1 & 7.2 \\
\addlinespace[1.8pt]

\multirow{2}{*}{DPCore~(\hyperlink{cite.DPCore}{ICML 2025})}
& \multirow{2}{*}{\xmark}
& Acc. \(\uparrow\)
& 57.7 & 64.9 & 69.3 & 73.2 & 76.3 & 68.3 \\
&
& ECE \(\downarrow\)
& 7.5 & 10.0 & 9.8 & 9.4 & 9.6 & 9.3 \\
\addlinespace[1.8pt]

\multirow{2}{*}{PAID~(\hyperlink{cite.PAID}{NeurIPS 2025})}
& \multirow{2}{*}{\xmark}
& Acc. \(\uparrow\)
& 55.1 & 62.5 & 68.2 & 72.8 & 76.2 & 67.0 \\
&
& ECE \(\downarrow\)
& 6.5 & 4.8 & 3.8 & 2.9 & 2.2 & 4.1 \\
\addlinespace[1.8pt]

\multirow{2}{*}{DOTA~(\hyperlink{cite.DOTA}{NeurIPS 2025})}
& \multirow{2}{*}{\cmark}
& Acc. \(\uparrow\)
& 57.0 & 64.1 & 69.5 & 73.6 & 76.6 & 68.1 \\
&
& ECE \(\downarrow\)
& 36.9 & 31.0 & 26.6 & 23.0 & 20.6 & 27.6 \\

\midrule
\rowcolor{mycolor}
& & Acc. \(\uparrow\)
& \textbf{58.9} \(\pm\) 0.07 & \textbf{65.8} \(\pm\) 0.08 & \textbf{70.8} \(\pm\) 0.05 & \textbf{74.9} \(\pm\) 0.05 &  \(77.5 \pm 0.05\) & \textbf{69.6} \(\pm\) 0.04 \\
\rowcolor{mycolor}
\multirow{-2}{*}{\ours (Ours)}
& \multirow{-2}{*}{\cmark}
& ECE \(\downarrow\)
& \(6.2 \pm 0.10\) & \(6.7 \pm 0.09\) & \(7.0 \pm 0.10\) & \(7.0 \pm 0.09\) & \(7.2 \pm 0.04\) & \(6.8 \pm 0.07\) \\

\bottomrule
\end{tabular}
}
\end{table}

\begin{table*}[t]
\centering
\caption{
\textbf{LHA results on ImageNet-3DCC.}
Accuracy (Acc., \%) and expected calibration error (ECE, \%) over 10 repeated corruption cycles (R1–R10) with ViT-Base at severity 5.
BP-free denotes backpropagation-free adaptation.
Bold indicates the best results; Source is shown for reference only.
}
\label{tab:3dcc-lha}
\small
\setlength{\tabcolsep}{6pt}
\renewcommand{\arraystretch}{1.1}
\resizebox{0.98\linewidth}{!}{
\begin{tabular}{lcc|cccccccccc|c}
\toprule
Method
& BP-free
& Metric
& R1
& R2
& R3
& R4
& R5
& R6
& R7
& R8
& R9
& R10
& \textbf{Avg.} \\
\midrule


\rowcolor{sourcecolor}
& & Acc. \(\uparrow\)
& 56.9 & 56.9 & 56.9 & 56.9 & 56.9 & 56.9 & 56.9 & 56.9 & 56.9 & 56.9 & 56.9 \\
\rowcolor{sourcecolor}
\multirow{-2}{*}{Source}
& \multirow{-2}{*}{--}
& ECE \(\downarrow\)
& 7.2 & 7.2 & 7.2 & 7.2 & 7.2 & 7.2 & 7.2 & 7.2 & 7.2 & 7.2 & 7.2 \\
\midrule

\multirow{2}{*}{CoTTA~(\hyperlink{cite.CoTTA}{CVPR 2022})}
& \multirow{2}{*}{\xmark}
& Acc. \(\uparrow\)
& 57.0 & 57.0 & 57.1 & 56.9 & 56.8 & 56.8 & 56.9 & 57.1 & 57.1 & 57.0 & 57.0 \\
&
& ECE \(\downarrow\)
& 8.1 & 17.3 & 23.8 & 27.4 & 29.5 & 31.1 & 32.1 & 33.2 & 33.7 & 34.3 & 27.0 \\
\addlinespace[1.8pt]

\multirow{2}{*}{ViDA~(\hyperlink{cite.ViDA}{ICLR 2024})}
& \multirow{2}{*}{\xmark}
& Acc. \(\uparrow\)
& 57.1 & 57.9 & 58.3 & 58.6 & 58.8 & 58.9 & 59.0 & 59.0 & 59.1 & 59.1 & 58.6 \\
&
& ECE \(\downarrow\)
& \textbf{6.7} & \textbf{6.0} & \textbf{6.8} & 8.9 & 11.2 & 13.1 & 14.7 & 16.1 & 17.2 & 18.1 & 11.9 \\
\addlinespace[1.8pt]

\multirow{2}{*}{REM~(\hyperlink{cite.REM}{ICML 2025})}
& \multirow{2}{*}{\xmark}
& Acc. \(\uparrow\)
& 60.0 & 58.1 & 58.7 & 58.1 & 56.8 & 50.5 & 36.7 & 0.3 & 0.1 & 0.1 & 37.9 \\
&
& ECE \(\downarrow\)
& 11.1 & 18.1 & 19.7 & 21.7 & 24.4 & 33.5 & 51.3 & 99.5 & 99.9 & 99.9 & 47.9 \\
\addlinespace[1.8pt]

\multirow{2}{*}{DPCore~(\hyperlink{cite.DPCore}{ICML 2025})}
& \multirow{2}{*}{\xmark}
& Acc. \(\uparrow\)
& 59.6 & 59.4 & 58.5 & 58.4 & 58.1 & 57.7 & 57.5 & 57.2 & 56.8 & 56.6 & 58.0 \\
&
& ECE \(\downarrow\)
& 9.2 & 8.3 & 8.5 & 8.8 & 9.0 & 9.0 & 9.3 & 9.5 & 9.6 & 9.6 & 9.1 \\
\addlinespace[1.8pt]

\multirow{2}{*}{PAID~(\hyperlink{cite.PAID}{NeurIPS 2025})}
& \multirow{2}{*}{\xmark}
& Acc. \(\uparrow\)
& 57.3 & 55.1 & 52.8 & 51.0 & 49.4 & 48.4 & 47.1 & 46.0 & 45.1 & 44.2 & 49.6 \\
&
& ECE \(\downarrow\)
& 7.6 & 8.1 & 8.3 & 8.4 & 8.6 & 8.8 & 9.1 & 9.4 & 9.7 & 9.8 & 8.8 \\
\addlinespace[1.8pt]

\multirow{2}{*}{DOTA~(\hyperlink{cite.DOTA}{NeurIPS 2025})}
& \multirow{2}{*}{\cmark}
& Acc. \(\uparrow\)
& 58.3 & 58.5 & 58.5 & 58.5 & 58.5 & 58.5 & 58.5 & 58.4 & 58.4 & 58.4 & 58.4 \\
&
& ECE \(\downarrow\)
& 37.6 & 38.7 & 38.8 & 38.8 & 38.8 & 38.9 & 38.9 & 38.9 & 38.9 & 38.9 & 38.7 \\

\midrule
\rowcolor{mycolor}
& & Acc. \(\uparrow\)
& \textbf{60.5} & \textbf{60.8} & \textbf{60.8} & \textbf{60.8}
& \textbf{60.8} & \textbf{60.8} & \textbf{60.8} & \textbf{60.8}
& \textbf{60.8} & \textbf{60.7} & \textbf{60.7} \\
\rowcolor{mycolor}
\multirow{-2}{*}{\ours (Ours)}
& \multirow{-2}{*}{\cmark}
& ECE \(\downarrow\)
& 6.9 & 8.0 & 8.2 & \textbf{8.3}
& \textbf{8.3} & \textbf{8.4} & \textbf{8.4} & \textbf{8.4}
& \textbf{8.4} & \textbf{8.4} & \textbf{8.2} \\

\bottomrule
\end{tabular}
}
\end{table*}
\vspace{-3mm}
\subsection{More Results on ImageNet-3DCC}
\vspace{-2mm}
Tables~\ref{tab:3dcc-csc}--\ref{tab:3dcc-lha} further evaluate \ours on ImageNet-3DCC under CSC, CDC, MDS, and LHA. Compared with ImageNet-C, ImageNet-3DCC introduces more diverse shifts involving depth of field, lighting and weather, video compression, and camera motion, providing a complementary test of adaptation under heterogeneous distribution changes. \ours achieves the highest average accuracy under CSC (60.5\%), CDC (60.1\%), and MDS (69.6\%), while maintaining competitive calibration. The long-horizon setting further stresses error accumulation over 10 repeated corruption cycles without reset. \ours remains stable throughout the stream, maintaining 60.5--60.8\% accuracy with an average ECE of 8.2\%, whereas several baselines exhibit substantial accuracy degradation or calibration drift. These results indicate that gain-guided intervention generalizes beyond ImageNet-C to more diverse corruption mechanisms, dynamic and mixed shifts, and prolonged continual adaptation.

\vspace{-3mm}
\subsection{Evaluation under Test-Time Adaptation}
\vspace{-2mm}
%
Beyond continual test-time adaptation, we further evaluate \ours under standard test-time adaptation (TTA) to examine its generalization to non-continual domain shifts. Following prior TTA evaluation, we consider shifts from ImageNet to ImageNet-R, ImageNet-V2, and ImageNet-Sketch, and compare against representative adaptation methods. Table~\ref{tab:domain_shift} reports the top-1 accuracy on each target domain and their average. \ours achieves the highest mean accuracy of 63.4\%, with the best performance on ImageNet-V2 and ImageNet-Sketch, demonstrating that the proposed gain-guided intervention remains effective beyond continual adaptation.

\begin{table}[t]
\centering
\caption{
\textbf{TTA results under domain shifts.} Top-1 accuracy on ImageNet-R, ImageNet-V2, and ImageNet-Sketch. BP-free denotes backpropagation-free adaptation. Bold indicates the best results.
}
\label{tab:domain_shift}
\small
\setlength{\tabcolsep}{6pt}
\renewcommand{\arraystretch}{1.2}
\begin{tabular}{lccccc}
\toprule
Method
& BP-free
& ImageNet-R
& ImageNet-V2
& ImageNet-Sketch
& \textbf{Avg.} \\
\midrule

Source
& --
& 59.5
& 75.4
& 44.9
& 59.9 \\

Tent~(\hyperlink{cite.Tent}{ICLR 2021})
& \xmark
& 63.9
& 75.2
& 49.1
& 62.7 \\

CoTTA~(\hyperlink{cite.CoTTA}{CVPR 2022})
& \xmark
& 63.5
& 75.4
& 50.0
& 63.0 \\

SAR~(\hyperlink{cite.SAR}{ICLR 2023})
& \xmark
& 63.3
& 75.1
& 48.7
& 62.4 \\

FOA~(\hyperlink{cite.FOA}{ICML 2024})
& \cmark
& 63.8
& 75.4
& 49.9
& 63.0 \\

REM~(\hyperlink{cite.REM}{ICML 2025})
& \xmark
& \textbf{64.3}
& 75.2
& 49.7
& 63.1 \\

\rowcolor{mycolor}
\ours (Ours)
& \cmark
& 63.0
& \textbf{75.8}
& \textbf{51.5}
& \textbf{63.4} \\

\bottomrule
\end{tabular}
\end{table}

\end{document}